\documentclass{article}

\usepackage[preprint]{neurips_2024}

\usepackage[utf8]{inputenc} 
\usepackage[T1]{fontenc}    
\usepackage{hyperref}       
\usepackage{url}            
\usepackage{booktabs}       
\usepackage{amsfonts}       
\usepackage{nicefrac}       
\usepackage{microtype}      
\usepackage{xcolor}         

\usepackage{wrapfig}
\usepackage{graphicx}
\usepackage{subfigure}
\usepackage{algorithm}
\usepackage{algorithmic}
\usepackage{amsfonts}
\usepackage{multirow}
\usepackage{amssymb}
\usepackage{newtxmath}
\usepackage{url}
\usepackage{bm}
\usepackage{marvosym}
\usepackage{enumitem}
\usepackage[title]{appendix}
\usepackage{subcaption}
\usepackage{listings}
\definecolor{codeblue}{rgb}{0.2,0.2,0.8}
\definecolor{codegray}{rgb}{0.5,0.5,0.5}
\definecolor{codepurple}{rgb}{0.58,0,0.82}
\definecolor{backcolour}{rgb}{0.95,0.95,0.92}
\definecolor{redcolour}{rgb}{0.8,0.0,0.0}
\usepackage[capitalize]{cleveref}
    \crefname{figure}{Figure}{Figures}
    \Crefname{figure}{Figure}{Figures}
    \crefname{section}{Section}{Sections}
    \Crefname{section}{Section}{Sections}
    \crefname{table}{Table}{Tables}
    \Crefname{table}{Table}{Tables}

\title{\includegraphics[width=0.9cm,keepaspectratio]{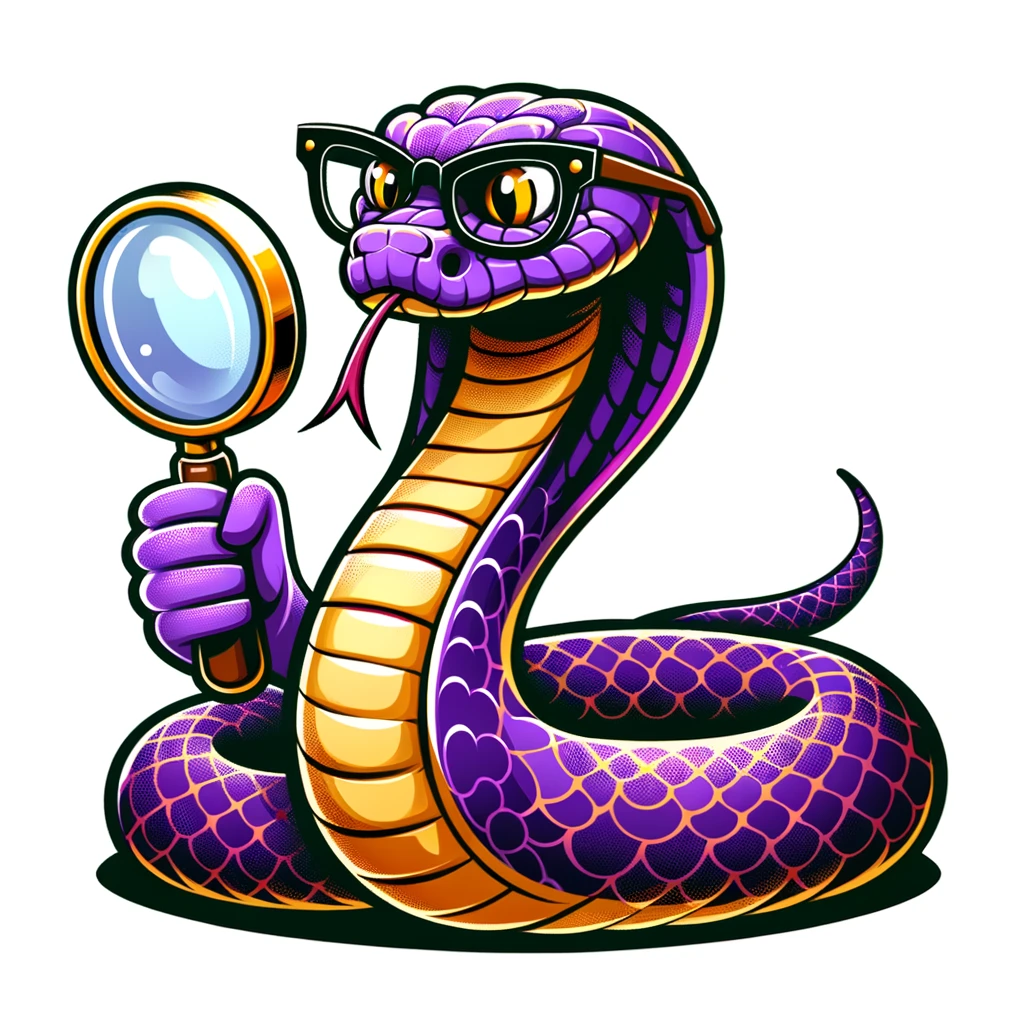}DeMamba: AI-Generated Video Detection on Million-Scale \includegraphics[width=3.2cm,keepaspectratio]{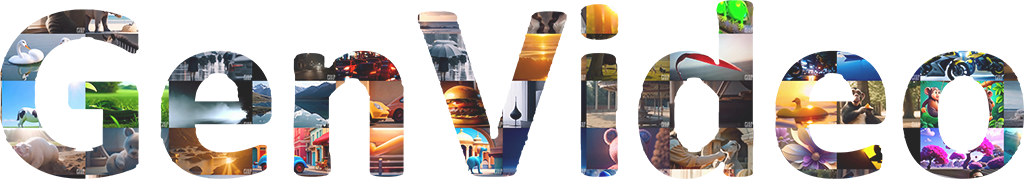} Benchmark}

\author{%
Haoxing Chen$^{1}\textsuperscript{†}$, 
Yan Hong$^{1}\textsuperscript{†}$, 
Zizheng Huang$^{1,2}$,
Zhuoer Xu$^{1}$ ,
Zhangxuan Gu$^{1*}$, \\
\textbf{
Yaohui Li$^{2}$, 
Jun Lan$^{1}$, 
Huijia Zhu$^{1}$, 
Jianfu Zhang$^{3}$\thanks{Corresponding author.\textsuperscript{†} Equal contribution.},
Weiqiang Wang$^{1}$,Huaxiong Li$^{2}$}
\\
\text{$^{1}$Ant Group }\\
\text{$^{2}$Nanjing University} \\ 
\text{$^{3}$Shanghai Jiao Tong University} \\
\texttt{hx.chen@hotmail.com}
}

\begin{document}

\maketitle
\begin{abstract}
Recently, video generation techniques have advanced rapidly. Given the popularity of video content on social media platforms, these models intensify concerns about the spread of fake information.
Therefore, there is a growing demand for detectors capable of distinguishing between fake AI-generated videos and mitigating the potential harm caused by fake information. However, the lack of large-scale datasets from the most advanced video generators poses a barrier to the development of such detectors.
To address this gap, we introduce the first AI-generated video detection dataset, GenVideo. It features the following characteristics: (1) a large volume of videos, including over one million AI-generated and real videos collected; (2) a rich diversity of generated content and methodologies, covering a broad spectrum of video categories and generation techniques.
We conducted extensive studies of the dataset and proposed two evaluation methods tailored for real-world-like scenarios to assess the detectors' performance: the cross-generator video classification task assesses the generalizability of trained detectors on generators; the degraded video classification task evaluates the robustness of detectors to handle videos that have degraded in quality during dissemination.
Moreover, we introduced a plug-and-play module, named Detail Mamba (DeMamba), designed to enhance the detectors by identifying AI-generated videos through the analysis of inconsistencies in temporal and spatial dimensions. Our extensive experiments demonstrate DeMamba's superior generalizability and robustness on GenVideo compared to existing detectors. 
We believe that the GenVideo dataset and the DeMamba module will significantly advance the field of AI-generated video detection. Our code and dataset will be aviliable at \url{https://github.com/chenhaoxing/DeMamba}.
\end{abstract}

\section{Introduction}
Advancements in generative models \citep{controlnet,diffute,SnapFusion} have been impressive, enabling the creation of highly realistic images with less effort and expertise. 
As these models become capable of generating sufficiently realistic images, more researchers are exploring how to improve video creation~\citep{SVD,Evalcrafter,Lavie,veo}.
Currently, certain generative algorithms, such as Sora~\citep{sora} and Gen2~\citep{Gen2}, are capable of producing high-quality videos through the use of straightforward inputs, including text and images.
While these generative algorithms can reduce manual labor and enhance creativity, they also introduce risks~\cite{barrett2023identifying}. For example, they could be utilized to misinform the public in critical domains such as politics or economics. A notable incident involved an AI-generated video of Taylor Swift that spread widely on Twitter, harming her reputation. This situation highlights the pressing need for technology that can detect these fake videos and avoid potential harm.

To assist in developing robust and highly generalizable detectors, we have created the first million-scale dataset of AI-generated videos, named GenVideo. 
GenVideo leverages state-of-the-art models to generate massive amounts of video, providing comprehensive training and validation for detectors of AI-generated videos.
Unlike deepfake video datasets~\citep{stil,TALL,HCIL} which focus on human face videos, GenVideo encompasses a broad spectrum of scene contents and motion variations, closely simulating the real-world authentication challenges posed by video generation models in various practical settings. 
GenVideo includes 1,078,838 generated videos and 1,223,511 real videos. The fake videos consist of those generated in-house and those collected from the internet, while the real videos come from the Youku-mPLUG~\citep{Youku-mPLUG}, Kinetics-400~\citep{K400}, and MSR-VTT~\citep{MSR-VTT} datasets. Due to the scale of the data, we can prevent detectors from merely learning the content differences between real and fake videos, instead focusing on subtle signs that determine video authenticity.
We propose two tasks that align with real-world detection challenges: (1) cross-generator video classification, where a trained detector is tasked with identifying videos from unseen generators; and (2) degraded video classification, where the detector assesses videos that have been degraded, such as those with low resolution, compression artifacts, or Gaussian blur. GenVideo can significantly advance the development of detectors aimed at identifying AI-generated videos in society.

\begin{figure*}[t]
	\centering
 \caption{Spatial and temporal artifacts in generated videos. We illustrate the spatial and temporal artifacts present in the generated videos. Artifacts: (a) errors in local appearance, (b) frequency inconsistency: average spectrum of video frames for real videos and fake videos generated, (c) temporal inconsistency.}
	\includegraphics[width=0.99\linewidth]{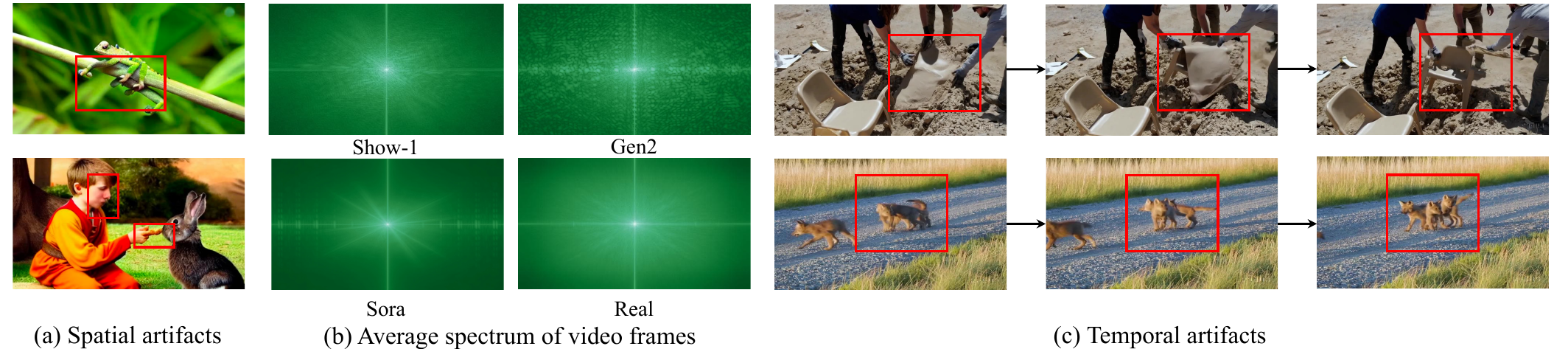}
	\label{Fig1_DeMamba}
\end{figure*}

In this paper, we evaluated state-of-the-art detection models~\citep{f3net,NPR,stil,XCLIP,CLIP} on GenVideo. However, the generalization capabilities of these models are compromised due to the limitations of existing image detection methods, which cannot model temporal inconsistencies, and video detection methods, which struggle to efficiently model local spatial inconsistencies.
In the show ``Detail''~\footnote{https://www.youtube.com/watch?v=NZmEY3n97P4}, Kobe Bryant offered insights into basketball nuances, including Jason Tatum's toe positioning when catching the ball, advice that significantly improved Tatum's ability to penetrate defenses. As shown in Figure~\ref{Fig1_DeMamba}, generated videos often exhibit both spatial and temporal artifacts, and modeling only one aspect (either spatial or temporal) may not be sufficient to cover all types of artifacts. Building a detector with satisfactory generalization performance requires modeling the spatial-temporal local details. In this paper, we introduce a plug-and-play module called Detail Mamba (DeMamba), which leverages a structured state space model to capture spatial-temporal inconsistencies across different regions, thereby discerning the authenticity of videos. Extensive experiments on GenVideo demonstrate that DeMamba can be used as a plug-and-play addition to existing feature extractors, significantly enhancing the generalizability and robustness of models.


Our contributions are summarized as follows:
\begin{itemize}[leftmargin=*]
    \item We introduce the first million-scale dataset for AI-generated video detection, GenVideo, which includes fake videos from various scenes, contents, and models.
    \item We design two tasks to evaluate the performance of detectors: cross-generator video classification and degraded video classification.
    \item We propose a plug-and-play detector, DeMamba, capable of modeling spatial-temporal inconsistencies. Extensive experimental results validate the generalizability and robustness of our DeMamba in identifying AI-generated videos.
\end{itemize}

\section{Related works}
\subsection{Video generation methods}
Video generation methods~\cite{henschel2024streamingt2v,zhou2024storydiffusion} have become powerful tools for producing high-quality video content from textual or image prompts.
Currently, video generation primarily encompasses two major tasks: Text-to-Video (T2V) and Image-to-Video (I2V).
T2V involves inputting a text prompt to the model to generate videos based on textual instructions, while I2V aims to generate videos based on an input image, describing videos' content or specific frames. 
Based on the types of these video generation methods, they can be separated into three categories: diffusion-based methods with Unets, diffusion-based methods with Transformers, and other methods.

\noindent\textbf{Diffusion-Unet}.
In the realm of video synthesis~\cite{ma2024magic,zhang2024moonshot,bar2024lumiere,wei2023dreamvideo,ho2022imagen,girdhar2023emu,feng2023dreamoving,xu2023magicanimate,hu2023animate,ni2023conditional,girdhar2023emu} recent advancements in diffusion-unet based methods have demonstrated significant progress. 
Text2Video-Zero\cite{khachatryan2023text2video} introduces a cost-effective approach, enriching latent codes with motion dynamics for temporal consistency.
Animatediff~\cite{guo2023animatediff} introduces a plug-and-play motion module for animating personalized text-to-image models without model-specific tuning, while Pia~\cite{zhang2023pia} designs a personalized image animator for motion controllability aligned with conditioned images. 
Similarly, Lavie~\cite{wang2023lavie} leverages temporal self-attentions and rotary positional encoding, emphasizing joint image-video fine-tuning for high-quality video outputs.
I2VGen-XL~\cite{I2vgen-xl} makes preliminary attempts in video generation with visual guidance from images.
VideoCrafter~\cite{chen2023videocrafter1, chen2024videocrafter2} and DynamiCrafter~\cite{xing2023dynamicrafter} enhance video quality through synthesized images and leveraging motion priors from text-to-video models to animate open-domain images. 
ModelScopeT2V~\cite{wang2023modelscope} incorporates spatial-temporal blocks to ensure consistent frame generation and smooth movement transitions. 
SVD~\cite{blattmann2023stable} evaluates the critical multi-stage training required for successful video latent diffusion models. 
VideoComposer~\cite{wang2024videocomposer} creatively concatenates image embeddings with style embeddings to enhance the visual continuity of generated videos. SEINE~\cite{chen2023seine} proposes a random-mask video diffusion model to push the boundaries of text-driven video synthesis.

\noindent\textbf{Diffusion-Transformer}.
In the evolving field of video generation, diffusion-transformer based methods~\citep{sora,OpenSora} have garnered considerable attention due to their flexibility and efficiency in handling sequential data. 
Latte~\citep{ma2024latte} enhances the transformer methodology by extracting spatio-temporal tokens from input videos and modeling video distribution in the latent space using a series of Transformer blocks.  Cogvideo~\citep{hong2022cogvideo} leverages a transformer-based model optimized with a multi-frame-rate hierarchical training strategy, which enhances learning efficiency and video quality. Sora~\citep{OpenSora,sora} adopts a DiT-based generative architecture~\citep{li2022dit}, for video generation, showcasing the versatility of transformer-based models in adapting to the specific demands of video synthesis.

\noindent\textbf{Others}.
In addition to diffusion-based models, Generative Adversarial Network(GAN)~\cite{shen2023mostgan,wang2023styleinv} and autoregressive models~\citep{veo,kondratyuk2023videopoet} are also applied for video generation. Notable contributions in this domain include\citep{yoo2023towards} and \citep{lei2023flashvideo}, who have explored the fundamentals of transformer applications in video generation. 
FlashVideo~\citep{lei2023flashvideo} focuses on accelerating transformer models for video generation. 
VideoPoet~\citep{kondratyuk2023videopoet} utilizes a decoder-only transformer architecture to process multimodal inputs and generate creative videos. 
Magvit~\citep{yu2023magvit} employs a masked generative video transformer that quantizes videos into spatial-temporal visual tokens using a 3D tokenizer. 
A few works~\citep{shen2023mostgan,ghosh2024raven,wang2023styleinv} introduce temporal layers into GAN for video generation.

\subsection{AI-Generated content detection}
AI-generated visual content may raise concerns about the spread of misinformation. As a result, considerable efforts have been made to design forgery detection models and establish benchmarks in this field. In recent years, a significant amount of research has focused on detecting generated images~\citep{guo2023hierarchical,lorenz2023detecting,wu2023generalizable,DIRE} with help of AI-Generated Image datasets~\citep{wang2022diffusiondb,zhu2023genimage,DIRE}, particularly those from unseen generative models. To date, studies in \citep{stil,TALL,HCIL} have addressed the detection of Deepfake videos, but there is a lack of research specifically dedicated to detecting generated videos on a wider range beyond human faces. We hope that this paper will make pioneering and insightful contributions to this research area.

\begin{table*}[t]
\centering
\caption{Statistics of real and generated videos in the GenVideo dataset.}
\resizebox{1\columnwidth}{!} {
\begin{tabular}{l|c|ccccc|c|c|c}
\toprule
\textbf{Video Source} & \textbf{Type} &\textbf{Task}  & \textbf{Time} & \textbf{Resolution} & \textbf{FPS} & \textbf{Length} & \textbf{Training Set} & \textbf{Testing Set} & \textbf{Total Count} \\
\midrule
Kinetics-400~\citep{kay2017kinetics} &\multirow{2}{*}{Real}  & - &17.05&224-340&-&5-10s&  260,232 & - & \multirow{2}{*}{1,213,511} \\
Youku-mPLUG~\citep{Youku-mPLUG} & & -&23.07&-& - &10-120s& 953,279 & - &  \\
\midrule
MSR-VTT~\citep{xu2016msr} & Real &-&16.05&-&-&10-30s& - & 10,000 & 10,000 \\
\midrule
ZeroScope~\citep{zeroscopexl} & \multirow{10}{*}{Fake} &T2V&23.07&1024$\times$576&8& 3s& 133,169 & - &  \\
I2VGen-XL~\citep{I2vgen-xl} & &I2V&23.12&1280$\times$720&8& 2s& 61,975 & - &  \\
SVD~\citep{blattmann2023stable} & &I2V&23.12&1024$\times$576&8&4s& 149,026 & - &  \\
VideoCrafter~\citep{chen2024videocrafter2} &  &T2V&24.01&1024$\times$576&8& 2s& 39,485 & - & \multirow{10}{*}{1,048,575} \\
Pika~\citep{pika} & &T2V\&I2V&24.02&1088$\times$640&24& 3s& 98,377 & - & \\
DynamiCrafter~\citep{xing2023dynamicrafter} & &I2V&24.03&1024$\times$576&8& 3s& 46,205 & - &\\
SD~\citep{zhang2023pia} & &T2V\&I2V&23-24&512-1024&8&2-6s& 200,720 & - &  \\
SEINE~\citep{chen2023seine} & &I2V&24.04&1024$\times$576&8& 2-4s& 24,737 & - &  \\
Latte~\citep{ma2024latte} & &T2V&24.03&512$\times$512&8&2s& 149,979 & - &  \\
OpenSora~\citep{OpenSora} & &T2V&24.03&512$\times$512&8&2s& 177,410 & - &  \\
\midrule
ModelScope~\citep{wang2023modelscope} & \multirow{10}{*}{Fake} &T2V&23.03&256$\times$256&8&4s& - & 700 & \multirow{10}{*}{8,588} \\
MorphStudio~\citep{Morphstudio} & &T2V&23.08&1280$\times$720&8&2s& - & 700 & \\
MoonValley~\citep{moonvalley} & &T2V&24.01&1024$\times$576&16&3s& - & 626 &  \\
HotShot~\citep{Hotshot} & &T2V&23.10&672$\times$384&8& 1s& - & 700 &  \\
Show\_1~\citep{zhang2023show} & &T2V&23.10&576$\times$320&8&4s& - & 700 &  \\
Gen2~\citep{esser2023structure} & &I2V\&T2V&23.09&896$\times$512&24&4s& - & 1,380 &  \\
Crafter~\citep{chen2023videocrafter1} & &T2V&23.04&256$\times$256&8&4s& - & 1,400 &  \\
Lavie~\citep{Lavie} & &T2V&23.09&1280$\times$2048&8&2s& - & 1,400 &  \\
Sora~\citep{sora} & &T2V&24.02&-&-&-60s &- & 56 &  \\
WildScrape & &T2V\&I2V&24&512-1024&8-16&2-6s& - & 926 &  \\
\midrule
\textbf{Total Count} & - &-&-&-&-&- & 2,294,594 & 19,588 & 2,314,182 \\
\bottomrule
\end{tabular}
}
\label{tab:dataset}
\end{table*}

\section{GenVideo} \label{sec:dataset}

\subsection{Overviews of GenVideo} \label{sec:overview}
In response to the critical need for evaluating the generalizability of datasets and detectors (\textit{i.e.}, the capacity of training detectors to accurately recognize unseen videos from the open world) and the robustness of these detectors (\textit{i.e.}, their ability to maintain high performance against various corruptions to fake videos), we have developed the GenVideo dataset.
This dataset is characterized by two main features:
\begin{itemize}[leftmargin=*]
    \item Large scale: The GenVideo dataset is organized hierarchically, encompassing cross generators such as diffusion-based generators and transformer-based generators, and cross architectures within the same type of generator like different motion modules combined with the same T2I base model \cite{guo2023animatediff,zhang2023pia}. This structure facilitates covering a broader range of generated content and producing fake videos on a larger scale. The training (\emph{resp.}, testing) set in GenVideo contains a total of $2,294,594$ (\emph{resp.}, $19,588$) video clips, comprising $1,213,511$ (\emph{resp.}, $10,000$) real videos and $1,081,083$ (\emph{resp.}, $8,588$) fake videos.
    \item Diverse content: GenVideo includes a wide array of high-quality fake videos sourced from open-source websites, along with videos produced using both user-trained and officially provided pre-trained video generation models, including T2V and I2V models. The generated video content encompasses a diverse range of scenes, including landscapes, people, buildings, objects, and more. The duration of the videos is primarily between 2 to 6 seconds, and the aspect ratios of the video resolutions vary widely. This diverse collection ensures a comprehensive set of fake videos, significantly enriching the understanding of AI-generated video detection across numerous real-world contexts, and enhancing the generalizability and robustness of detectors.
\end{itemize}

\noindent\textbf{Evaluation objectives}.
To avoid trivial detection caused by the same distribution from the same generator, as observed in previous AI-generated image detection datasets~\citep{CNNDet,GenImage}, we conduct two tasks to verify the performance of detection models: cross-generator generalization and degraded video classification. 
Cross-generator generalization refers to the model being trained on data generated by some generators and validated on unseen data generated by other generators, which is meant to test the model's generalization ability. Degraded video classification, on the other hand, is used to validate the model's robustness by testing its ability to recognize videos of different types of degradation.

\subsection{Organization of GenVideo} \label{sec:structure}
The GenVideo dataset primarily consists of real videos and fake videos shown in \cref{tab:dataset}. The real videos are mainly sourced from existing datasets related to video action dataset~\citep{K400} and video description dataset~\citep{Youku-mPLUG,xu2016msr}. The fake videos are obtained through external web scraping, internal generation pipelines based on open-source projects, and a number of existing video evaluation datasets~\citep{liu2023evalcrafter}.

Considering the emergence of video generation models, which primarily focus on diffusion-based methods~\citep{xing2023dynamicrafter,zhang2023pia,I2vgen-xl,blattmann2023stable,chen2024videocrafter2,xing2023dynamicrafter,zhang2023pia,chen2023seine} and methods based on auto-regressive models~\cite{ma2024latte,OpenSora}, the training set of the GenVideo dataset predominantly comprises videos generated by these two popular types of algorithms shown in \cref{tab:dataset}. Additionally, following~\citep{liu2023evalcrafter}, we generate $98,377$ videos using the service provided by Pika website~\cite{pika}. To balance the quantity ratio between real videos and fake videos, we sampled 260,232 and 953,279 video clips from the existing video datasets Kinetics-400~\citep{kay2017kinetics} and Youku-mPLUG~\citep{Youku-mPLUG}, respectively, to form the white sample of the training set.

For the testing set, the real videos are sourced from the MSR-VTT dataset~\citep{xu2016msr}, which is a large video description dataset. The fake videos are mainly sourced from two parts: the first part comes from the Evalcrafter benchmark~\citep{liu2023evalcrafter}, which is used to assess the temporal smoothness, quality, and other metrics of different generation models. The second part of the data comes from external web scraping, covering generated videos from existing popular video generation methods~\citep{zhang2024moonshot,yang2024direct,bar2024lumiere,ma2024magic,wang2024videocomposer,ren2024consisti2v,wang2023recipe,qing2023hierarchical,ho2022imagen,ge2023preserve,guo2023i2v,tian2024emo,kondratyuk2023videopoet,yoo2023towards,ni2023conditional,zeng2023make,wei2023dreamvideo,feng2023dreamoving,xu2023magicanimate,hu2023animate,yu2023magvit}, which includes diffusion-based methods~\citep{SVD,xing2023dynamicrafter,wang2023modelscope,zeng2023make,wei2023dreamvideo,feng2023dreamoving,xu2023magicanimate,hu2023animate}, auto-regressive-based models~\citep{kondratyuk2023videopoet,ma2024latte,sora}, and other models~\citep{yoo2023towards,yu2023magvit}.
This data encompasses most of the currently available video generation methods and advanced derivative methods of mainstream video generation techniques. Those scraped data are denoted as WildScape in \cref{tab:dataset}.

\subsection{Video collection details of GenVideo} \label{sec:collection}
We synthesize fake videos and gather real videos to construct the GenVideo dataset utilizing the hierarchical structure and the corresponding generators. 
It's crucial to underline that the primary objective of an AI-generated video detection dataset is to achieve \textit{robust and generalizable detection capabilities}, rather than solely focusing on \textit{video quality for assessment purposes}. 
A diverse and large-scale collection ensures that the dataset encompasses a wide range of video categories, facilitating detailed evaluations of AI-generated video detection algorithms and their effectiveness across various contexts.

\noindent\textbf{Real video collection:}
Considering that fake videos from generators are limited to specific domains determined by training datasets such as Kinetics-400~\citep{kay2017kinetics} and Youku-mPLUG~\citep{xu2023youku}, we sample parts of videos from those datasets as the real part of the GenVideo dataset. Specifically, we randomly sample $953,279$ videos from Youku-mPLUG~\citep{xu2023youku} and randomly slice 10-second segments from each video to form real samples.

\noindent\textbf{Fake video collection:}
The guiding principle for collecting fake videos is to ensure maximal diversity in content and generators. We prioritize generating additional fake videos using the most recent generators due to their superior quality.
To collect diverse fake videos from different resources as training samples, we have established a video generation pipeline for text-to-video generation and image-to-video generation. This pipeline facilitates the production of videos using popular generative mechanisms, including diffusion-based models~\citep{xing2023dynamicrafter,zhang2023pia,guo2023animatediff,I2vgen-xl,blattmann2023stable,chen2024videocrafter2,xing2023dynamicrafter,zhang2023pia,chen2023seine}, transformer-based methods~\citep{ma2024latte,OpenSora} and service-based method Pika website~\citep{pika}. 
For image-to-video generation, we employed various text-to-image models to produce diverse images, including different versions of Stable Diffusion (SD~\citep{rombach2022high}, SDXL~\citep{podell2023sdxl}). 
In order to produce videos with rich semantics through different generative approaches, including semantic-diverse text and image prompts across person, objects and diverse scenes, we firstly construct a rich prompt dictionary. In detail, we selected common 100 categories such as humans, animals,  plants and etc as foreground keywords, and typical 20 scenes like ``in the park'' or ``on the lawn'' as background keywords. Leveraging a large language model~\citep{le2023bloom}, these foreground and background keywords are expanded into about 4000 comprehensive textual prompts. Besides, we also randomly sample 1600 textual prompts from VBench~\cite{huang2024vbench} with consideration of semantic diversity and style diversity.
Secondly, each prompt from constructed prompt dictionary is fed into T2V (\emph{resp.,} T2I) generators to produce diverse videos (\emph{resp.,} images). I2V generators leverage images generated from T2I generators to produce videos.

To assemble a representative testing set, we investigated current video generators based on different model architectures~\citep{SVD,sora,ghosh2024raven} and scraped example videos from their projects, as illustrated by WildScape in \cref{tab:dataset}. This includes prominent video generation models such as VideoPoet~\citep{kondratyuk2023videopoet}, Emu~\citep{girdhar2023emu}, and Sora~\citep{sora}. Additionally, we collected videos generated by various condition-guided models~\citep{wei2023dreamvideo,feng2023dreamoving,xu2023magicanimate} that focus on social contexts and characters. We also included some non-mainstream generation algorithms, such as those based on latent flow diffusion models~\citep{ni2023conditional}, masked generative video transformer~\citep{yu2023magvit}, or autoregressive models~\citep{gupta2023photorealistic}. 
This approach ensures coverage of both popular algorithms and those generating high-quality content, particularly around character-centric videos.
We integrated existing video quality evaluation dataset~\cite{Evalcrafter} that include typical generation methods and demonstrate relatively high generation quality.

\section{DeMamba}
\begin{figure}[t]
	\centering
	\includegraphics[width=0.95\textwidth]{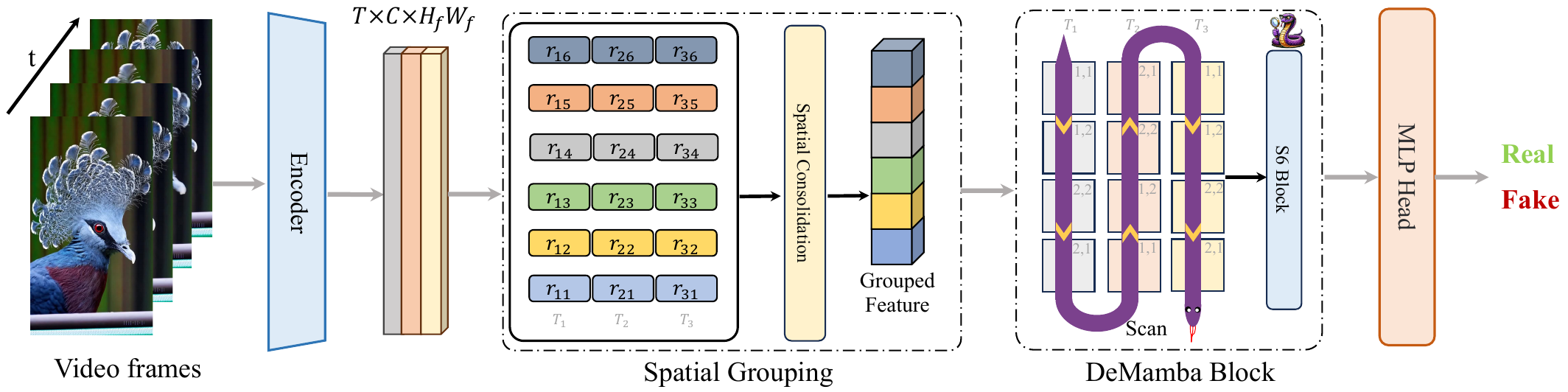}
  \caption{The overall framework of our Detail Mamba.}
	\label{fig:DeMamba_framework}
\end{figure}

\subsection{Preliminaries}
Structured State Space Sequence models (S4)~\citep{Gu_ICLR,rcc_lssl,SSSL} are grounded in continuous systems, facilitating the mapping of a one-dimension function or sequence, denoted as $x(t) \in \mathbb{R}^L$ to $y(t) \in \mathbb{R}^L$, via an intermediary hidden state $h(t) \in \mathbb{R}^N$. In a formal context, S4 leverage the subsequent ordinary differential equation to represent the input data:
\begin{gather}
    h'(t) = \mathbf{A}h(t) + \mathbf{B}x(t), \quad
    y(t) = \mathbf{C}h(t),
\end{gather}
where $\mathbf{A} \in \mathbb{R}^{N \times N}$ embodies the system's evolutionary matrix, with $\mathbf{B} \in \mathbb{R}^{N \times L}$ and $\mathbf{C} \in \mathbb{R}^{L \times N}$ serving as the projection matrices. To navigate the transition from continuous to discrete modeling in contemporary S4, the Mamba framework utilizes a timescale parameter $\mathbf{\bigtriangleup}$, facilitating the conversion of $\mathbf{A}$ and $\mathbf{B}$ into their discrete equivalents $\mathbf{\bar{A}}$ and $\mathbf{\bar{B}}$ through the Zero-Order Hold methodology~\citep{Gu_ICLR}, expressed as:
\begin{gather}
    \mathbf{\bar{A}} = {\rm exp}(\mathbf{\bigtriangleup A}),\quad
    \mathbf{\bar{B}} = \mathbf{\bigtriangleup A}^{-1}({\rm exp}(\mathbf{\bigtriangleup A}-\mathbf{I}))\cdot \mathbf{\bigtriangleup B},\quad
    h_t = \mathbf{\bar{A}}h_{t-1} + \mathbf{\bar{B}}x_t,\quad
    y_t = \mathbf{C}h_t.
\end{gather}
Contrary to traditional models that primarily rely on linear time-invariant S4, Mamba \citep{gu2023mamba} distinguishes itself by implementing a Selection mechanism computed with Scan for S4 (S6). 
Within the S6 framework, parameters $\mathbf{B} \in \mathbb{R}^{B \times L \times N}$, $\mathbf{C} \in \mathbb{R}^{B \times L \times N}$, and $\mathbf{\bigtriangleup} \in \mathbb{R}^{B \times L \times D}$ are inherently derived from the input $\mathbf{x} \in \mathbb{R}^{B \times L \times D}$, formulating an intrinsic structure for contextual perceptiveness and adaptive modulation of weights.

\subsection{AI-generated video detection with DeMamba module}
\textbf{Overview.} As illustrated in Figure~\ref{fig:DeMamba_framework}, our proposed method comprises a feature encoder, a DeMamba block, and an MLP classification head. Specifically, we employ state-of-the-art vision encoders (\textit{e.g.}, CLIP~\citep{CLIP} and XCLIP~\citep{XCLIP}) to encode the input video frames $\bm{X}^v \in \mathbb{R}^{3 \times T \times H \times W}$ into a sequence of features, denoted by $\bm{F} \in \mathbb{R}^{T \times C \times H_f \times W_f}$, where $C$ symbolizes the channel dimensionality, and $H_f$, $W_f$ represent the spatial dimensions, \textit{i.e.}, height and width of the feature maps, respectively. Following this, the extracted features are spatially grouped, and the DeMamba module is applied to model the intra-group feature consistency. Finally, we aggregate the features from different groups to determine whether the input video is generated by AI.

\textbf{DeMamba block.} We first apply spatial consolidation: given the feature $\bm{F}$, we split it into $s^2$ zones along both the height and width dimension where each zone of $\bm{F}$ is denoted as $\bm{F}_{jk}\in \mathbb{R}^{T\times C\times (H_f/s)\times (W_f/s)}$, where $j,k = \{1,...,s\}$. In Figure \ref{fig:DeMamba_framework}, we adapt the 1D Mamba layer for handling spatial-temporal input by expanding its capability to a 3D scan. 
In the previous Mamba approaches~\citep{gu2023mamba,VisionMamba,VMamba}, a sweep-scan mechanism was utilized, which might not effectively capture the inherent contextual relationships between adjacent tokens. 
To address this limitation, we propose a continuous scan strategy for each segmented region, aimed at maintaining spatial continuity throughout the entire scanning phase. 
Suppose a zone consists of four spatial positions: (1,1), (1,2), (2,1), and (2,2), corresponding to the top-left, top-right, bottom-left, and bottom-right corners, respectively. The sweep-scan order is (1,1) -> (1,2) -> (2,1) -> (2,2), whereas in the continuous scan, the order is (1,1) -> (1,2) -> (2,2) -> (2,1).
This method organizes spatial tokens based on their proximity and subsequently aligns them sequentially across successive frames.
It facilitates the coherent integration of spatial and temporal dynamics, enhancing the capability of the model to capture complex spatial-temporal relationships. After modeling the spatial-temporal inconsistency of each partitioned region using DeMamba, we can obtain the feature $\bm{F}'_{jk}\in \mathbb{R}^{T\times C\times (F_hF_w/s^2))}$, where $j,k =  \{1,...,s\}$.

\textbf{Classification head.} 
To leverage more comprehensive features for classification, we aggregate both global and local features. Specifically, we temporally and spatially average the input features $\bm{F}$ before the DeMamba block to obtain the global feature $\bm{F}^{\rm global}\in \mathbb{R}^{C}$, and average pool the temporal and spatial features $\bm{F}'_{jk}$ after the DeMamba processing into pooled features $\bm{F}^{\rm pool}_{jk} \in \mathbb{R}^{C}$. Then we concatenate the local features with the global features and apply a simple MLP for classification:
\begin{equation}
    y_{\rm pred} = {\rm Sigmoid}({\rm MLP}([\bm{F}^{\rm global};\bm{F}^{\rm pool}_{11};....;\bm{F}^{\rm pool}_{ss}])).
\end{equation}
Finally, we use binary cross-entropy loss to train our model to classify real/fake videos.


\section{Experiments}

\subsection{Implementation details}
\textbf{Datasets.}
To comprehensively analyze the performance of various detectors, we divided the dataset into two distinct parts: the basic training set $D_{\rm train}$ and the out-of-domain test set $D_{\rm v-ood}$. $D_{\rm train}$ and $D_{\rm v-ood}$ contain fake videos created by different generative methods and different real videos. $D_{\rm train}$ and $D_{\rm v-ood}$ contain fake videos produced by different generative methods and real videos from different sources. $D_{\rm train}$ includes 1,213,511 real videos and 1,048,575 generated videos produced using 10 baseline generative methods. $D_{\rm v-ood}$ contains 10,000 real videos and 8,588 generated videos created with 10 generative methods. 
For detailed information about the data, please refer to \textbf{\cref{sec:dataset} and \cref{sec:collection}}.

\begin{table*}[h]
    \centering
    \caption{Training parameter settings.}
    \resizebox{0.7\textwidth}{!}{
    \begin{tabular}{cccccccc}
    \toprule
        \multirow{2}{*}{Model} & \multirow{2}{*}{Batchsize (Video level)} & \multirow{2}{*}{LR}   & \multirow{2}{*}{Epochs} & \multirow{2}{*}{Optimizer} & \multirow{2}{*}{Scheduler} & \multirow{2}{*}{Loss}\\
        & & & \\
        \midrule
        F3Net & 128 & 1e-5 &  30& \multirow{15}{*}{AdamW} & [20, 25], lr * 0.1 & \multirow{15}{*}{BCE}\\
        NPR & 128 & 1e-5&  30& & [20, 25], lr * 0.1 &\\
        STIL & 128 & 1e-5& 30& & [20, 25], lr * 0.1 &\\
        VideoMAE-B & 128 & 1e-5& 30& & [20, 25], lr * 0.1 &\\
        CLIP-B-LP & 128 & 1e-6& 30& & [20, 25], lr * 0.1 &\\
        DeMamba-CLIP-LP & 128 & 1e-6& 30& & [20, 25], lr * 0.1 &\\
        XCLIP-B-LP & 128 & 1e-6& 30& & [20, 25], lr * 0.1 &\\
        DeMamba-XCLIP-LP & 128 & 1e-6& 30& & [20, 25], lr * 0.1 &\\
        CLIP-B-FT & 128 & 1e-6& 10 & & - &\\
        MINTIME-CLIP-B & 128 & 1e-6& 10& & - &\\
        FTCN-CLIP-B & 128 & 1e-6& 10& & - &\\
        TALL & 128 & 1e-6& 10& & - &\\
        XCLIP-B-FT & 128 & 1e-6& 10&& - &\\
        DeMamba-CLIP-FT & 128 & 1e-6& 10 &&  - &\\
        DeMamba-XCLIP-FT & 128 & 1e-6& 10 && - &\\
         \bottomrule
    \end{tabular}}
    \label{param_set}
\end{table*}

\begin{table*}[t]
    \centering
    \caption{Comparisons to the SOTAs in F1 score (F1) and average precision (AP) on the many-to-many generalization task.}
    \label{tab_m2m}
    \resizebox{\textwidth}{!}{
    \begin{tabular}{cccccccccccccc}
    \toprule
       \multirow{2}{*}{Model} & Detection&\multirow{2}{*}{Metric}&\multirow{2}{*}{Sora}&Morph&\multirow{2}{*}{Gen2}&\multirow{2}{*}{HotShot}&\multirow{2}{*}{Lavie}&\multirow{2}{*}{Show-1}&Moon&\multirow{2}{*}{Crafter}&Model&Wild&\multirow{2}{*}{Avg.}\\
       &level&&&Studio&&&&&Valley&&Scope&Scrape\\
       \midrule
       \multirow{3}{*}{F3Net}&\multirow{3}{*}{Image}&
       R& 
       0.8393& 0.9971 &0.9862
       &0.7757&0.5700 &0.3657 &0.9952&0.9971
       & 0.8943& 0.7678&0.8188 
       \\
       &&F1&0.5000&0.9406&0.9628&0.8169&0.6988&0.4904&
0.9332&0.9688&0.8873&0.8251&0.8024\\
       &&AP& 0.6827& 0.9989& 0.9967&0.8935
       &0.8524& 0.6317& 0.9958&0.9989& 0.9380
       &0.8841& 0.8873
       \\
        \midrule
       \multirow{3}{*}{NPR}&\multirow{3}{*}{Image}&
R& 0.9107 &0.9957& 0.9949& 0.2429& 0.8964& 0.5771 &0.9712
       &0.9986&0.9429&0.8780&0.8408\\
       &&F1&0.2786&0.8441&0.9131&0.3028&0.8627&0.5944&
0.8170&0.9164&0.8184&0.8163&0.7164
       \\
       
       &&AP&0.6717 &0.9914& 0.9920& 0.2276& 0.9391& 0.6176&0.9633&0.9972& 0.9415& 0.9040& 0.8245
       
       \\
       \midrule
       \multirow{3}{*}{STIL}&\multirow{3}{*}{Video}
        &R
       & 0.7857 & 0.9814 & 0.9804 & 0.7600 & 0.6179 & 0.5329 & 0.9936 & 0.9736 & 0.9457 & 0.6501 & 0.8222
       \\
       &&F1&0.3805&0.9068&0.9458&0.7824&0.7232&0.6217&0.9039&0.9433&0.8884&0.7267&0.7823\\
       &&AP& 0.5721 & 0.9908 & 0.9932 & 0.8619 & 0.8224 & 0.7043 & 0.9925 & 0.9896 & 0.9718 & 0.8132& 0.8712
       \\
       \midrule
       \multirow{3}{*}{VideoMAE-B}&\multirow{3}{*}{Video}&
        
        R&
       0.6786 & 0.9600 &0.9841 & 0.9614& 0.7714 & 0.8043 &0.9744&0.9693& 0.9629
       &0.6836&0.8750\\
        
        &&F1&0.623&0.9593&0.9819&0.9600&0.8608&0.8722&0.9644&0.9745&0.9615&0.7977&0.8955
       \\
       &&AP&0.6649 & 0.9885 & 0.9977 & 0.9927& 0.9655 & 0.9531 & 0.9949&
       0.9969&0.9927 & 0.9074& 0.9454
       \\
       \midrule       
\multirow{3}{*}{MINTIME-CLIP-B}&\multirow{3}{*}{Video}&
R& 0.8929&1.0000 &0.9899 &0.2643 &0.9679 &0.9814 &0.9984 &  1.0000&0.8429 &0.8238 &0.8762
\\
&&F1&0.4902 &0.9340 & 0.9606 &0.3760 &0.9499 &0.9246 &0.9266 & 0.9659& 0.8495&0.8539 &0.8231\\
&&AP&0.8321 &0.9999 &0.9967 &0.5084 &0.9920 &0.9927 &0.9976 &0.9999 & 0.9183&0.9177 & 0.9155\\
\midrule
       \multirow{3}{*}{FTCN-CLIP-B}&\multirow{3}{*}{Video}&
R&0.8750 & 1.0000 & 0.9891 & 0.1771 & 0.9771 & 0.9186 & 1.0000 & 1.0000 &  0.8529& 0.8283& 0.8618 \\
&&F1& 0.7840 & 0.9859 & 0.9873 & 0.2922 & 0.9813 & 0.9442 & 0.9843 & 0.9929 & 0.9073 & 0.8960 & 0.8755\\
&&AP&0.9179 & 0.9999 & 0.9979 & 0.4594 & 0.9976 & 0.9780 & 0.9999 & 0.9999 & 0.9469& 0.9232&0.9221\\
\midrule

       \multirow{3}{*}{TALL}&\multirow{3}{*}{Video}&
R&0.9107& 0.9828& 0.9783& 0.8300& 0.7657& 0.7957&0.9952 & 0.9893 &0.9414 &0.6631 & 0.8852 \\
&&F1&0.2615& 0.8240& 0.8964& 0.7430&  0.7782 &0.7234& 0.8133 &0.9032 & 0.8027 & 0.6736 &0.7419   \\
&&AP&0.7115& 0.9689& 0.9851& 0.7938&0.8459 &0.7938&0.9879 &0.9902 &0.9270 &0.7647 &0.8791\\
       \midrule
       \multirow{3}{*}{CLIP-B-PT}&\multirow{3}{*}{Image}&
       R&
       0.8571&0.8243&0.9036&0.7100&0.7929&
    0.7543&0.8962&0.8629&0.8214&0.7516&0.8174\\
    &&F1& 0.0215&0.2078&0.1816&0.3271&0.1918&0.2053&0.3509&0.2068&0.2357&0.2289&0.2157
       \\
       &&AP&0.6780&0.4356&0.7088&0.2997 &0.5297
       &0.3536 &0.5552&0.6603&0.4423&0.4299
       &0.4483
       \\
       \midrule
       \multirow{3}{*}{DeMamba-CLIP-PT}&\multirow{3}{*}{Video}&
       R& 0.5893&0.9643&0.9312
       &0.6800& 0.6936& 0.6900& 0.8914& 0.9186& 0.9614& 0.5659& $0.7886_{\textcolor[RGB]{0,116,0}{-0.0308}}$\\
       &&F1&
       0.234&0.8610&0.8992&0.6949&0.7574&0.7015&0.8105&0.8934&0.8588&0.6379&$0.7349_{\textcolor[RGB]{116,0,0}{+0.5060}}$
       \\
       
       &&AP& 0.2587&0.9514& 0.9623
       &0.7343& 0.8331& 0.7549& 0.9017& 0.9506& 0.9505& 0.6995&$0.7997_{\textcolor[RGB]{116,0,0}{+0.3514}}$
       
       \\\midrule
       \multirow{3}{*}{CLIP-B-FT}&\multirow{3}{*}{Image}&
 R&
       0.9464 & 0.9986& 0.9138& 0.7729& 0.8814& 0.8600& 0.9968& 0.9979& 0.8429&0.8467&0.9057\\
       &&F1&0.2818&0.8422&0.8691&0.7204&0.8521&0.7698&0.8252&0.9137&0.7608&0.7955&0.7631
       \\
       &&AP&0.8067 & 0.9967& 0.9524& 0.8220& 0.9348& 0.8862& 0.9955& 0.9979& 0.8693&0.8908&0.9152
       \\
       \midrule
        \multirow{3}{*}{DeMamba-CLIP-B-FT}&\multirow{3}{*}{Video}&R& 0.9571 & 1.0000 & 0.9870 & 0.6914 & 0.9243 & 0.9329 & 1.0000 & 1.0000 & 0.8357 & 0.8294 &$0.9158_{\textcolor[RGB]{116,0,0}{+0.0101}}$\\
        
        &&F1& 0.6463&0.9615&0.9739&0.7803&0.9414&0.9276&0.9572&0.9804&0.8723&0.8782&$0.8919_{\textcolor[RGB]{116,0,0}{+0.1288}}$
       \\
       &&AP& 0.8550 & 1.0000 & 0.9959 & 0.7615 & 0.9678 & 0.9699 & 0.9997 & 1.0000 & 0.8980 & 0.8972 & $0.9345_{\textcolor[RGB]{116,0,0}{+0.1930}}$
       \\
        \midrule
       \multirow{3}{*}{XCLIP-B-PT}&\multirow{3}{*}{Video}&
       R
       &0.8134 & 0.8215& 0.8335&0.8098& 0.8182&0.8155& 0.8214& 0.8298& 0.8193& 0.8110&0.8193\\
       &&F1& 0.0458&0.3666&0.5237&0.3619&0.5196&0.3640&0.3424&0.5250&0.3656&0.4240&0.3839
       \\
        &&AP&0.1639& 0.7216& 0.8777& 0.3986& 0.6557& 0.5426& 0.7523& 0.8480& 0.6160& 0.5528&0.6129 
       \\
       \midrule
       \multirow{3}{*}{DeMamba-XCLIP-PT}&\multirow{3}{*}{Video}&R
       &0.6607&0.9586&0.9464& 0.7786& 0.7536& 0.8029& 0.9089& 0.9250& 0.9600& 0.6641&$0.8359_{\textcolor[RGB]{116,0,0}{+0.0166}}$\\
        &&F1&0.1244&0.7223&0.8232&0.6293&0.7172&0.6427&0.6758&0.8137&0.7226&0.6058&$0.6477_{\textcolor[RGB]{116,0,0}{+0.2638}}$
       \\
       &&AP&0.1826& 0.9350& 0.9472& 0.6994& 0.7808& 0.7150& 0.8395& 0.9223& 0.9354& 0.6810&$0.7638_{\textcolor[RGB]{116,0,0}{+0.1509}}$
       \\
       \midrule

       \multirow{3}{*}{XCLIP-B-FT}&\multirow{3}{*}{Video}&
        R&
       0.8214&0.9957 &0.9362 & 0.6129 & 0.7936&0.6971 &0.9792&0.9979& 0.7714 & 0.8359&0.8441\\
       &&F1&0.3147&0.8805&0.9041&0.6530&0.8242&0.7099&0.8602&0.9370&0.7570&0.8212&0.7662\\
       &&AP&0.6442& 0.9973&0.9678& 0.7098 & 0.9035& 0.7728&0.9734&0.9984
       &0.8201& 0.8897&0.8677\\
       \midrule
       \multirow{3}{*}{DeMamba-XCLIP-FT}&\multirow{3}{*}{Video}&
        R
       &0.9821& 1.0000 & 0.9986 & 0.6543 & 0.9486 & 0.9886 & 1.0000& 1.0000&0.9286 &0.8909 & $\textbf{0.9392}_{\textcolor[RGB]{116,0,0}{+0.0951}}$\\
       &&F1&0.6467& 0.9602& 0.9790& 0.7539& 0.9537&0.9551&0.9557&0.9797&0.9240&0.9120&$\textbf{0.9020}_{\textcolor[RGB]{116,0,0}{+0.1358}}$
        \\
        &&AP&0.9332 & 1.0000 & 0.9997 & 0.8555 & 0.9897 & 0.9960& 0.9998& 1.0000& 0.9777& 0.9575&$\textbf{0.9710}_{\textcolor[RGB]{116,0,0}{+0.1043}}$
       \\
        \bottomrule
    \end{tabular}}
    \label{main_results}
\end{table*}

\begin{table*}[t]
    \centering
    \caption{Comparisons to the SOTAs in mean accuracy (ACC) and average precision (AP) on the one-to-many generalization task.}
    \label{12many}
    \resizebox{\textwidth}{!}{
    \begin{tabular}{ccccccccccccccc}
    \toprule
    \multirow{2}{*}{Training}&\multirow{3}{*}{Model}&\multirow{2}{*}{Detection}&\multirow{3}{*}{Metric}&\multicolumn{10}{c}{Testing subset}&\multirow{3}{*}{Avg.}
    \\\cline{5-14}
    &&&&\multirow{2}{*}{Sora}&Morph&\multirow{2}{*}{Gen2}&\multirow{2}{*}{HotShot}&\multirow{2}{*}{Lavie}&\multirow{2}{*}{Show-1}&Moon&\multirow{2}{*}{Crafter}&Model&Wild&\\
    subset&&level&&&Studio&&&&&Valley&&Scope&Scrape\\
    \midrule

    \multirow{21}{*}{Pika}
    &\multirow{3}{*}{NPR}&\multirow{3}{*}{Image}&
    R& 0.5536 & 0.7757& 0.7188&0.4860& 0.0721& 0.0429& 0.8626& 0.6029
    &0.7143& 0.3153& 0.5144\\
    &&&F1&0.4627&0.8410&0.8197&0.0871&0.1293&0.0772&
0.8894&0.7368&0.8019&0.4604&0.5306
    \\
    &&&AP&0.4574 & 0.9155& 0.9271&0.2180& 0.4432 & 0.2274& 0.9504& 0.9003& 0.8491& 0.6088& 0.6497 
    \\
       
       \cmidrule(lr){2-15}
    &\multirow{3}{*}{STIL}&\multirow{3}{*}{Image}&
    R& 0.7500
     & 0.7943 & 0.9449 & 0.5786 & 0.5314 & 0.6414 & 0.9712 & 0.8529 & 0.6943 & 0.6242 &0.7383 \\
    &&&F1& 0.0984&0.5530&0.7580&0.4355&0.5128&0.4708&0.6107&0.7130&0.5008&0.5117&0.5165
    \\
       &&&AP&0.2235 & 0.7162 & 0.9319 & 0.4061 & 0.5324 & 0.4773 &  0.9494 & 0.8582 & 0.5899 & 0.6191 & 0.6304
       \\
    \cmidrule(lr){2-15}
    &\multirow{3}{*}{MINTIME-CLIP-B}&\multirow{3}{*}{Video}&
    R& 0.1607&0.2686&0.1471&0.0557&0.3757&0.0900&0.2188&0.4071&0.0114&0.4309 & 0.2166\\
    &&&F1& 0.2338&0.4178&0.2545&0.1039&0.5428&0.1626&0.3535&0.5752&0.0222&0.5969 & 
0.3263
\\
    &&&AP& 0.1692&0.7433&0.6066&0.3903&0.8229&0.4108&0.7479&0.8862&0.2385&0.7225 & 0.5738\\
    \cmidrule(lr){2-15}
    &\multirow{3}{*}{FTCN-CLIP-B}&\multirow{3}{*}{Video}&
    R&0.9107&0.9300&0.9232&0.1771&0.1750&0.0643&0.9968&0.8479&0.7457&0.5173 & 0.6287\\
    &&&F1&0.5896&0.9195&0.9371&0.2770&0.2855&0.1088&0.9483&0.8948&0.8106&0.6513 & 0.6423\\
    &&&AP&0.8369&0.9705&0.9718&0.3747&0.4490&0.1733&0.9975&0.9541&0.7971&0.6686 & 0.7194\\
    \cmidrule(lr){2-15}
    &\multirow{3}{*}{TALL}&\multirow{3}{*}{Video}&
    R& 0.6071& 0.7043 &0.9449  & 0.5914 &0.5757  &0.6400  &0.9633  & 0.8871 & 0.6057  &0.6210&0.7141 \\
    &&&F1&0.1038& 0.5609  &0.8029  &0.4932  &0.5817  &0.5231  & 0.6726  &  0.7746&0.5024  &0.5569& 0.5572\\
    &&&AP&0.1582& 0.6363   &0.9347  &0.4400  &0.5907  &0.5106  &  0.9209 &  0.8785&0.5111&0.6443&0.6225 \\
           \cmidrule(lr){2-15}
    &\multirow{3}{*}{XCLIP-B-FT}&\multirow{3}{*}{Video}&
        R& 0.6786 & 0.9129  & 0.9623 & 0.1200 & 0.2236 & 0.0914 & 0.9984 & 0.8343 & 0.7557 & 0.5184 &0.6096 \\
    &&&F1&0.5891&0.9301&0.9679&0.2049&0.3581&0.1579&0.9712&0.8974&0.8361&0.6662&0.6579
    \\
    &&&AP& 0.7108 & 0.9753 & 0.9944 & 0.4468 & 0.7269 &  0.3837 & 0.9996 & 0.9732 & 0.8800 & 0.7400 &0.7831
    \\

    \cmidrule(lr){2-15}
    &\multirow{3}{*}{DeMamba-XCLIP-FT}&\multirow{3}{*}{Video}&R &0.9286 & 0.9729 & 0.9848 & 0.3829 & 0.5350 & 0.4143 & 0.9984 & 0.9407 & 0.7729 & 0.6415 &$\textbf{0.7572}_{\textcolor[RGB]{116,0,0}{+0.1476}}$ \\
    &&&F1&
    0.4298&0.8990&0.9460&0.4864&0.6559&0.5160&0.9017&0.9235&0.7869&0.7183&
$\textbf{0.7263}_{\textcolor[RGB]{116,0,0}{+0.0684}}$
    \\
    &&&AP&0.7775 & 0.9842 & 0.9916 & 0.5297 & 0.7672 &0.5624 & 0.9980 & 0.9791 & 0.8283 & 0.7481 &  $\textbf{0.8166}_{\textcolor[RGB]{116,0,0}{+0.0335}}$
    
    \\
       \midrule
       
    \multirow{21}{*}{SEINE}
        &\multirow{3}{*}{NPR}&\multirow{3}{*}{Image}&
        R& 0.4643& 0.7857&0.6370& 0.2186
    & 0.0700& 0.0329& 0.9297& 0.8929&0.3386& 0.2484&0.4618
    \\&&&F1& 0.4685&0.8592&0.7684&0.3469&0.1283&0.0612&0.9401&0.9332&0.4907&0.3882&0.5385
    \\
    &&&AP&0.3630& 0.9263& 0.8502& 0.5268& 0.2569& 0.1105& 0.9780& 0.9778&0.6464&0.4748& 0.6111
    \\

       \cmidrule(lr){2-15}
    &\multirow{3}{*}{STIL}&\multirow{3}{*}{Video}&R& 0.7143 & 0.8043 & 0.8848 & 0.6771 & 0.5457 & 0.5571 & 0.9393 & 0.8957 & 0.7200 & 0.5011 &0.7239\\
    &&&F1 &0.0922&0.5533&0.7240&0.4864&0.5199&0.4181&0.5921&0.7317&0.5099&0.4292&0.5057
    \\
    &&&AP&0.2389 & 0.7101 & 0.8818 & 0.5217 & 0.5449 & 0.4123 & 0.8473 & 0.8738 & 0.5872 & 0.4651 & 0.6083
    \\
    \cmidrule(lr){2-15}
    &\multirow{3}{*}{MINTIME-CLIP-B}&\multirow{3}{*}{Video}&
    R&0.6964&0.9257&0.9116&0.1071&0.2293&0.1071&0.9968&0.8600&0.7371&0.5637 & 0.6135\\
    &&&F1&0.5532&0.9297&0.9378&0.1827&0.3633&0.1827&0.9630&0.9087&0.8177&0.6988 & 0.6538\\
    &&&AP&0.6596&0.9781&0.9790&0.3688&0.6215&0.3311&0.9978&0.9701&0.7963&0.7402  & 0.7443\\
    \cmidrule(lr){2-15}
    &\multirow{3}{*}{FTCN-CLIP-B}&\multirow{3}{*}{Video}&
    R&0.8929&0.9786&0.6297&0.4043&0.1650&0.1557&0.9984&0.9971&0.6514&0.4060&0.6279

 \\
    &&&F1&0.8130&0.9765&0.7664&0.5660&0.2802&0.2636&0.9858&0.9922&0.7775&0.5697&0.6991 \\
    &&&AP&0.8836&0.9972&0.8736&0.8028&0.6072&0.4321&0.9998&0.9998&0.8139&0.6255&0.8036 \\
    \cmidrule(lr){2-15}
    &\multirow{3}{*}{TALL}&\multirow{3}{*}{Video}&
    R&0.5714&0.7643&0.7957&0.6443&0.3486 &0.5629 &0.8514&0.9143 &0.6714&0.4438&0.6568\\
    &&&F1&0.1730&0.7053&0.7957&0.6289&0.4498&0.5727&0.7403&0.86423&0.6469&0.5077& 0.6085\\
    &&&AP&0.2716&0.7836&0.8838&0.6896&0.5518&0.6093&0.8240&0.9426&0.7007&0.5483& 0.6805 \\
    \cmidrule(lr){2-15}
    &\multirow{3}{*}{XCLIP-B-FT}&\multirow{3}{*}{Video}&
    R& 0.8571 & 0.9543 & 0.7623 & 0.6586 & 0.3593 & 0.3700 & 0.9968 & 0.9900 & 0.7557 & 0.4978 & 0.7201\\
    &&&F1 &0.7705&0.9632&0.8580&0.7814&0.5234&0.5291&0.9826&0.9879&0.8468&0.6548&0.7898
    \\
    &&&AP& 0.8589 & 0.9797 & 0.9440 & 0.9281 & 0.8168 &  0.7768 & 0.9848 & 0.9891 & 0.9227 & 0.6791 & 0.8880
    
    \\

       \cmidrule(lr){2-15}
    &\multirow{3}{*}{DeMamba-XCLIP-FT}&\multirow{3}{*}{Video}&R & 0.9464 & 0.9843 & 0.9217 &  0.8243 & 0.5229 & 0.5400 & 0.9952 & 0.9914 & 0.7929 &  0.5788 &$\textbf{0.8098}_{\textcolor[RGB]{116,0,0}{+0.0897}}$\\
   &&&F1& 0.5000&0.9255&0.9240&0.8381&0.6559&0.6412&0.9228&0.9609&0.8192&0.6855&
$\textbf{0.7873}_{\textcolor[RGB]{0,116,0}{-0.0025}}$
    \\
    &&&AP&0.8374 & 0.9901 & 0.9766 & 0.9082 & 0.8411 & 0.7330 & 0.9972 & 0.9973 & 0.8972 & 0.7645 & $\textbf{0.8943}_{\textcolor[RGB]{116,0,0}{+0.0063}}$
    
    \\
       \midrule
       
    \multirow{21}{*}{OpenSora}
        &\multirow{3}{*}{NPR}&\multirow{3}{*}{Image}&
    R
    &0.5536& 0.7629& 0.5551&0.5857&0.7650&0.2243 &0.7492 &0.8307&0.2986 & 0.6037& 0.5929\\
    &&&F1& 0.1263&0.6520&0.6008&0.5406&0.7448&0.2490&0.6248&0.7835&0.3184&0.5918&0.5232  
    \\
    &&&AP&0.2565&0.7524& 0.6502&0.5512&0.8242&0.2075 &0.7265 &0.8684&0.2813& 0.6450& 0.5763
    \\

       \cmidrule(lr){2-15}
    &\multirow{3}{*}{STIL}&\multirow{3}{*}{Video}&
     R& 0.3214
     & 0.4543 & 0.5645 & 0.3514& 0.4507 & 0.3457 & 0.5783 & 0.6314 & 0.1986 & 0.4395 &0.4336 \\
    &&&F1 &0.1313&0.5282&0.6644&0.4332&0.5686&0.4277&
0.6167&0.7153&0.2712&0.5349&0.4892
    \\
    &&&AP& 0.0694 & 0.5562 & 0.7599 &0.4355& 0.6806 & 0.4401 & 0.6384 & 0.8059 & 0.2939 & 0.5758 &0.5256

    \\
    \cmidrule(lr){2-15}
    &\multirow{3}{*}{MINTIME-CLIP-B}&\multirow{3}{*}{Video}&
    R&0.8571&0.9214&0.7971&0.6343&0.2357&0.3057&0.9712&0.9636&0.7257&0.4773 & 0.6889\\
    &&&F1& 0.6000&0.9208&0.8675&0.7396&0.3695&0.4412&0.9426&0.9619&0.8044&0.6212 & 0.7269 \\
    &&&AP&0.7981&0.9739&0.9370&0.8451&0.6232&0.5819&0.9891&0.9926&0.8529&0.6925 & 0.8286\\
    \cmidrule(lr){2-15}
    &\multirow{3}{*}{FTCN-CLIP-B}&\multirow{3}{*}{Video}&
    R&0.1964&0.1786&0.1442&0.1957&0.1579&0.0329&0.2716&0.3229&0.0214&0.3413 & 0.1863 \\
    &&&F1& 0.3235&0.3027&0.2519&0.3270&0.2725&0.0635&0.4252&0.4879&0.0419&0.5084 & 0.3005\\
    &&&AP&0.3178&0.6675&0.4927&0.6954&0.7089&0.2382&0.7876&0.8500&0.2263&0.7049 & 0.5689\\
    \cmidrule(lr){2-15}
    &\multirow{3}{*}{TALL}&\multirow{3}{*}{Video}&
    R&0.3393&0.5771&0.5225&0.3986&0.5893&0.6129&0.5655&0.6736&0.2429&0.3952 & 0.4916\\
    &&&F1&0.1357&0.6173&0.6256&0.4713&0.6790&0.6432&0.5975&0.7405&0.3163&0.4890&0.5315 \\
    &&&AP& 0.0798&0.6447&0.7246&0.4924&0.7830&0.6916&0.6166&0.8329&0.3132&0.5313&0.5710 \\
    \cmidrule(lr){2-15}
    &\multirow{3}{*}{XCLIP-B-FT}&\multirow{3}{*}{Video}&
     R& 0.6786 & 0.7586 & 0.6746 & 0.7086 & 0.7314 & 0.4357 & 0.7987 & 0.8629  & 0.3343 & 0.6317 &0.6615 \\
    &&&F1& 0.2686&0.7479&0.7443&0.7162&0.7833&0.5096&0.7595&0.8638&0.4167&0.6875&0.6497
    \\
    &&&AP& 0.4828 & 0.8139 & 0.8184 & 0.7738 & 0.8608 &  0.5187 & 0.8341 & 0.9318 & 0.3927 & 0.7274 & 0.7154
    \\
    
       \cmidrule(lr){2-15}
    &\multirow{3}{*}{DeMamba-XCLIP-FT}&\multirow{3}{*}{Video}&R&
     0.5536& 0.8743 & 0.8130 & 0.7314 & 0.8521 & 0.7314 & 0.8962 &  0.9007& 0.4486 & 0.5810 & $\textbf{0.7382}_{\textcolor[RGB]{116,0,0}{+0.0767}}$\\
    &&&F1& 0.1742&
0.7742&0.8094&0.6905&0.8333&0.6905&0.7706&
0.8604&0.4895&0.6209&$\textbf{0.6713}_{\textcolor[RGB]{116,0,0}{+0.0216}}$
    \\
    &&&AP& 0.2589 & 0.8663 & 0.8727 & 0.7438 & 0.9112 & 0.7601 & 0.8641 & 0.9383 & 0.4874 &  0.6792 & $\textbf{0.7382}_{\textcolor[RGB]{116,0,0}{+0.0228}}$
    \\
    \bottomrule
    \end{tabular}}
    \label{main_results}
\end{table*}

\begin{table*}[t]
    \centering
    \caption{Robustness evaluation of different detectors on many-to-many generalization task.}
     \resizebox{\textwidth}{!}{
    \begin{tabular}{ccccccccccccc}
    \toprule
       \multirow{2}{*}{Model} & Detection&\multirow{2}{*}{Metric}&\multirow{2}{*}{Original}&\multicolumn{2}{c}{Compression}&\multicolumn{2}{c}{Transformation}&\multicolumn{2}{c}{Watemarks}&Gaussian&Color
       \\\cline{5-10}
       &level&&&${\rm CRF}=28$&JEPG&Flip&Crop&Text&Image&noise&transform\\
       \midrule
       \multirow{3}{*}{F3Net}&\multirow{3}{*}{Image}&
       R&0.8188 & $0.7721_{\textcolor[RGB]{0,116,0}{-0.0467}}$ & $0.8108_{\textcolor[RGB]{0,116,0}{-0.0080}}$ & $0.8041_{\textcolor[RGB]{0,116,0}{-0.0147}}$  & $0.6782_{\textcolor[RGB]{0,116,0}{-0.1406}}$ & $0.7513_{\textcolor[RGB]{0,116,0}{-0.0675}}$ & $0.7668_{\textcolor[RGB]{0,116,0}{-0.0520}}$ & $0.8025_{\textcolor[RGB]{0,116,0}{-0.0163}}$ & $0.8102_{\textcolor[RGB]{0,116,0}{-0.0086}}$\\
       &&F1&0.8024&
       $0.7967_{\textcolor[RGB]{0,116,0}{-0.0057}}$ &
       $0.6984_{\textcolor[RGB]{0,116,0}{-0.1040}}$ &
    $0.7945_{\textcolor[RGB]{0,116,0}{-0.0079}}$ &
    $0.5988_{\textcolor[RGB]{0,116,0}{-0.2036}}$ &
    $0.7858_{\textcolor[RGB]{0,116,0}{-0.0166}}$ &
    $0.7944_{\textcolor[RGB]{0,116,0}{-0.0080}}$ &
    $0.7910_{\textcolor[RGB]{0,116,0}{-0.0114}}$ &
    $0.7912_{\textcolor[RGB]{0,116,0}{-0.0112}}$ 

       \\
       &&AP&0.8873 & $0.8739_{\textcolor[RGB]{0,116,0}{-0.0134}}$ & $0.8102_{\textcolor[RGB]{0,116,0}{-0.0771}}$ & $0.8770_{\textcolor[RGB]{0,116,0}{-0.0173}}$ & $0.6705_{\textcolor[RGB]{0,116,0}{-0.2168}}$ & $0.8701_{\textcolor[RGB]{0,116,0}{-0.0172}}$ & $0.8793_{\textcolor[RGB]{0,116,0}{-0.0070}}$ & $0.8717_{\textcolor[RGB]{0,116,0}{-0.0156}}$ & $0.8717_{\textcolor[RGB]{0,116,0}{-0.0156}}$\\\midrule
       \multirow{3}{*}{NPR}&\multirow{3}{*}{Image}&
        R&0.8408 & $0.8012_{\textcolor[RGB]{0,116,0}{-0.0396}}$ & $0.7375_{\textcolor[RGB]{0,116,0}{-0.1033}}$ & $0.8365_{\textcolor[RGB]{0,116,0}{-0.0043}}$ & $0.7144_{\textcolor[RGB]{0,116,0}{-0.1264}}$ & $0.8285_{\textcolor[RGB]{0,116,0}{-0.0123}}$ & $0.8200_{\textcolor[RGB]{0,116,0}{-0.0208}}$ & $0.8392_{\textcolor[RGB]{0,116,0}{-0.0016}}$ & $0.8378_{\textcolor[RGB]{0,116,0}{-0.0030}}$\\
        &&F1 & 0.7164  & 
        $0.6382_{\textcolor[RGB]{0,116,0}{-0.0782}}$ &
        $0.4588_{\textcolor[RGB]{0,116,0}{-0.2576}}$ &
        $0.6896_{\textcolor[RGB]{0,116,0}{-0.0268}}$ &
        $0.1655_{\textcolor[RGB]{0,116,0}{-0.5509}}$ &
        $0.5580_{\textcolor[RGB]{0,116,0}{-0.1584}}$ &
        $0.6583_{\textcolor[RGB]{0,116,0}{-0.0581}}$ &
        $0.4791_{\textcolor[RGB]{0,116,0}{-0.2373}}$ &
        $0.6564_{\textcolor[RGB]{0,116,0}{-0.0600}}$ 
       \\
       &&AP& 0.8245 & $0.7612_{\textcolor[RGB]{0,116,0}{-0.0633}}$ & $0.5372_{\textcolor[RGB]{0,116,0}{-0.2873}}$ & $0.7905_{\textcolor[RGB]{0,116,0}{-0.0340}}$ & $0.3636_{\textcolor[RGB]{0,116,0}{-0.4609}}$ & $0.6716_{\textcolor[RGB]{0,116,0}{-0.1528}}$ & $0.7702_{\textcolor[RGB]{0,116,0}{-0.0543}}$ & $0.5755_{\textcolor[RGB]{0,116,0}{-0.2490}}$ & $0.7681_{\textcolor[RGB]{0,116,0}{-0.0564}}$\\\midrule
       \multirow{3}{*}{STIL}&\multirow{3}{*}{Video}&
       R & 0.8222 & $0.7712_{\textcolor[RGB]{0,116,0}{-0.0510}}$ & $0.7215_{\textcolor[RGB]{0,116,0}{-0.1007}}$  & $0.8076_{\textcolor[RGB]{0,116,0}{-0.0146}}$ & $0.7678_{\textcolor[RGB]{0,116,0}{-0.0544}}$ & $0.7694_{\textcolor[RGB]{0,116,0}{-0.0528}}$ & $0.7401_{\textcolor[RGB]{0,116,0}{-0.0821}}$ & $0.8205_{\textcolor[RGB]{0,116,0}{-0.0017}}$ & $0.8195_{\textcolor[RGB]{0,116,0}{-0.0027}}$\\
       &&F1& 0.7823&
       $0.7412_{\textcolor[RGB]{0,116,0}{-0.0411}}$ &
       $0.5124_{\textcolor[RGB]{0,116,0}{-0.2699}}$ &
        $0.7435_{\textcolor[RGB]{0,116,0}{-0.0388}}$ &
        $0.6589_{\textcolor[RGB]{0,116,0}{-0.1234}}$ &
        $0.7276_{\textcolor[RGB]{0,116,0}{-0.0547}}$ &
        $0.7301_{\textcolor[RGB]{0,116,0}{-0.0522}}$ &
        $0.6472_{\textcolor[RGB]{0,116,0}{-0.1351}}$ &
        $0.7566_{\textcolor[RGB]{0,116,0}{-0.0257}}$ &
       \\
       &&AP& 0.8712 & $0.8504_{\textcolor[RGB]{0,116,0}{-0.0208}}$ & $0.5938_{\textcolor[RGB]{0,116,0}{-0.2774}}$ & $0.8521_{\textcolor[RGB]{0,116,0}{-0.0191}}$ & $0.7638_{\textcolor[RGB]{0,116,0}{-0.1074}}$ & $0.8259_{\textcolor[RGB]{0,116,0}{-0.0453}}$ & $0.8309_{\textcolor[RGB]{0,116,0}{-0.0403}}$ & $0.7454_{\textcolor[RGB]{0,116,0}{-0.1258}}$&$0.8665_{\textcolor[RGB]{0,116,0}{-0.0047}}$\\
        \midrule
        \multirow{3}{*}{MINTIME-CLIP-B}&\multirow{3}{*}{Video}&
        R & 0.8724&$0.8215_{\textcolor[RGB]{0,116,0}{-0.0509}}$  &$0.8199_{\textcolor[RGB]{0,116,0}{-0.0525}}$&$0.8215_{\textcolor[RGB]{0,116,0}{-0.0508}}$ &$0.5534_{\textcolor[RGB]{0,116,0}{-0.3190}}$ &$0.8084_{\textcolor[RGB]{0,116,0}{-0.0640}}$ &$0.7972_{\textcolor[RGB]{0,116,0}{-0.0752}}$&$0.8628_{\textcolor[RGB]{0,116,0}{-0.0096}}$ &$0.8715_{\textcolor[RGB]{0,116,0}{-0.0009}}$\\
        &&F1 &0.8700 &$0.8486_{\textcolor[RGB]{0,116,0}{-0.0214}}$  &$\textbf{0.8417}_{\textcolor[RGB]{0,116,0}{-0.0283}}$ &$0.8486_{\textcolor[RGB]{0,116,0}{-0.0214}}$ &$0.5485_{\textcolor[RGB]{0,116,0}{-0.3215}}$ &$0.8468_{\textcolor[RGB]{0,116,0}{-0.0232}}$& $0.8379_{\textcolor[RGB]{0,116,0}{-0.0321}}$ &$0.4734_{\textcolor[RGB]{0,116,0}{-0.3966}}$& $0.8658_{\textcolor[RGB]{0,116,0}{-0.0042}}$\\
        &&AP &0.9369 &$0.9207_{\textcolor[RGB]{0,116,0}{-0.0158}}$  &$\textbf{0.9143}_{\textcolor[RGB]{0,116,0}{-0.0226}}$ &$0.9207_{\textcolor[RGB]{0,116,0}{-0.0158}}$ &$0.6199_{\textcolor[RGB]{0,116,0}{-0.3170}}$ &$0.9071_{\textcolor[RGB]{0,116,0}{-0.0298}}$ &$0.9227_{\textcolor[RGB]{0,116,0}{-0.0142}}$ &$0.8146_{\textcolor[RGB]{0,116,0}{-0.1223}}$ & $0.9364_{\textcolor[RGB]{0,116,0}{-0.0005}}$
        \\
       \midrule
        \multirow{3}{*}{FTCN-CLIP-B}&\multirow{3}{*}{Video}&
        R & 0.8935&$0.8519_{\textcolor[RGB]{0,116,0}{-0.0416}}$ &$0.8127_{\textcolor[RGB]{0,116,0}{-0.0808}}$ &$0.8858_{\textcolor[RGB]{0,116,0}{-0.0077}}$ &$0.5546_{\textcolor[RGB]{0,116,0}{-0.3389}}$& $0.8284_{\textcolor[RGB]{0,116,0}{-0.0651}}$&$0.8547_{\textcolor[RGB]{0,116,0}{-0.0388}}$&$0.8874_{\textcolor[RGB]{0,116,0}{-0.0061}}$ & $0.8904_{\textcolor[RGB]{0,116,0}{-0.0031}}$\\
        &&F1 &0.8739 &$0.8730_{\textcolor[RGB]{0,116,0}{-0.0009}}$& $0.5495_{\textcolor[RGB]{0,116,0}{-0.3244}}$&$0.8484_{\textcolor[RGB]{0,116,0}{-0.0451}}$ &$0.5573_{\textcolor[RGB]{0,116,0}{-0.3166}}$&$0.8689_{\textcolor[RGB]{0,116,0}{-0.0050}}$ &$0.8648_{\textcolor[RGB]{0,116,0}{-0.0091}}$ & $0.6002_{\textcolor[RGB]{0,116,0}{-0.2735}}$ & $0.8638_{\textcolor[RGB]{0,116,0}{-0.0101}}$\\
        &&AP &0.9418 &$0.9294_{\textcolor[RGB]{0,116,0}{-0.0124}}$ &$0.6884_{\textcolor[RGB]{0,116,0}{-0.2534}}$ &$0.9316_{\textcolor[RGB]{0,116,0}{-0.0102}}$ &$0.6015_{\textcolor[RGB]{0,116,0}{-0.3403}}$ & $0.9256_{\textcolor[RGB]{0,116,0}{-0.0162}}$& $0.9386_{\textcolor[RGB]{0,116,0}{-0.0032}}$ &$0.8503_{\textcolor[RGB]{0,116,0}{-0.0915}}$ & $0.9389_{\textcolor[RGB]{0,116,0}{-0.0029}}$\\
\midrule
        \multirow{3}{*}{TALL}&\multirow{3}{*}{Video}&
        R & 0.8852&$0.8413_{\textcolor[RGB]{0,116,0}{-0.0439}}$& $0.4272_{\textcolor[RGB]{0,116,0}{-0.4580}}$& $0.8785_{\textcolor[RGB]{0,116,0}{-0.0067}}$ & $0.7233_{\textcolor[RGB]{0,116,0}{-0.1619}}$& $0.8061_{\textcolor[RGB]{0,116,0}{-0.0791}}$ & $0.8045_{\textcolor[RGB]{0,116,0}{-0.0807}}$ & $0.8834_{\textcolor[RGB]{0,116,0}{-0.0018}}$ & $0.8816_{\textcolor[RGB]{0,116,0}{-0.0036}}$\\
        &&F1 & 0.7419& $0.7411_{\textcolor[RGB]{0,116,0}{-0.0008}}$ & $0.5309_{\textcolor[RGB]{0,116,0}{-0.2110}}$& $0.7248_{\textcolor[RGB]{0,116,0}{-0.0171}}$ & $0.5844_{\textcolor[RGB]{0,116,0}{-0.1575}}$& $0.7329_{\textcolor[RGB]{0,116,0}{-0.0090}}$ & $0.7360_{\textcolor[RGB]{0,116,0}{-0.0059}}$ & $0.5996_{\textcolor[RGB]{0,116,0}{-0.1423}}$ & $0.6719_{\textcolor[RGB]{0,116,0}{-0.0700}}$\\
        &&AP & 0.8791& $0.8673_{\textcolor[RGB]{0,116,0}{-0.0118}}$ & $0.6439_{\textcolor[RGB]{0,116,0}{-0.2352}}$& $0.8774_{\textcolor[RGB]{0,116,0}{-0.0017}}$ & $0.6744_{\textcolor[RGB]{0,116,0}{-0.2047}}$& $0.8608_{\textcolor[RGB]{0,116,0}{-0.0183}}$ & $0.8767_{\textcolor[RGB]{0,116,0}{-0.0024}}$& $0.7880_{\textcolor[RGB]{0,116,0}{-0.0911}}$ & $0.8439_{\textcolor[RGB]{0,116,0}{-0.0352}}$\\
       \midrule    
       \multirow{3}{*}{CLIP-B-FT}&\multirow{3}{*}{Image}&
        R & 0.9057 & $0.8253_{\textcolor[RGB]{0,116,0}{-0.0804}}$ & $0.7632_{\textcolor[RGB]{0,116,0}{-0.1425}}$ & $0.8878_{\textcolor[RGB]{0,116,0}{-0.0179}}$ & $0.6782_{\textcolor[RGB]{0,116,0}{-0.2275}}$ & $0.7877_{\textcolor[RGB]{0,116,0}{-0.1180}}$ & $0.7832_{\textcolor[RGB]{0,116,0}{-0.1225}}$ & $0.8812_{\textcolor[RGB]{0,116,0}{-0.0245}}$ & $0.8809_{\textcolor[RGB]{0,116,0}{-0.0248}}$\\
        &&F1 & 0.7631 &
        $0.7476_{\textcolor[RGB]{0,116,0}{-0.0155}}$ & 
        $0.6982_{\textcolor[RGB]{0,116,0}{-0.0649}}$ & 
        $0.7533_{\textcolor[RGB]{0,116,0}{-0.0098}}$ & 
        $0.6273_{\textcolor[RGB]{0,116,0}{-0.1358}}$ & 
        $0.7125_{\textcolor[RGB]{0,116,0}{-0.0506}}$ & 
        $0.7537_{\textcolor[RGB]{0,116,0}{-0.0094}}$ &
        $0.6743_{\textcolor[RGB]{0,116,0}{-0.0888}}$ &
        $0.7542_{\textcolor[RGB]{0,116,0}{-0.0089}}$ 
       \\
       &&AP& 0.9152 & $0.9042_{\textcolor[RGB]{0,116,0}{-0.0110}}$ & $0.8533_{\textcolor[RGB]{0,116,0}{-0.0619}}$ & $0.9105_{\textcolor[RGB]{0,116,0}{-0.0047}}$ & $0.7635_{\textcolor[RGB]{0,116,0}{-0.1517}}$ & $0.8894_{\textcolor[RGB]{0,116,0}{-0.0258}}$ & $0.9105_{\textcolor[RGB]{0,116,0}{-0.0047}}$ & $0.8455_{\textcolor[RGB]{0,116,0}{-0.0697}}$ & $0.9105_{\textcolor[RGB]{0,116,0}{-0.0047}}$\\
       \midrule       \multirow{3}{*}{DeMamba-CLIP-FT}&\multirow{3}{*}{Video}&
       R& 0.9158&$0.8572_{\textcolor[RGB]{0,116,0}{-0.0586}}$&$0.8479_{\textcolor[RGB]{0,116,0}{-0.0679}}$&$0.9047_{\textcolor[RGB]{0,116,0}{-0.0111}}$&$0.6172_{\textcolor[RGB]{0,116,0}{-0.2986}}$ & $0.8352_{\textcolor[RGB]{0,116,0}{-0.0806}}$ & $0.8664_{\textcolor[RGB]{0,116,0}{-0.0494}}$ & $0.9037_{\textcolor[RGB]{0,116,0}{-0.0121}}$ & $0.9105_{\textcolor[RGB]{0,116,0}{-0.0053}}$ \\
       &&F1 & 0.8919 &  $0.8750_{\textcolor[RGB]{0,116,0}{-0.0169}}$ &
       $0.6325_{\textcolor[RGB]{0,116,0}{-0.2594}}$ &
       $0.8912_{\textcolor[RGB]{0,116,0}{-0.0007}}$ &
       $0.5233_{\textcolor[RGB]{0,116,0}{-0.3686}}$ &
       $0.8455_{\textcolor[RGB]{0,116,0}{-0.0464}}$ &
       $0.8895_{\textcolor[RGB]{0,116,0}{-0.0024}}$ &
       $0.7672_{\textcolor[RGB]{0,116,0}{-0.1247}}$ &
       $0.8910_{\textcolor[RGB]{0,116,0}{-0.0009}}$ &
       \\
       &&AP&0.9345&$0.9093_{\textcolor[RGB]{0,116,0}{-0.0252}}$&$0.7403_{\textcolor[RGB]{0,116,0}{-0.1942}}$& $0.9344_{\textcolor[RGB]{0,116,0}{-0.0001}}$ & $0.6112_{\textcolor[RGB]{0,116,0}{-0.3233}}$ & $0.9008_{\textcolor[RGB]{0,116,0}{-337}}$ & $0.9335_{\textcolor[RGB]{0,116,0}{-0.0010}}$ & $0.8299_{\textcolor[RGB]{0,116,0}{-0.1046}}$ & $0.9339_{\textcolor[RGB]{0,116,0}{-0.0006}}$ \\\midrule
       \multirow{3}{*}{XCLIP-B-FT}&\multirow{3}{*}{Video}&
       R & 0.8441& $0.8387_{\textcolor[RGB]{0,116,0}{-0.0054}}$ & $0.5377_{\textcolor[RGB]{0,116,0}{-0.3064}}$  & $0.8378_{\textcolor[RGB]{0,116,0}{-0.0063}}$ & $0.5597_{\textcolor[RGB]{0,116,0}{-0.2844}}$ & $0.8388_{\textcolor[RGB]{0,116,0}{-0.0023}}$ & $0.8254_{\textcolor[RGB]{0,116,0}{-0.0187}}$ & $0.8195_{\textcolor[RGB]{0,116,0}{-0.0216}}$ & $0.8421_{\textcolor[RGB]{0,116,0}{-0.0020}}$ \\
       &&F1&0.7662 & $0.7329_{\textcolor[RGB]{0,116,0}{-0.0333}}$&
       $0.4188_{\textcolor[RGB]{0,116,0}{-0.3474}}$&
      $0.7592_{\textcolor[RGB]{0,116,0}{-0.0070}}$&
        $0.3062_{\textcolor[RGB]{0,116,0}{-0.4600}}$&
    $0.7595_{\textcolor[RGB]{0,116,0}{-0.0067}}$&
    $0.7593_{\textcolor[RGB]{0,116,0}{-0.0069}}$&
    $0.6322_{\textcolor[RGB]{0,116,0}{-0.1340}}$&
    $0.7488_{\textcolor[RGB]{0,116,0}{-0.0174}}$
       \\
       &&AP& 0.8677 & $0.8653_{\textcolor[RGB]{0,116,0}{-0.0024}}$ & $0.5411_{\textcolor[RGB]{0,116,0}{-0.3266}}$ & $0.8676_{\textcolor[RGB]{0,116,0}{-0.0001}}$ & $0.4610_{\textcolor[RGB]{0,116,0}{-0.4067}}$ & $0.8671_{\textcolor[RGB]{0,116,0}{-0.0006}}$ & $0.8674_{\textcolor[RGB]{0,116,0}{-0.0003}}$ & $0.6805_{\textcolor[RGB]{0,116,0}{-0.1489}}$ & $0.8601_{\textcolor[RGB]{0,116,0}{-0.0076}}$\\
       \midrule
       \multirow{3}{*}{DeMamba-XCLIP-FT}&\multirow{3}{*}{Video}&
       R&\textbf{0.9392}&$\textbf{0.9052}_{\textcolor[RGB]{0,116,0}{-0.0340}}$ & $\textbf{0.9326}_{\textcolor[RGB]{0,116,0}{-0.0116}}$ & $\textbf{0.9376}_{\textcolor[RGB]{0,116,0}{-0.0016}}$ & $\textbf{0.7298}_{\textcolor[RGB]{0,116,0}{-0.2244}}$ & $\textbf{0.8915}_{\textcolor[RGB]{0,116,0}{-0.0477}}$ & $\textbf{0.9094}_{\textcolor[RGB]{0,116,0}{-0.0288}}$ & $\textbf{0.9382}_{\textcolor[RGB]{0,116,0}{-0.0010}}$& $\textbf{0.9342}_{\textcolor[RGB]{0,116,0}{-0.0050}}$\\
       &&F1 &\textbf{0.9020} & $\textbf{0.8770}_{\textcolor[RGB]{0,116,0}{-0.0250}}$ & 
       $0.6823_{\textcolor[RGB]{0,116,0}{-0.2197}}$ &
    $\textbf{0.8820}_{\textcolor[RGB]{0,116,0}{-0.0200}}$ &
     $\textbf{0.6595}_{\textcolor[RGB]{0,116,0}{-0.2425}}$ &
      $\textbf{0.8755}_{\textcolor[RGB]{0,116,0}{-0.0265}}$ &
       $\textbf{0.8990}_{\textcolor[RGB]{0,116,0}{-0.0030}}$ &
        $\textbf{0.7962}_{\textcolor[RGB]{0,116,0}{-0.1058}}$ &
         $\textbf{0.9005}_{\textcolor[RGB]{0,116,0}{-0.0005}}$ &
       \\
       &&AP&\textbf{0.9710}& 
       $\textbf{0.9600}_{\textcolor[RGB]{0,116,0}{-0.0110}}$ & $0.7629_{\textcolor[RGB]{0,116,0}{-0.2081}}$ &$\textbf{0.9676}_{\textcolor[RGB]{0,116,0}{-0.0034}}$ & $\textbf{0.6718}_{\textcolor[RGB]{0,116,0}{-0.2992}}$ & $\textbf{0.9430}_{\textcolor[RGB]{0,116,0}{-0.0280}}$ & $\textbf{0.9707}_{\textcolor[RGB]{0,116,0}{-0.0003}}$ & $\textbf{0.8505}_{\textcolor[RGB]{0,116,0}{-0.1205}}$& $\textbf{0.9705}_{\textcolor[RGB]{0,116,0}{-0.0005}}$
       \\
        \bottomrule
    \end{tabular}}
    \label{robustness_results}
\end{table*}

\begin{table*}[t]
    \centering
    
    \caption{Ablation study of DeMamba.}
    \label{ablation_table}
    \resizebox{\textwidth}{!}{
    \begin{tabular}{ccccccccccccc}
    \toprule
       \multirow{2}{*}{Model}& \multirow{2}{*}{Metric}&\multirow{2}{*}{Sora}&Morph&\multirow{2}{*}{Gen2}&\multirow{2}{*}{HotShot}&\multirow{2}{*}{Lavie}&\multirow{2}{*}{Show-1}&Moon&\multirow{2}{*}{Crafter}&Model&Wild&\multirow{2}{*}{Avg.}\\
       &&&Studio&&&&&Valley&&Scope&Scrape\\
       
       \midrule
       \multirow{3}{*}{w/o Demamba}&
       R&0.8214&0.9957 &0.9362 & 0.6129 & 0.7936&0.6971 &0.9792&0.9979
       & 0.7714 & 0.8359&$0.8411_{\textcolor[RGB]{0,116,0}{-0.0971}}$\\
       &       F1&0.3147&
0.8805&
0.9041&
0.6530&
0.8242&
0.7099&
0.8602&
0.9370&
0.7570&
0.8212&
$0.7662_{\textcolor[RGB]{0,116,0}{-0.1358}}$\\
       
        &AP&0.6442& 0.9973&0.9678& 0.7098 & 0.9035& 0.7728&0.9734&0.9984
       &0.8201& 0.8897&$0.8677_{\textcolor[RGB]{0,116,0}{-0.1043}}$\\ \midrule
       
       \multirow{3}{*}{w/o global}&
        R & 0.9733& 0.9985 &0.9945 & 0.6766 & 0.9355 & 0.9788 & 1.0000& 1.0000& 0.9134 &0.8654 & $0.9336_{\textcolor[RGB]{0,116,0}{-0.0046}}$\\
       &F1& 0.4252&
0.9053&
0.9475&
0.7183&
0.9176&
0.8960&
0.8962&
0.9508&
0.8618&
0.8562&
$0.8375_{\textcolor[RGB]{0,116,0}{-0.0645}}$
       \\
       &AP & 0.8732 & 0.9935 & 0.9922 & 0.8244 & 0.9799 & 0.9942& 0.9942& 0.9985& 0.9732& 0.9498&$0.9573_{\textcolor[RGB]{0,116,0}{-0.0137}}$
       \\ \midrule
       \multirow{3}{*}{DeMamba-XCLIP-FT}&R
       &0.9821& 1.0000 & 0.9986 & 0.6543 & 0.9486 & 0.9886 & 1.0000& 1.0000&0.9286 &0.8909 & \textbf{0.9382}
       \\
       &F1&0.6467& 0.9602& 0.9790& 0.7539& 0.9537&0.9551&0.9557&0.9797&0.9240&0.9120&\textbf{0.9020}
        \\
       &AP&0.9332 & 1.0000 & 0.9997 & 0.8555 & 0.9897 & 0.9960& 0.9998& 1.0000& 0.9777& 0.9575&\textbf{0.9710}
       \\

        \bottomrule
    \end{tabular}}
\end{table*}

\begin{table*}[t]
    \centering
    
    \caption{Influence of different zone sizes in DeMamba.}
    \label{zone_size}
    \resizebox{\textwidth}{!}{
    \begin{tabular}{cccccccccccccc}
    \toprule
       \multirow{2}{*}{Model} & Zone&\multirow{2}{*}{Metric}&\multirow{2}{*}{Sora}&Morph&\multirow{2}{*}{Gen2}&\multirow{2}{*}{HotShot}&\multirow{2}{*}{Lavie}&\multirow{2}{*}{Show-1}&Moon&\multirow{2}{*}{Crafter}&Model&Wild&\multirow{2}{*}{Avg.}\\
       &size&&&Studio&&&&&Valley&&Scope&Scrape\\
       
        \midrule
        &\multirow{3}{*}{$1\times 1$}&R& 1.0000 & 0.9986 & 0.8478 & 0.9557 & 0.8143 & 0.8229 & 1.0000 & 0.9979 & 0.7829 & 0.8110 & 0.9031
        \\
        &&F1& 0.3218& 
0.8556& 
0.8398& 
0.8334& 
0.8216& 
0.7624& 
0.8414& 
0.9215& 
0.7390& 
0.7849& 
0.7722\\
        &&AP& 0.9380 & 0.9964 & 0.9178 & 0.9603 & 0.9004 & 0.8613 & 0.9992 & 0.9981 & 0.8212 & 0.8696 & 0.9262
        \\
        \cmidrule(lr){2-14}
       &\multirow{3}{*}{$2\times 2$}&
        R& 0.9571 & 1.0000 & 0.9870 & 0.6914 & 0.9243 & 0.9329 & 1.0000 & 1.0000 & 0.8357 & 0.8294 &0.9158 \\
       &&F1& 0.6463&
0.9615&
0.9739&
0.7803&
0.9414&
0.9276&
0.9572&
0.9804&
0.8723&
0.8782&
\textbf{0.8919}
       \\
       DeMamba&&AP& 0.8550 & 1.0000 & 0.9959 & 0.7615 & 0.9678 & 0.9699 & 0.9997 & 1.0000 & 0.8980 & 0.8972 & \textbf{0.9345}
       \\
       \cmidrule(lr){2-14}
       -CLIP-FT&\multirow{3}{*}{$7\times 7$}&
        R&
       0.9821 & 1.0000 & 0.9841 & 0.6500 & 0.9400 &0.9543& 1.0000 & 1.0000 & 0.8086 & 0.8434 &0.9163\\
       &&F1& 0.3724&
0.8855&
0.9308&
0.6811&
0.9085&
0.8630&
0.8737&
0.9393&
0.7828&
0.8271&
0.8064
       \\
       
       &&AP&0.8728 & 0.9999 & 0.9906 & 0.7351 & 0.9747 & 0.9700 & 0.9976 & 0.9997 & 0.8575 & 0.8995 & 0.9297
       \\
       \cmidrule(lr){2-14}
       &\multirow{3}{*}{$14\times 14$}&
       R& 1.0000  &  0.9986 &  0.9341 &  0.8857 &  0.8171 &  0.7014 & 1.0000 &  0.9993 &  0.7643 &  0.8002 & 0.8901\\
       &&F1& 0.4628&
0.9149&
0.9214&
0.8550&
0.8555&
0.7430&
0.9059&
0.9556&
0.7845&
0.8245&
0.8223
       \\
        &&AP& 0.9494 & 0.9942 & 0.9703 & 0.9356 & 0.9402 & 0.8456 & 0.9912 &0.9922 & 0.7760 & 0.8946  & 0.9289
        
        \\
       \midrule
       &\multirow{3}{*}{$1\times 1$}&
       R& 0.9107 & 1.0000 & 0.9935 & 0.7371 & 0.8950 & 0.9629 & 1.0000 & 1.0000 & 0.9471&0.8121&0.9258 \\
       &&F1& 0.5952&
0.9569&
0.9748&
0.8066&
0.9227&
0.9387&
0.9521&
0.9780&
0.9298&
0.8644&
0.8919
       \\
        &&AP& 0.8665 & 0.9999 & 0.9983 & 0.9064 & 0.9785 & 0.9881 & 0.9991 & 0.9999 & 0.9843 &0.9218 &  0.9643
        \\
        \cmidrule(lr){2-14}
        &\multirow{3}{*}{$2\times 2$}&
      R
       &0.9821& 1.0000 & 0.9986 & 0.6543 & 0.9486 & 0.9886 & 1.0000& 1.0000&0.9286 &0.8909 & 0.9392\\
        &&F1&0.6467& 0.9602& 0.9790& 0.7539& 0.9537&0.9551&0.9557&0.9797&0.9240&0.9120&\textbf{0.9020}
        \\
       DeMamba&&AP&0.9332 & 1.0000 & 0.9997 & 0.8555 & 0.9897 & 0.9960& 0.9998& 1.0000& 0.9777& 0.9575&\textbf{0.9710}

       \\
       \cmidrule(lr){2-14}
       -XCLIP-FT&\multirow{3}{*}{$7\times 7$}&
        R&0.9643&1.0000&0.9957 & 0.5600 & 0.8871& 0.9600
       &1.0000& 1.0000& 0.9443&0.7948&0.9106\\
       &&F1& 0.7013&
0.9689&
0.9821&
0.6895&
0.9244&
0.9485&
0.9653&
0.9842&
0.9409&
0.8622&
0.8967
       \\
        &&AP& 0.9602& 1.0000& 0.9997& 0.8285& 0.9969& 0.9952& 0.9998& 1.0000& 0.8651& 0.9745& 0.9620
       \\
       \cmidrule(lr){2-14}
       &\multirow{3}{*}{$14\times 14$}&
       R& 0.7679 & 1.0000 & 0.9906& 0.2286 &0.8471 &0.9514 &0.9952 & 1.0000& 0.8771 & 0.7225& 0.8380\\
      && F1& 0.7748&
0.9908&
0.9909&
0.3670&
0.9126&
0.9659&
0.9873&
0.9954&
0.9253&
0.8326&
0.8742
       \\
        &&AP& 0.8043 & 1.0000 & 0.9976 & 0.6569 & 0.9849 &0.9931 & 0.9979& 0.9991 & 0.9687 & 0.8831 & 0.9286
        \\
        \bottomrule
    \end{tabular}}
    \label{main_results}
\end{table*}

\begin{table*}[t]
    \centering
    \caption{Influence of scanning order in DeMamba.}
    \label{scan}
    \resizebox{\textwidth}{!}{
    \begin{tabular}{cccccccccccccc}
    \toprule
       \multirow{2}{*}{Model} & Scan&\multirow{2}{*}{Metric}&\multirow{2}{*}{Sora}&Morph&\multirow{2}{*}{Gen2}&\multirow{2}{*}{HotShot}&\multirow{2}{*}{Lavie}&\multirow{2}{*}{Show-1}&Moon&\multirow{2}{*}{Crafter}&Model&Wild&\multirow{2}{*}{Avg.}\\
       &Rrder&&&Studio&&&&&Valley&&Scope&Scrape\\
       
        \midrule
       &\multirow{3}{*}{sweep} &
       R & 0.9521& 1.0000 & 0.9255 & 0.6177 & 0.9269 & 0.9633 & 0.9987& 0.9932 &0.8153 &0.7311 &0.8924 \\
       &&F1& 0.4649&
0.9211&
0.9200&
0.6906&
0.9212&
0.9029&
0.9124&
0.9557&
0.8207&
0.7856&
$0.8295_{\textcolor[RGB]{0,116,0}{-0.0725}}$\\
       DeMamba&&AP& 0.9332 & 0.9995 & 0.9732 & 0.8555 & 0.9588 & 0.9754& 0.9966& 0.9754 & 0.8713& 0.8082& $0.9347_{\textcolor[RGB]{0,116,0}{-0.0363}}$
       \\
       \cmidrule(lr){2-14}
       -XCLIP-FT&\multirow{3}{*}{continuous} &
        R& 0.9821& 1.0000 & 0.9986 & 0.6543 & 0.9486 & 0.9886 & 1.0000& 1.0000&0.9286 &0.8909 &0.9392 \\
       &&F1&0.6467& 0.9602& 0.9790& 0.7539& 0.9537&0.9551&0.9557&0.9797&0.9240&0.9120&\textbf{0.9020}
       \\
        &&AP&0.9332 & 1.0000 & 0.9997 & 0.8555 & 0.9897 & 0.9960& 0.9998& 1.0000& 0.9777& 0.9575&\textbf{0.9710}
        \\
        \bottomrule
    \end{tabular}}
    \label{main_results}
\end{table*}

\textbf{Evaluation metrics}. Consistent with the methodologies employed in prior studies, our evaluation framework primarily focuses on reporting accuracy (ACC) and average precision (AP) to assess the effectiveness of the detectors. The accuracy calculation is based on a threshold value of 0.5. For image-based detection techniques, we consolidate frame-level predictions to derive the corresponding video-level predictions, ensuring a coherent analysis across different media formats. It is noteworthy that when evaluating the performance on a dataset generated by a specific synthesis method, we calculate the ACC of that synthesis method based on the dataset itself. Additionally, in the process of computing the AP, we take into account real videos to achieve a more comprehensive assessment.

\textbf{Baselines.} Three type of image-level methods and six video-level methods are treated as baselines for comparison: F3Net~\cite{f3net}, NPR~\citep{NPR}, CLIP~\citep{CLIP}, STIL~\citep{stil}, XCLIP~\citep{XCLIP}, VideoMAE~\citep{videomae}, MINTIME~\citep{MINTIME}, FTCN~\citep{FTCN}, TALL~\citep{TALL}.

\textbf{Training settings.} As shown in Table~\ref{param_set}, we present the different model training parameter settings for many-to-many and one-to-many generalization tasks. All of our experiments were conducted on a system equipped with 8 Tesla A-100 80G GPUs and an Intel(R) Xeon(R) Platinum 8369B CPU @ 2.90GHz. 

\subsection{Task1: cross generator generalization}
Due to the rapid iteration of generation methods, we propose a cross-dataset generalization task to test the generalization performance of detectors. Specifically, it consists of two types of generalization tasks: 1) the many-to-many generalization task, and 2) the one-to-many generalization task.

\textbf{Many-to-many generalization task.} This task involves training on 10 baseline categories and then testing on each subset and the average detection performance on $D_{\rm v-ood}$. As shown in Table \ref{tab_m2m}, video models achieve better recognition accuracy compared to image models because video models can model temporal sequences. Moreover, our DeMamba model can be effectively integrated into existing models, achieving significant improvements. For example, integrating the DeMamba module into XCLIP results in DeMamba-XCLIP-FT achieving an average recall/F1/AP of 0.9392/0.9020/0.9710 respectively, which marks an improvement of 11.26\%/17.72\%/12.02\% in recall/F1/AP over the original XCLIP. Note that PT (Partially Tuning) indicates that the backbone is frozen, with only the other parts being fine-tuned, while FT (Full Training) tuning the entire model.

\textbf{One-to-many generalization task.} Following AI-generated image detection setting~\citep{NPR,On_Det,DIRE}, we also perform a one-to-many generalization task.
Unlike the many-to-many generalization task, the one-to-many generalization task involves training on one baseline category and then testing on each subset and the average detection performance on $D_{\rm v-ood}$.
As shown in Table \ref{12many}, our DeMamba-XCLIP-FT achieves better generalization performance in three one-to-many generalization tasks due to the learning of spatial-temporal inconsistency in DeMamba.

\subsection{Task2: degraded video classification}
In practical detection scenarios, the robustness of the detector to perturbations is also of paramount importance. In this regard, we investigated the impact of perturbations on the detector on 8 different types: H.264 compression, JPEG compression, FLIP, Crop, text watermark, image watermark, Gaussian noise, and color transform. More specific details about the perturbations can be found in \textbf{\cref{A3_de}}. Table \ref{robustness_results} shows the performance of the models trained in the many-to-many task under the influence of these perturbations. We can see that in the case of degraded data, DeMamba-XCLIP-FT still achieves the best performance, indicating that our model has good robustness in the face of degraded data.

\subsection{Ablation study}
\label{A4}
\noindent
\textbf{Ablation testing.} We conducted ablation experiments to validate the effectiveness of DeMamba. As shown in Table~\ref{ablation_table}, DeMamba effectively enhances the generalization performance of the model. Additionally, when using fused features, the model achieves its best performance.

\noindent
\textbf{Influence of different zone sizes.} We investigated the impact of zone size in dividing zones in DeMamba on modeling temporal inconsistency. As shown in Table~\ref{zone_size}, the best performance is observed when the zone size is 2. Smaller zones enable the model to concentrate more on local details, leading to superior modeling performance. However, excessively small zones may result in the loss of spatial contextual information. Therefore, selecting an appropriate zone size is crucial.

\noindent
\textbf{Influence of scanning orders.}  As shown in Table \ref{scan}, the continuous scan proposed in this paper effectively enhances performance compared to the traditional scanning method.

\noindent
\begin{wrapfigure}{r}{0.45\textwidth}
    \centering
    \includegraphics[width=0.45\textwidth]{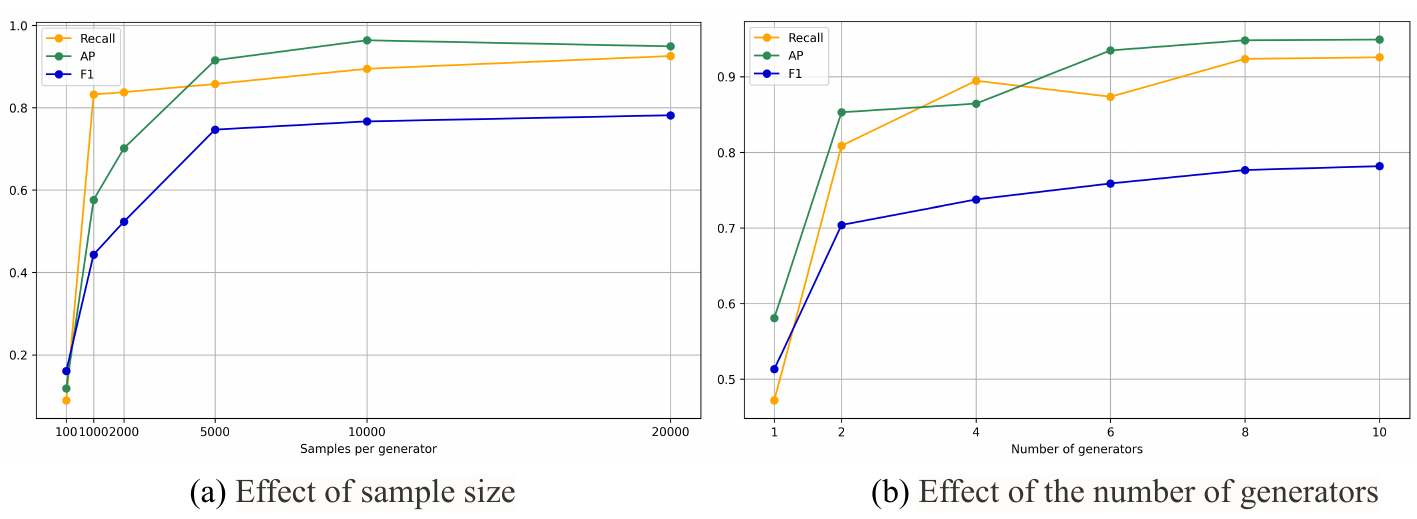}
    \caption{Performance of training on scaled-up datasets on the testing set.}
    \label{scaling}
\end{wrapfigure}

\textbf{Necessity of large volume of GenVideo for generalizability.} To clarify the impact of data volume on detection performance, we conducted experiments using DeMamba on various scaled subsets of the GenVideo dataset. We designed two experiments to explore: 1) variation in the number of training samples within each generator: For the data generated by each generator, we randomly selected between 100 and 20,000 video samples to train the model. 2) variation in the number of training samples across different generators: We randomly selected subsets to simulate scenarios with 1 to 10 generators in the dataset. In each scenario, we collected 20,000 video samples for the videos generated by each generator. Note that, for real videos, we randomly selected an equal number of samples to the total number of training generated video data to participate in the training. The test sets used in these experiments are the same as those mentioned in the main paper, they consist of the full test set videos, and the model selected for the experiment is DeMamba-XCLIP-FT. The comparative results of these experiments are detailed in Figure 4. The results clearly indicate a significant association between the scale of the dataset and the improvement in detection performance, further affirming the critical role of massive data in enhancing detection capabilities.

\begin{wraptable}{r}{0.6\textwidth} 
    \centering
    \caption{Training/inference time and QPS for different models.}
    \resizebox{0.6\textwidth}{!}{
    \begin{tabular}{cccc}
    \hline
    Model & Training time (h) & Inference time (s) & QPS \\
    \hline
    F3Net & 40 & 188 & 2.47 \\
    NPR & 25.1 & 149 & 2.66 \\
    STIL & 11.7 & 174 & 2.42 \\
    VideoMAE & 52.5 & 283 & 1.32 \\
    MINTIME & 18.1 & 148 & 2.61 \\
    FTCN & 11.8 & 148 & 2.68 \\
    TALL & 6 & 150 & 2.71 \\
    CLIP-B-FT & 10.1 & 148 & 2.56 \\
    DeMamba-CLIP-B-FT & 14 & 150 & 2.54 \\
    XCLIP-FT & 11.5 & 149 & 2.55 \\
    DeMamba-XCLIP-FT & 15.1 & 153 & 2.53 \\
    \hline
    \end{tabular}}
    \label{time}
\end{wraptable}
\textbf{Training and inference efficiency.} As detailed in Table~\ref{time}, we report the training and inference times on 8 Tesla A100-80G GPUs, with QPS measured on a single GPU at batch size 1. Incorporating DeMamba into XCLIP does not significantly impact inference time compared to CLIP and XCLIP. Slower training is due to Mamba's training characteristics, but it offers efficient inference post-training. Our method maintains competitive computational times relative to other video models, outperforming F3Net, STIL, and TALL.

\section{Broader impacts}
Our research focuses on utilizing machine learning to detect generated videos. We have introduced the first million-scale AI-generated video detection dataset and developed the DeMamba model. These efforts are crucial for protecting digital content and preventing the spread of misinformation. However, there is a potential for these tools to be misused, leading to competition between video generation and detection technologies. We aim to advocate for the ethical use of technology and promote creative research into tools that verify media authenticity. We believe this will help protect the public from the harm of misinformation, enhance the clarity and authenticity of information dissemination, and ensure the protection of personal privacy.

\section{Conclusion and limitation}
This paper introduces GenVideo, a dataset specifically designed for detecting fake videos generated by generative models. GenVideo is characterized by its large-scale nature, as well as the rich diversity of generated content and methods. We propose two tasks that mimic real-world scenarios, namely the cross-generator video classification task and the degraded video classification task, to evaluate the detection performance of existing detectors on GenVideo. Additionally, we introduce a plug-and-play effective detection model called Detail Mamba (DeMamba), which distinguishes AI-generated videos by analyzing inconsistencies in the spatial-temporal dimensions. This model has demonstrated its strong generalization and robustness across multiple tasks. We hope that this research will inspire the creation and improvement of other detection technologies, providing new avenues for the development of authentic and reliable AI-generated content applications.

The main limitation of this article lies in the suboptimal training efficiency of the proposed DeMamba, a common issue with the Mamba model. Consequently, we encourage the community to design more lightweight and generalized detection models to facilitate the regulation of AI-generated content.

\newpage
\bibliographystyle{unsrtnat}
\bibliography{ref}

\begin{thebibliography}{97}
\providecommand{\natexlab}[1]{#1}
\providecommand{\url}[1]{\texttt{#1}}
\expandafter\ifx\csname urlstyle\endcsname\relax
  \providecommand{\doi}[1]{doi: #1}\else
  \providecommand{\doi}{doi: \begingroup \urlstyle{rm}\Url}\fi

\bibitem[Zhang et~al.(2023{\natexlab{a}})Zhang, Rao, and Agrawala]{controlnet}
Lvmin Zhang, Anyi Rao, and Maneesh Agrawala.
\newblock Adding conditional control to text-to-image diffusion models.
\newblock In \emph{CVPR}, pages 3813--3824, 2023{\natexlab{a}}.

\bibitem[Chen et~al.(2023{\natexlab{a}})Chen, Xu, Gu, Lan, Zheng, Li, Meng, Zhu, and Wang]{diffute}
Haoxing Chen, Zhuoer Xu, Zhangxuan Gu, Jun Lan, Xing Zheng, Yaohui Li, Changhua Meng, Huijia Zhu, and Weiqiang Wang.
\newblock Diffute: Universal text editing diffusion model.
\newblock In \emph{NeurIPS}, 2023{\natexlab{a}}.

\bibitem[Li et~al.(2023)Li, Wang, Jin, Hu, Chemerys, Fu, Wang, Tulyakov, and Ren]{SnapFusion}
Yanyu Li, Huan Wang, Qing Jin, Ju~Hu, Pavlo Chemerys, Yun Fu, Yanzhi Wang, Sergey Tulyakov, and Jian Ren.
\newblock Snapfusion: Text-to-image diffusion model on mobile devices within two seconds.
\newblock In \emph{NeurIPS}, 2023.

\bibitem[Blattmann et~al.(2023{\natexlab{a}})Blattmann, Dockhorn, Kulal, Mendelevitch, Kilian, Lorenz, Levi, English, Voleti, Letts, et~al.]{SVD}
Andreas Blattmann, Tim Dockhorn, Sumith Kulal, Daniel Mendelevitch, Maciej Kilian, Dominik Lorenz, Yam Levi, Zion English, Vikram Voleti, Adam Letts, et~al.
\newblock Stable video diffusion: Scaling latent video diffusion models to large datasets.
\newblock \emph{arXiv preprint arXiv:2311.15127}, 2023{\natexlab{a}}.

\bibitem[Liu et~al.(2023{\natexlab{a}})Liu, Cun, Liu, Wang, Zhang, Chen, Liu, Zeng, Chan, and Shan]{Evalcrafter}
Yaofang Liu, Xiaodong Cun, Xuebo Liu, Xintao Wang, Yong Zhang, Haoxin Chen, Yang Liu, Tieyong Zeng, Raymond Chan, and Ying Shan.
\newblock Evalcrafter: Benchmarking and evaluating large video generation models.
\newblock \emph{arXiv preprint arXiv:2310.11440}, 2023{\natexlab{a}}.

\bibitem[Wang et~al.(2023{\natexlab{a}})Wang, Chen, Ma, Zhou, Huang, Wang, Yang, He, Yu, Yang, et~al.]{Lavie}
Yaohui Wang, Xinyuan Chen, Xin Ma, Shangchen Zhou, Ziqi Huang, Yi~Wang, Ceyuan Yang, Yinan He, Jiashuo Yu, Peiqing Yang, et~al.
\newblock Lavie: High-quality video generation with cascaded latent diffusion models.
\newblock \emph{arXiv preprint arXiv:2309.15103}, 2023{\natexlab{a}}.

\bibitem[GoogleAI()]{veo}
GoogleAI.
\newblock Veo.
\newblock \url{https://deepmind.google/technologies/veo/}.
\newblock Accessed: 2024-05.

\bibitem[Brooks et~al.(2024)Brooks, Peebles, Holmes, DePue, Guo, Jing, Schnurr, Taylor, Luhman, Luhman, Ng, Wang, and Ramesh]{sora}
Tim Brooks, Bill Peebles, Connor Holmes, Will DePue, Yufei Guo, Li~Jing, David Schnurr, Joe Taylor, Troy Luhman, Eric Luhman, Clarence Ng, Ricky Wang, and Aditya Ramesh.
\newblock Video generation models as world simulators, 2024.
\newblock URL \url{https://openai.com/index/sora/}.

\bibitem[Research(2023)]{Gen2}
Runway Research.
\newblock Text driven video generation, 2023.
\newblock URL \url{https://research.runwayml.com/gen2}.

\bibitem[Barrett et~al.(2023)Barrett, Boyd, Bursztein, Carlini, Chen, Choi, Chowdhury, Christodorescu, Datta, Feizi, et~al.]{barrett2023identifying}
Clark Barrett, Brad Boyd, Elie Bursztein, Nicholas Carlini, Brad Chen, Jihye Choi, Amrita~Roy Chowdhury, Mihai Christodorescu, Anupam Datta, Soheil Feizi, et~al.
\newblock Identifying and mitigating the security risks of generative ai.
\newblock \emph{Foundations and Trends{\textregistered} in Privacy and Security}, 6\penalty0 (1):\penalty0 1--52, 2023.

\bibitem[Gu et~al.(2021{\natexlab{a}})Gu, Chen, Yao, Ding, Li, Huang, and Ma]{stil}
Zhihao Gu, Yang Chen, Taiping Yao, Shouhong Ding, Jilin Li, Feiyue Huang, and Lizhuang Ma.
\newblock Spatiotemporal inconsistency learning for deepfake video detection.
\newblock In \emph{ACM Multimedia}, pages 3473--3481, 2021{\natexlab{a}}.

\bibitem[Xu et~al.(2023{\natexlab{a}})Xu, Liang, Jia, Yang, Zhang, and He]{TALL}
Yuting Xu, Jian Liang, Gengyun Jia, Ziming Yang, Yanhao Zhang, and Ran He.
\newblock {TALL:} thumbnail layout for deepfake video detection.
\newblock In \emph{ICCV}, pages 22601--22611, 2023{\natexlab{a}}.

\bibitem[Gu et~al.(2022{\natexlab{a}})Gu, Yao, Chen, Ding, and Ma]{HCIL}
Zhihao Gu, Taiping Yao, Yang Chen, Shouhong Ding, and Lizhuang Ma.
\newblock Hierarchical contrastive inconsistency learning for deepfake video detection.
\newblock In \emph{ECCV}, volume 13672, pages 596--613, 2022{\natexlab{a}}.

\bibitem[Xu et~al.(2023{\natexlab{b}})Xu, Ye, Wu, Yan, Miao, Ye, Xu, Hu, Shi, Xu, Li, Qian, Que, Zhang, Zeng, and Huang]{Youku-mPLUG}
Haiyang Xu, Qinghao Ye, Xuan Wu, Ming Yan, Yuan Miao, Jiabo Ye, Guohai Xu, Anwen Hu, Yaya Shi, Guangwei Xu, Chenliang Li, Qi~Qian, Maofei Que, Ji~Zhang, Xiao Zeng, and Fei Huang.
\newblock Youku-mplug: {A} 10 million large-scale chinese video-language dataset for pre-training and benchmarks.
\newblock \emph{arXiv preprint arXiv:2306.04362}, 2023{\natexlab{b}}.

\bibitem[Kay et~al.(2017{\natexlab{a}})Kay, Carreira, Simonyan, Zhang, Hillier, Vijayanarasimhan, Viola, Green, Back, Natsev, Suleyman, and Zisserman]{K400}
Will Kay, Jo{\~{a}}o Carreira, Karen Simonyan, Brian Zhang, Chloe Hillier, Sudheendra Vijayanarasimhan, Fabio Viola, Tim Green, Trevor Back, Paul Natsev, Mustafa Suleyman, and Andrew Zisserman.
\newblock The kinetics human action video dataset.
\newblock \emph{arXiv preprint arXiv:1705.06950}, 2017{\natexlab{a}}.

\bibitem[Xu et~al.(2016{\natexlab{a}})Xu, Mei, Yao, and Rui]{MSR-VTT}
Jun Xu, Tao Mei, Ting Yao, and Yong Rui.
\newblock {MSR-VTT:} {A} large video description dataset for bridging video and language.
\newblock In \emph{CVPR}, pages 5288--5296, 2016{\natexlab{a}}.

\bibitem[Qian et~al.(2020)Qian, Yin, Sheng, Chen, and Shao]{f3net}
Yuyang Qian, Guojun Yin, Lu~Sheng, Zixuan Chen, and Jing Shao.
\newblock Thinking in frequency: Face forgery detection by mining frequency-aware clues.
\newblock In \emph{ECCV}, volume 12357, pages 86--103, 2020.

\bibitem[Tan et~al.(2024)Tan, Liu, Zhao, Wei, Gu, Liu, and Wei]{NPR}
Chuangchuang Tan, Huan Liu, Yao Zhao, Shikui Wei, Guanghua Gu, Ping Liu, and Yunchao Wei.
\newblock Rethinking the up-sampling operations in cnn-based generative network for generalizable deepfake detection.
\newblock In \emph{CVPR}, 2024.

\bibitem[Ni et~al.(2022)Ni, Peng, Chen, Zhang, Meng, Fu, Xiang, and Ling]{XCLIP}
Bolin Ni, Houwen Peng, Minghao Chen, Songyang Zhang, Gaofeng Meng, Jianlong Fu, Shiming Xiang, and Haibin Ling.
\newblock Expanding language-image pretrained models for general video recognition.
\newblock In \emph{ECCV}, volume 13664, pages 1--18, 2022.

\bibitem[Radford et~al.(2021)Radford, Kim, Hallacy, Ramesh, Goh, Agarwal, Sastry, Askell, Mishkin, Clark, et~al.]{CLIP}
Alec Radford, Jong~Wook Kim, Chris Hallacy, Aditya Ramesh, Gabriel Goh, Sandhini Agarwal, Girish Sastry, Amanda Askell, Pamela Mishkin, Jack Clark, et~al.
\newblock Learning transferable visual models from natural language supervision.
\newblock In \emph{ICML}, pages 8748--8763, 2021.

\bibitem[Henschel et~al.(2024)Henschel, Khachatryan, Hayrapetyan, Poghosyan, Tadevosyan, Wang, Navasardyan, and Shi]{henschel2024streamingt2v}
Roberto Henschel, Levon Khachatryan, Daniil Hayrapetyan, Hayk Poghosyan, Vahram Tadevosyan, Zhangyang Wang, Shant Navasardyan, and Humphrey Shi.
\newblock Streamingt2v: Consistent, dynamic, and extendable long video generation from text.
\newblock \emph{arXiv preprint arXiv:2403.14773}, 2024.

\bibitem[Zhou et~al.(2024)Zhou, Zhou, Cheng, Feng, and Hou]{zhou2024storydiffusion}
Yupeng Zhou, Daquan Zhou, Ming-Ming Cheng, Jiashi Feng, and Qibin Hou.
\newblock Storydiffusion: Consistent self-attention for long-range image and video generation.
\newblock \emph{arXiv preprint arXiv:2405.01434}, 2024.

\bibitem[Ma et~al.(2024{\natexlab{a}})Ma, Zhou, Yeh, Wang, Li, Yang, Dong, Keutzer, and Feng]{ma2024magic}
Ze~Ma, Daquan Zhou, Chun-Hsiao Yeh, Xue-She Wang, Xiuyu Li, Huanrui Yang, Zhen Dong, Kurt Keutzer, and Jiashi Feng.
\newblock Magic-me: Identity-specific video customized diffusion.
\newblock \emph{arXiv preprint arXiv:2402.09368}, 2024{\natexlab{a}}.

\bibitem[Zhang et~al.(2024)Zhang, Li, Le, Shou, Xiong, and Sahoo]{zhang2024moonshot}
David~Junhao Zhang, Dongxu Li, Hung Le, Mike~Zheng Shou, Caiming Xiong, and Doyen Sahoo.
\newblock Moonshot: Towards controllable video generation and editing with multimodal conditions.
\newblock \emph{arXiv preprint arXiv:2401.01827}, 2024.

\bibitem[Bar-Tal et~al.(2024)Bar-Tal, Chefer, Tov, Herrmann, Paiss, Zada, Ephrat, Hur, Li, Michaeli, et~al.]{bar2024lumiere}
Omer Bar-Tal, Hila Chefer, Omer Tov, Charles Herrmann, Roni Paiss, Shiran Zada, Ariel Ephrat, Junhwa Hur, Yuanzhen Li, Tomer Michaeli, et~al.
\newblock Lumiere: A space-time diffusion model for video generation.
\newblock \emph{arXiv preprint arXiv:2401.12945}, 2024.

\bibitem[Wei et~al.(2023)Wei, Zhang, Qing, Yuan, Liu, Liu, Zhang, Zhou, and Shan]{wei2023dreamvideo}
Yujie Wei, Shiwei Zhang, Zhiwu Qing, Hangjie Yuan, Zhiheng Liu, Yu~Liu, Yingya Zhang, Jingren Zhou, and Hongming Shan.
\newblock Dreamvideo: Composing your dream videos with customized subject and motion.
\newblock \emph{arXiv preprint arXiv:2312.04433}, 2023.

\bibitem[Ho et~al.(2022)Ho, Chan, Saharia, Whang, Gao, Gritsenko, Kingma, Poole, Norouzi, Fleet, et~al.]{ho2022imagen}
Jonathan Ho, William Chan, Chitwan Saharia, Jay Whang, Ruiqi Gao, Alexey Gritsenko, Diederik~P Kingma, Ben Poole, Mohammad Norouzi, David~J Fleet, et~al.
\newblock Imagen video: High definition video generation with diffusion models.
\newblock \emph{arXiv preprint arXiv:2210.02303}, 2022.

\bibitem[Girdhar et~al.(2023)Girdhar, Singh, Brown, Duval, Azadi, Rambhatla, Shah, Yin, Parikh, and Misra]{girdhar2023emu}
Rohit Girdhar, Mannat Singh, Andrew Brown, Quentin Duval, Samaneh Azadi, Sai~Saketh Rambhatla, Akbar Shah, Xi~Yin, Devi Parikh, and Ishan Misra.
\newblock Emu video: Factorizing text-to-video generation by explicit image conditioning.
\newblock \emph{arXiv preprint arXiv:2311.10709}, 2023.

\bibitem[Feng et~al.(2023)Feng, Liu, Yu, Yao, Hui, Guo, Lin, Xue, Shi, Li, et~al.]{feng2023dreamoving}
Mengyang Feng, Jinlin Liu, Kai Yu, Yuan Yao, Zheng Hui, Xiefan Guo, Xianhui Lin, Haolan Xue, Chen Shi, Xiaowen Li, et~al.
\newblock Dreamoving: A human video generation framework based on diffusion models.
\newblock \emph{arXiv preprint arXiv:2312.05107}, 2023.

\bibitem[Xu et~al.(2023{\natexlab{c}})Xu, Zhang, Liew, Yan, Liu, Zhang, Feng, and Shou]{xu2023magicanimate}
Zhongcong Xu, Jianfeng Zhang, Jun~Hao Liew, Hanshu Yan, Jia-Wei Liu, Chenxu Zhang, Jiashi Feng, and Mike~Zheng Shou.
\newblock Magicanimate: Temporally consistent human image animation using diffusion model.
\newblock \emph{arXiv preprint arXiv:2311.16498}, 2023{\natexlab{c}}.

\bibitem[Hu et~al.(2023)Hu, Gao, Zhang, Sun, Zhang, and Bo]{hu2023animate}
Li~Hu, Xin Gao, Peng Zhang, Ke~Sun, Bang Zhang, and Liefeng Bo.
\newblock Animate anyone: Consistent and controllable image-to-video synthesis for character animation.
\newblock \emph{arXiv preprint arXiv:2311.17117}, 2023.

\bibitem[Ni et~al.(2023)Ni, Shi, Li, Huang, and Min]{ni2023conditional}
Haomiao Ni, Changhao Shi, Kai Li, Sharon~X Huang, and Martin~Renqiang Min.
\newblock Conditional image-to-video generation with latent flow diffusion models.
\newblock In \emph{CVPR}, pages 18444--18455, 2023.

\bibitem[Khachatryan et~al.(2023)Khachatryan, Movsisyan, Tadevosyan, Henschel, Wang, Navasardyan, and Shi]{khachatryan2023text2video}
Levon Khachatryan, Andranik Movsisyan, Vahram Tadevosyan, Roberto Henschel, Zhangyang Wang, Shant Navasardyan, and Humphrey Shi.
\newblock Text2video-zero: Text-to-image diffusion models are zero-shot video generators.
\newblock In \emph{ICCV}, pages 15954--15964, 2023.

\bibitem[Guo et~al.(2023{\natexlab{a}})Guo, Yang, Rao, Wang, Qiao, Lin, and Dai]{guo2023animatediff}
Yuwei Guo, Ceyuan Yang, Anyi Rao, Yaohui Wang, Yu~Qiao, Dahua Lin, and Bo~Dai.
\newblock Animatediff: Animate your personalized text-to-image diffusion models without specific tuning.
\newblock \emph{arXiv preprint arXiv:2307.04725}, 2023{\natexlab{a}}.

\bibitem[Zhang et~al.(2023{\natexlab{b}})Zhang, Xing, Zeng, Fang, and Chen]{zhang2023pia}
Yiming Zhang, Zhening Xing, Yanhong Zeng, Youqing Fang, and Kai Chen.
\newblock Pia: Your personalized image animator via plug-and-play modules in text-to-image models.
\newblock \emph{arXiv preprint arXiv:2312.13964}, 2023{\natexlab{b}}.

\bibitem[Wang et~al.(2023{\natexlab{b}})Wang, Chen, Ma, Zhou, Huang, Wang, Yang, He, Yu, Yang, et~al.]{wang2023lavie}
Yaohui Wang, Xinyuan Chen, Xin Ma, Shangchen Zhou, Ziqi Huang, Yi~Wang, Ceyuan Yang, Yinan He, Jiashuo Yu, Peiqing Yang, et~al.
\newblock Lavie: High-quality video generation with cascaded latent diffusion models.
\newblock \emph{arXiv preprint arXiv:2309.15103}, 2023{\natexlab{b}}.

\bibitem[Wang et~al.(2023{\natexlab{c}})Wang, Yuan, Zhang, Chen, Wang, Zhang, Shen, Zhao, and Zhou]{I2vgen-xl}
Xiang Wang, Hangjie Yuan, Shiwei Zhang, Dayou Chen, Jiuniu Wang, Yingya Zhang, Yujun Shen, Deli Zhao, and Jingren Zhou.
\newblock I2vgen-xl, 2023{\natexlab{c}}.
\newblock URL \url{https://modelscope.cn/models/damo/Image-to-Video/summary}.

\bibitem[Chen et~al.(2023{\natexlab{b}})Chen, Xia, He, Zhang, Cun, Yang, Xing, Liu, Chen, Wang, et~al.]{chen2023videocrafter1}
Haoxin Chen, Menghan Xia, Yingqing He, Yong Zhang, Xiaodong Cun, Shaoshu Yang, Jinbo Xing, Yaofang Liu, Qifeng Chen, Xintao Wang, et~al.
\newblock Videocrafter1: Open diffusion models for high-quality video generation.
\newblock \emph{arXiv preprint arXiv:2310.19512}, 2023{\natexlab{b}}.

\bibitem[Chen et~al.(2024)Chen, Zhang, Cun, Xia, Wang, Weng, and Shan]{chen2024videocrafter2}
Haoxin Chen, Yong Zhang, Xiaodong Cun, Menghan Xia, Xintao Wang, Chao Weng, and Ying Shan.
\newblock Videocrafter2: Overcoming data limitations for high-quality video diffusion models.
\newblock \emph{arXiv preprint arXiv:2401.09047}, 2024.

\bibitem[Xing et~al.(2023)Xing, Xia, Zhang, Chen, Wang, Wong, and Shan]{xing2023dynamicrafter}
Jinbo Xing, Menghan Xia, Yong Zhang, Haoxin Chen, Xintao Wang, Tien-Tsin Wong, and Ying Shan.
\newblock Dynamicrafter: Animating open-domain images with video diffusion priors.
\newblock \emph{arXiv preprint arXiv:2310.12190}, 2023.

\bibitem[Wang et~al.(2023{\natexlab{d}})Wang, Yuan, Chen, Zhang, Wang, and Zhang]{wang2023modelscope}
Jiuniu Wang, Hangjie Yuan, Dayou Chen, Yingya Zhang, Xiang Wang, and Shiwei Zhang.
\newblock Modelscope text-to-video technical report.
\newblock \emph{arXiv preprint arXiv:2308.06571}, 2023{\natexlab{d}}.

\bibitem[Blattmann et~al.(2023{\natexlab{b}})Blattmann, Dockhorn, Kulal, Mendelevitch, Kilian, Lorenz, Levi, English, Voleti, Letts, et~al.]{blattmann2023stable}
Andreas Blattmann, Tim Dockhorn, Sumith Kulal, Daniel Mendelevitch, Maciej Kilian, Dominik Lorenz, Yam Levi, Zion English, Vikram Voleti, Adam Letts, et~al.
\newblock Stable video diffusion: Scaling latent video diffusion models to large datasets.
\newblock \emph{arXiv preprint arXiv:2311.15127}, 2023{\natexlab{b}}.

\bibitem[Wang et~al.(2023{\natexlab{e}})Wang, Yuan, Zhang, Chen, Wang, Zhang, Shen, Zhao, and Zhou]{wang2024videocomposer}
Xiang Wang, Hangjie Yuan, Shiwei Zhang, Dayou Chen, Jiuniu Wang, Yingya Zhang, Yujun Shen, Deli Zhao, and Jingren Zhou.
\newblock Videocomposer: Compositional video synthesis with motion controllability.
\newblock \emph{NeurIPS}, 36, 2023{\natexlab{e}}.

\bibitem[Chen et~al.(2023{\natexlab{c}})Chen, Wang, Zhang, Zhuang, Ma, Yu, Wang, Lin, Qiao, and Liu]{chen2023seine}
Xinyuan Chen, Yaohui Wang, Lingjun Zhang, Shaobin Zhuang, Xin Ma, Jiashuo Yu, Yali Wang, Dahua Lin, Yu~Qiao, and Ziwei Liu.
\newblock Seine: Short-to-long video diffusion model for generative transition and prediction.
\newblock In \emph{ICLR}, 2023{\natexlab{c}}.

\bibitem[Ope(2024)]{OpenSora}
Open-sora: Democratizing efficient video production for all, 2024.
\newblock URL \url{https://github.com/hpcaitech/Open-Sora}.

\bibitem[Ma et~al.(2024{\natexlab{b}})Ma, Wang, Jia, Chen, Liu, Li, Chen, and Qiao]{ma2024latte}
Xin Ma, Yaohui Wang, Gengyun Jia, Xinyuan Chen, Ziwei Liu, Yuan-Fang Li, Cunjian Chen, and Yu~Qiao.
\newblock Latte: Latent diffusion transformer for video generation.
\newblock \emph{arXiv preprint arXiv:2401.03048}, 2024{\natexlab{b}}.

\bibitem[Hong et~al.(2022)Hong, Ding, Zheng, Liu, and Tang]{hong2022cogvideo}
Wenyi Hong, Ming Ding, Wendi Zheng, Xinghan Liu, and Jie Tang.
\newblock Cogvideo: Large-scale pretraining for text-to-video generation via transformers.
\newblock \emph{arXiv preprint arXiv:2205.15868}, 2022.

\bibitem[Li et~al.(2022)Li, Xu, Lv, Cui, Zhang, and Wei]{li2022dit}
Junlong Li, Yiheng Xu, Tengchao Lv, Lei Cui, Cha Zhang, and Furu Wei.
\newblock Dit: Self-supervised pre-training for document image transformer.
\newblock In \emph{ACM Multimedia}, pages 3530--3539, 2022.

\bibitem[Shen et~al.(2023)Shen, Li, and Elhoseiny]{shen2023mostgan}
Xiaoqian Shen, Xiang Li, and Mohamed Elhoseiny.
\newblock Mostgan-v: Video generation with temporal motion styles.
\newblock In \emph{CVPR}, pages 5652--5661, 2023.

\bibitem[Wang et~al.(2023{\natexlab{f}})Wang, Jiang, and Loy]{wang2023styleinv}
Yuhan Wang, Liming Jiang, and Chen~Change Loy.
\newblock Styleinv: A temporal style modulated inversion network for unconditional video generation.
\newblock In \emph{ICCV}, pages 22851--22861, 2023{\natexlab{f}}.

\bibitem[Kondratyuk et~al.(2023)Kondratyuk, Yu, Gu, Lezama, Huang, Hornung, Adam, Akbari, Alon, Birodkar, et~al.]{kondratyuk2023videopoet}
Dan Kondratyuk, Lijun Yu, Xiuye Gu, Jos{\'e} Lezama, Jonathan Huang, Rachel Hornung, Hartwig Adam, Hassan Akbari, Yair Alon, Vighnesh Birodkar, et~al.
\newblock Videopoet: A large language model for zero-shot video generation.
\newblock \emph{arXiv preprint arXiv:2312.14125}, 2023.

\bibitem[Yoo et~al.(2023)Yoo, Kim, Lee, Kim, and Hong]{yoo2023towards}
Jaehoon Yoo, Semin Kim, Doyup Lee, Chiheon Kim, and Seunghoon Hong.
\newblock Towards end-to-end generative modeling of long videos with memory-efficient bidirectional transformers.
\newblock In \emph{CVPR}, pages 22888--22897, 2023.

\bibitem[Lei et~al.(2023)Lei, Ding, et~al.]{lei2023flashvideo}
Bin Lei, Caiwen Ding, et~al.
\newblock Flashvideo: A framework for swift inference in text-to-video generation.
\newblock \emph{arXiv preprint arXiv:2401.00869}, 2023.

\bibitem[Yu et~al.(2023)Yu, Cheng, Sohn, Lezama, Zhang, Chang, Hauptmann, Yang, Hao, Essa, et~al.]{yu2023magvit}
Lijun Yu, Yong Cheng, Kihyuk Sohn, Jos{\'e} Lezama, Han Zhang, Huiwen Chang, Alexander~G Hauptmann, Ming-Hsuan Yang, Yuan Hao, Irfan Essa, et~al.
\newblock Magvit: Masked generative video transformer.
\newblock In \emph{CVPR}, pages 10459--10469, 2023.

\bibitem[Ghosh et~al.(2024)Ghosh, Sanyal, Schmid, and Sch{\"o}lkopf]{ghosh2024raven}
Partha Ghosh, Soubhik Sanyal, Cordelia Schmid, and Bernhard Sch{\"o}lkopf.
\newblock Raven: Rethinking adversarial video generation with efficient tri-plane networks.
\newblock \emph{arXiv preprint arXiv:2401.06035}, 2024.

\bibitem[Guo et~al.(2023{\natexlab{b}})Guo, Liu, Ren, Grosz, Masi, and Liu]{guo2023hierarchical}
Xiao Guo, Xiaohong Liu, Zhiyuan Ren, Steven Grosz, Iacopo Masi, and Xiaoming Liu.
\newblock Hierarchical fine-grained image forgery detection and localization.
\newblock In \emph{ICCV}, pages 3155--3165, 2023{\natexlab{b}}.

\bibitem[Lorenz et~al.(2023)Lorenz, Durall, and Keuper]{lorenz2023detecting}
Peter Lorenz, Ricard~L Durall, and Janis Keuper.
\newblock Detecting images generated by deep diffusion models using their local intrinsic dimensionality.
\newblock In \emph{ICCV}, pages 448--459, 2023.

\bibitem[Wu et~al.(2023)Wu, Zhou, and Zhang]{wu2023generalizable}
Haiwei Wu, Jiantao Zhou, and Shile Zhang.
\newblock Generalizable synthetic image detection via language-guided contrastive learning.
\newblock \emph{arXiv preprint arXiv:2305.13800}, 2023.

\bibitem[Wang et~al.(2023{\natexlab{g}})Wang, Bao, Zhou, Wang, Hu, Chen, and Li]{DIRE}
Zhendong Wang, Jianmin Bao, Wengang Zhou, Weilun Wang, Hezhen Hu, Hong Chen, and Houqiang Li.
\newblock {DIRE} for diffusion-generated image detection.
\newblock In \emph{ICCV}, pages 22388--22398, 2023{\natexlab{g}}.

\bibitem[Wang et~al.(2022)Wang, Montoya, Munechika, Yang, Hoover, and Chau]{wang2022diffusiondb}
Zijie~J Wang, Evan Montoya, David Munechika, Haoyang Yang, Benjamin Hoover, and Duen~Horng Chau.
\newblock Diffusiondb: A large-scale prompt gallery dataset for text-to-image generative models.
\newblock \emph{arXiv preprint arXiv:2210.14896}, 2022.

\bibitem[Zhu et~al.(2023{\natexlab{a}})Zhu, Chen, Yan, Huang, Lin, Li, Tu, Hu, Hu, and Wang]{zhu2023genimage}
Mingjian Zhu, Hanting Chen, Qiangyu Yan, Xudong Huang, Guanyu Lin, Wei Li, Zhijun Tu, Hailin Hu, Jie Hu, and Yunhe Wang.
\newblock Genimage: A million-scale benchmark for detecting ai-generated image.
\newblock \emph{arXiv preprint arXiv:2306.08571}, 2023{\natexlab{a}}.

\bibitem[Kay et~al.(2017{\natexlab{b}})Kay, Carreira, Simonyan, Zhang, Hillier, Vijayanarasimhan, Viola, Green, Back, Natsev, et~al.]{kay2017kinetics}
Will Kay, Joao Carreira, Karen Simonyan, Brian Zhang, Chloe Hillier, Sudheendra Vijayanarasimhan, Fabio Viola, Tim Green, Trevor Back, Paul Natsev, et~al.
\newblock The kinetics human action video dataset.
\newblock \emph{arXiv preprint arXiv:1705.06950}, 2017{\natexlab{b}}.

\bibitem[Xu et~al.(2016{\natexlab{b}})Xu, Mei, Yao, and Rui]{xu2016msr}
Jun Xu, Tao Mei, Ting Yao, and Yong Rui.
\newblock Msr-vtt: A large video description dataset for bridging video and language.
\newblock In \emph{CVPR}, pages 5288--5296, 2016{\natexlab{b}}.

\bibitem[zer(2024)]{zeroscopexl}
Zeroscope-v2-xl, 2024.
\newblock URL \url{https://huggingface.co/cerspense/zeroscope_v2_XL}.

\bibitem[pik(2022)]{pika}
Pika art, 2022.
\newblock URL \url{https://pika.art/}.

\bibitem[Mor(2023)]{Morphstudio}
Morph studio, 2023.
\newblock URL \url{https : / / www.morphstudio.com/}.

\bibitem[moonvalley.ai(2022)]{moonvalley}
moonvalley.ai.
\newblock moonvalley.ai, 2022.
\newblock URL \url{https://moonvalley.ai/}.

\bibitem[Hot(2023)]{Hotshot}
Hotshot, 2023.
\newblock URL \url{https://huggingface.co/hotshotco/Hotshot-XL}.

\bibitem[Zhang et~al.(2023{\natexlab{c}})Zhang, Wu, Liu, Zhao, Ran, Gu, Gao, and Shou]{zhang2023show}
David~Junhao Zhang, Jay~Zhangjie Wu, Jia-Wei Liu, Rui Zhao, Lingmin Ran, Yuchao Gu, Difei Gao, and Mike~Zheng Shou.
\newblock Show-1: Marrying pixel and latent diffusion models for text-to-video generation.
\newblock \emph{arXiv preprint arXiv:2309.15818}, 2023{\natexlab{c}}.

\bibitem[Esser et~al.(2023)Esser, Chiu, Atighehchian, Granskog, and Germanidis]{esser2023structure}
Patrick Esser, Johnathan Chiu, Parmida Atighehchian, Jonathan Granskog, and Anastasis Germanidis.
\newblock Structure and content-guided video synthesis with diffusion models.
\newblock In \emph{ICCV}, pages 7346--7356, 2023.

\bibitem[Wang et~al.(2020)Wang, Wang, Zhang, Owens, and Efros]{CNNDet}
Sheng{-}Yu Wang, Oliver Wang, Richard Zhang, Andrew Owens, and Alexei~A. Efros.
\newblock Cnn-generated images are surprisingly easy to spot... for now.
\newblock In \emph{CVPR}, pages 8692--8701, 2020.

\bibitem[Zhu et~al.(2023{\natexlab{b}})Zhu, Chen, Yan, Huang, Lin, Li, Tu, Hu, Hu, and Wang]{GenImage}
Mingjian Zhu, Hanting Chen, Qiangyu Yan, Xudong Huang, Guanyu Lin, Wei Li, Zhijun Tu, Hailin Hu, Jie Hu, and Yunhe Wang.
\newblock Genimage: {A} million-scale benchmark for detecting ai-generated image.
\newblock In \emph{NeurIPS}, 2023{\natexlab{b}}.

\bibitem[Liu et~al.(2023{\natexlab{b}})Liu, Cun, Liu, Wang, Zhang, Chen, Liu, Zeng, Chan, and Shan]{liu2023evalcrafter}
Yaofang Liu, Xiaodong Cun, Xuebo Liu, Xintao Wang, Yong Zhang, Haoxin Chen, Yang Liu, Tieyong Zeng, Raymond Chan, and Ying Shan.
\newblock Evalcrafter: Benchmarking and evaluating large video generation models.
\newblock \emph{arXiv preprint arXiv:2310.11440}, 2023{\natexlab{b}}.

\bibitem[Yang et~al.(2024)Yang, Hou, Huang, Ma, Wan, Zhang, Chen, and Liao]{yang2024direct}
Shiyuan Yang, Liang Hou, Haibin Huang, Chongyang Ma, Pengfei Wan, Di~Zhang, Xiaodong Chen, and Jing Liao.
\newblock Direct-a-video: Customized video generation with user-directed camera movement and object motion.
\newblock \emph{arXiv preprint arXiv:2402.03162}, 2024.

\bibitem[Ren et~al.(2024)Ren, Yang, Zhang, Wei, Du, Huang, and Chen]{ren2024consisti2v}
Weiming Ren, Harry Yang, Ge~Zhang, Cong Wei, Xinrun Du, Stephen Huang, and Wenhu Chen.
\newblock Consisti2v: Enhancing visual consistency for image-to-video generation.
\newblock \emph{arXiv preprint arXiv:2402.04324}, 2024.

\bibitem[Wang et~al.(2023{\natexlab{h}})Wang, Zhang, Yuan, Qing, Gong, Zhang, Shen, Gao, and Sang]{wang2023recipe}
Xiang Wang, Shiwei Zhang, Hangjie Yuan, Zhiwu Qing, Biao Gong, Yingya Zhang, Yujun Shen, Changxin Gao, and Nong Sang.
\newblock A recipe for scaling up text-to-video generation with text-free videos.
\newblock \emph{arXiv preprint arXiv:2312.15770}, 2023{\natexlab{h}}.

\bibitem[Qing et~al.(2023)Qing, Zhang, Wang, Wang, Wei, Zhang, Gao, and Sang]{qing2023hierarchical}
Zhiwu Qing, Shiwei Zhang, Jiayu Wang, Xiang Wang, Yujie Wei, Yingya Zhang, Changxin Gao, and Nong Sang.
\newblock Hierarchical spatio-temporal decoupling for text-to-video generation.
\newblock \emph{arXiv preprint arXiv:2312.04483}, 2023.

\bibitem[Ge et~al.(2023)Ge, Nah, Liu, Poon, Tao, Catanzaro, Jacobs, Huang, Liu, and Balaji]{ge2023preserve}
Songwei Ge, Seungjun Nah, Guilin Liu, Tyler Poon, Andrew Tao, Bryan Catanzaro, David Jacobs, Jia-Bin Huang, Ming-Yu Liu, and Yogesh Balaji.
\newblock Preserve your own correlation: A noise prior for video diffusion models.
\newblock In \emph{ICCV}, pages 22930--22941, 2023.

\bibitem[Guo et~al.(2023{\natexlab{c}})Guo, Zheng, Hou, Gao, Deng, Ma, Hu, Zha, Huang, Wan, et~al.]{guo2023i2v}
Xun Guo, Mingwu Zheng, Liang Hou, Yuan Gao, Yufan Deng, Chongyang Ma, Weiming Hu, Zhengjun Zha, Haibin Huang, Pengfei Wan, et~al.
\newblock I2v-adapter: A general image-to-video adapter for video diffusion models.
\newblock \emph{arXiv preprint arXiv:2312.16693}, 2023{\natexlab{c}}.

\bibitem[Tian et~al.(2024)Tian, Wang, Zhang, and Bo]{tian2024emo}
Linrui Tian, Qi~Wang, Bang Zhang, and Liefeng Bo.
\newblock Emo: Emote portrait alive-generating expressive portrait videos with audio2video diffusion model under weak conditions.
\newblock \emph{arXiv preprint arXiv:2402.17485}, 2024.

\bibitem[Zeng et~al.(2023)Zeng, Wei, Zheng, Zou, Wei, Zhang, and Li]{zeng2023make}
Yan Zeng, Guoqiang Wei, Jiani Zheng, Jiaxin Zou, Yang Wei, Yuchen Zhang, and Hang Li.
\newblock Make pixels dance: High-dynamic video generation.
\newblock \emph{arXiv preprint arXiv:2311.10982}, 2023.

\bibitem[Xu et~al.(2023{\natexlab{d}})Xu, Ye, Wu, Yan, Miao, Ye, Xu, Hu, Shi, Xu, et~al.]{xu2023youku}
Haiyang Xu, Qinghao Ye, Xuan Wu, Ming Yan, Yuan Miao, Jiabo Ye, Guohai Xu, Anwen Hu, Yaya Shi, Guangwei Xu, et~al.
\newblock Youku-mplug: A 10 million large-scale chinese video-language dataset for pre-training and benchmarks.
\newblock \emph{arXiv preprint arXiv:2306.04362}, 2023{\natexlab{d}}.

\bibitem[Rombach et~al.(2022)Rombach, Blattmann, Lorenz, Esser, and Ommer]{rombach2022high}
Robin Rombach, Andreas Blattmann, Dominik Lorenz, Patrick Esser, and Bj{\"o}rn Ommer.
\newblock High-resolution image synthesis with latent diffusion models.
\newblock In \emph{CVPR}, pages 10684--10695, 2022.

\bibitem[Podell et~al.(2023)Podell, English, Lacey, Blattmann, Dockhorn, M{\"u}ller, Penna, and Rombach]{podell2023sdxl}
Dustin Podell, Zion English, Kyle Lacey, Andreas Blattmann, Tim Dockhorn, Jonas M{\"u}ller, Joe Penna, and Robin Rombach.
\newblock Sdxl: Improving latent diffusion models for high-resolution image synthesis.
\newblock \emph{arXiv preprint arXiv:2307.01952}, 2023.

\bibitem[Le~Scao et~al.(2023)Le~Scao, Fan, Akiki, Pavlick, Ili{\'c}, Hesslow, Castagn{\'e}, Luccioni, Yvon, Gall{\'e}, et~al.]{le2023bloom}
Teven Le~Scao, Angela Fan, Christopher Akiki, Ellie Pavlick, Suzana Ili{\'c}, Daniel Hesslow, Roman Castagn{\'e}, Alexandra~Sasha Luccioni, Fran{\c{c}}ois Yvon, Matthias Gall{\'e}, et~al.
\newblock Bloom: A 176b-parameter open-access multilingual language model.
\newblock 2023.

\bibitem[Huang et~al.(2024)Huang, He, Yu, Zhang, Si, Jiang, Zhang, Wu, Jin, Chanpaisit, et~al.]{huang2024vbench}
Ziqi Huang, Yinan He, Jiashuo Yu, Fan Zhang, Chenyang Si, Yuming Jiang, Yuanhan Zhang, Tianxing Wu, Qingyang Jin, Nattapol Chanpaisit, et~al.
\newblock Vbench: Comprehensive benchmark suite for video generative models.
\newblock In \emph{Proceedings of the IEEE/CVF Conference on Computer Vision and Pattern Recognition}, pages 21807--21818, 2024.

\bibitem[Gupta et~al.(2023)Gupta, Yu, Sohn, Gu, Hahn, Fei-Fei, Essa, Jiang, and Lezama]{gupta2023photorealistic}
Agrim Gupta, Lijun Yu, Kihyuk Sohn, Xiuye Gu, Meera Hahn, Li~Fei-Fei, Irfan Essa, Lu~Jiang, and Jos{\'e} Lezama.
\newblock Photorealistic video generation with diffusion models.
\newblock \emph{arXiv preprint arXiv:2312.06662}, 2023.

\bibitem[Gu et~al.(2022{\natexlab{b}})Gu, Goel, and R{\'{e}}]{Gu_ICLR}
Albert Gu, Karan Goel, and Christopher R{\'{e}}.
\newblock Efficiently modeling long sequences with structured state spaces.
\newblock In \emph{ICLR}, 2022{\natexlab{b}}.

\bibitem[Gu et~al.(2021{\natexlab{b}})Gu, Johnson, Goel, Saab, Dao, Rudra, and R{\'{e}}]{rcc_lssl}
Albert Gu, Isys Johnson, Karan Goel, Khaled Saab, Tri Dao, Atri Rudra, and Christopher R{\'{e}}.
\newblock Combining recurrent, convolutional, and continuous-time models with linear state space layers.
\newblock In \emph{NeurIPS}, pages 572--585, 2021{\natexlab{b}}.

\bibitem[Smith et~al.(2023)Smith, Warrington, and Linderman]{SSSL}
Jimmy T.~H. Smith, Andrew Warrington, and Scott~W. Linderman.
\newblock Simplified state space layers for sequence modeling.
\newblock In \emph{ICLR}, 2023.

\bibitem[Gu and Dao(2023)]{gu2023mamba}
Albert Gu and Tri Dao.
\newblock Mamba: Linear-time sequence modeling with selective state spaces.
\newblock \emph{arXiv preprint arXiv:2312.00752}, 2023.

\bibitem[Zhu et~al.(2024)Zhu, Liao, Zhang, Wang, Liu, and Wang]{VisionMamba}
Lianghui Zhu, Bencheng Liao, Qian Zhang, Xinlong Wang, Wenyu Liu, and Xinggang Wang.
\newblock Vision mamba: Efficient visual representation learning with bidirectional state space model.
\newblock \emph{arXiv preprint arXiv:2401.09417}, 2024.

\bibitem[Liu et~al.(2024)Liu, Tian, Zhao, Yu, Xie, Wang, Ye, and Liu]{VMamba}
Yue Liu, Yunjie Tian, Yuzhong Zhao, Hongtian Yu, Lingxi Xie, Yaowei Wang, Qixiang Ye, and Yunfan Liu.
\newblock Vmamba: Visual state space model.
\newblock \emph{arXiv preprint arXiv:2401.10166}, 2024.

\bibitem[Tong et~al.(2022)Tong, Song, Wang, and Wang]{videomae}
Zhan Tong, Yibing Song, Jue Wang, and Limin Wang.
\newblock Videomae: Masked autoencoders are data-efficient learners for self-supervised video pre-training.
\newblock In \emph{NeurIPS}, 2022.

\bibitem[Coccomini et~al.(2024)Coccomini, Kordopatis{-}Zilos, Amato, Caldelli, Falchi, Papadopoulos, and Gennaro]{MINTIME}
Davide~Alessandro Coccomini, Giorgos Kordopatis{-}Zilos, Giuseppe Amato, Roberto Caldelli, Fabrizio Falchi, Symeon Papadopoulos, and Claudio Gennaro.
\newblock {MINTIME:} multi-identity size-invariant video deepfake detection.
\newblock \emph{{IEEE} Trans. Inf. Forensics Secur.}, 19:\penalty0 6084--6096, 2024.

\bibitem[Zheng et~al.(2021)Zheng, Bao, Chen, Zeng, and Wen]{FTCN}
Yinglin Zheng, Jianmin Bao, Dong Chen, Ming Zeng, and Fang Wen.
\newblock Exploring temporal coherence for more general video face forgery detection.
\newblock In \emph{ICCV}, pages 15024--15034, 2021.

\bibitem[Corvi et~al.(2023)Corvi, Cozzolino, Zingarini, Poggi, Nagano, and Verdoliva]{On_Det}
Riccardo Corvi, Davide Cozzolino, Giada Zingarini, Giovanni Poggi, Koki Nagano, and Luisa Verdoliva.
\newblock On the detection of synthetic images generated by diffusion models.
\newblock In \emph{ICASSP}, pages 1--5, 2023.

\end{thebibliography}

\newpage
\appendix
\section*{\Large Appendix}
\section{Model details}
We provide detailed information about the methods used in this paper, as shown in Table~\ref{tab:model_detail}. Our model only requires a small addition of parameters on the XCLIP-B model to achieve significant performance improvements.

\begin{table*}[h]
    \centering
    \caption{Model details and its performance on many-to-many generalization task.}
    \label{tab:model_detail}
    \resizebox{\textwidth}{!}{
    \begin{tabular}{cccccccc}
    \toprule
    Model & Detection level&Input size & Param (M) & FLOPs(G)&  R&F1 & AP\\
    \midrule
    CLIP-B&Image&$8\times 224\times224$&151.46&149.34
    &0.9057& 0.7631&0.9152\\
    NPR&Image&$8\times 224\times224$&\textbf{1.44}&\textbf{14.08}
    & 0.8408& 0.7164&0.8245\\
    F3Net&Image&$8\times 299\times299$&48.31&145.04
    & 0.8188&0.8024&0.8873\\
    VideoMAE-B&Video&$16\times 224\times224$&86.54&147.52 &0.8750& 0.8955& 0.9454
    \\
    STIL&Video&$8\times 224\times224$&22.69&38.06&0.8222&0.7823&0.8712
    \\
    MINTIME-CLIP-B&Video&$8\times 224\times224$&201.35&221.0& 0.8762& 0.8231& 0.9155\\
    FTCN-CLIP-B&Video&$8\times 224\times224$&174.82&146.69& 0.8618&0.8755&0.9221\\
    TALL&Video&$8\times 224\times224$&86.92&15.2& 0.8852&0.7419&0.8791\\
    XCLIP-B&Video&$8\times 224\times224$&121.26&141.28 &  0.6615&0.6497&0.7154
    \\
    DeMamba-XCLIP&Video&$8\times 224\times224$&125.37&147.61& \textbf{0.9392} &\textbf{0.9020} & \textbf{0.9710} 
    \\
    \bottomrule
    \end{tabular}}
\end{table*}

\section{Experiment details}
\subsection{Implementation details}
\label{A3_tr}

\textbf{Data pre-processing.} For each video, we uniformly sample frames for alignment. For videos longer than 3 seconds, we sample 2 frames every second. For videos shorter than 3 seconds, our sampling frequency is $\frac{8}{\rm length}$ seconds. The pseudo-code for Pytorch-like is as follows:
\begin{lstlisting}
    video_length = get_video_length(video_path)
    os.makedirs(os.path.dirname(image_path), exist_ok=True)
    if video_length >= 3 :
        inter_val = 2
        os.system(f"cd {image_path} | ffmpeg -loglevel quiet -i {video_path} 
        -r {inter_val} {image_path}%d.jpg")
    else:
        inter_val = math.ceil(8 / video_length)
        os.system(f"cd {image_path} | ffmpeg -loglevel quiet -i {video_path} 
        -r {inter_val} {image_path}%d.jpg")
\end{lstlisting}

\textbf{Dataset augmentation.} During training and testing, we randomly select 8 or 16 consecutive frames from the video after frame sampling, and resize each frame to $224\times 224$. To enhance generalizability of models, we introduced random data augmentation during training, including HorizontalFlip, ImageCompression, GaussNoise, GaussianBlur, and Grayscale. The pseudo-code for Pytorch-like is as follows:
\begin{lstlisting}
    aug_list  = [albumentations.Resize(224, 224)]
    if random.random() < 0.5:
        aug_list.append(albumentations.HorizontalFlip(p=1.0))
    if random.random() < 0.5:
        quality_score = random.randint(50, 100)
        aug_list.append(albumentations.ImageCompression(
        quality_lower=quality_score, quality_upper=quality_score))
    if random.random() < 0.3:
        aug_list.append(albumentations.GaussNoise(p=1.0))
    if random.random() < 0.3:
        aug_list.append(albumentations.GaussianBlur(blur_limit=(3, 5), p=1.0))
    if random.random() < 0.001:
        aug_list.append(albumentations.ToGray(p=1.0))
    aug_list.append(albumentations.Normalize(mean=(0.485, 0.456, 0.406), 
    std=(0.229, 0.224, 0.225), max_pixel_value=255.0, p=1.0))
    trans = albumentations.Compose(aug_list)
\end{lstlisting}

\subsection{Details in degraded video classification task}
We applied the following transformations to the videos in $D_{\rm v-ood}$ and utilized models trained on many-to-many tasks for testing. Here, we provide the specific implementation details for each task of degraded video classification:
\label{A3_de}
\begin{itemize}
    \item[(1)] H.264 compression: H.264, also known as Advanced Video Coding (AVC), is a widely used video compression standard. In this paper, we set the crf to 28 to compress the video.
    \item[(2)] JPEG compression: JPEG compression is a widely used image compression standard designed for efficient compression of digital images. The JPEG algorithm is based on the characteristics of the human visual system, taking advantage of the insensitivity of human eyes to the loss of image details, thus achieving lossy compression of data. In this paper, we set the quality to 35 for the degradation experiment.
    \item[(3)] FLIP: We randomly select either Horizontal Flip or Vertical Flip with equal probability for the degradation experiment.
    \item[(4)] Crop: We randomly crop the video from the original video with a scale of 71\% to 93\%.
    \item[(5)] Text watermark: We randomly add textual watermarks on the random position of videos.
    \item[(6)] Image watermark: We randomly added visual watermarks on the random position of videos.
    \item[(7)] Gaussian noise: We add Gaussian blur to the video with a setting of $\sigma=7$.
    \item[(8)] Color transform: We randomly select one color transformation from brightness, contrast, saturation, hue and set the parameter to 0.5.
\end{itemize}

\section{Visualization of dataset}
As shown in Figure 3-22, we provide visualizations of samples from the dataset. From these figures, it can be seen that our dataset possesses diverse content.

\begin{figure*}[h]
	\centering
	\includegraphics[width=0.99\linewidth]{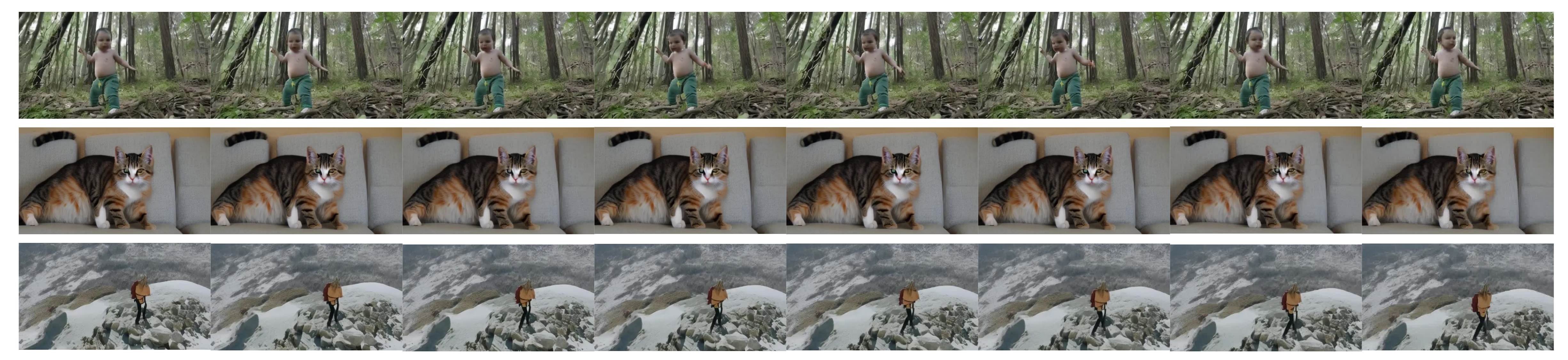}
	\caption{ZeroScope~\citep{wang2023modelscope} generated samples visualization.}
\end{figure*}

\begin{figure*}[h]
	\centering
	\includegraphics[width=0.99\linewidth]{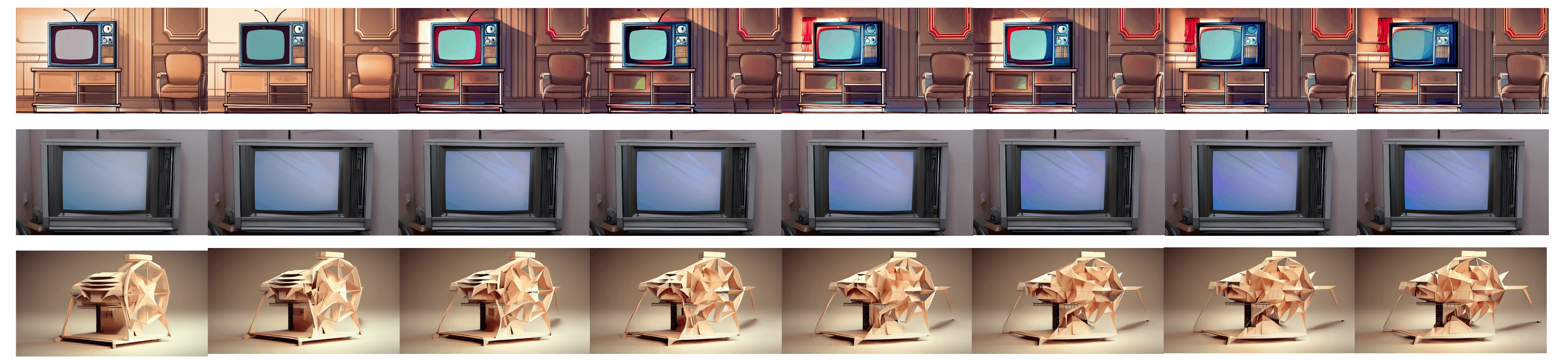}
	\caption{I2VGen-XL~\citep{I2vgen-xl} generated samples visualization.}
\end{figure*}

\begin{figure*}[h]
	\centering
	\includegraphics[width=0.99\linewidth]{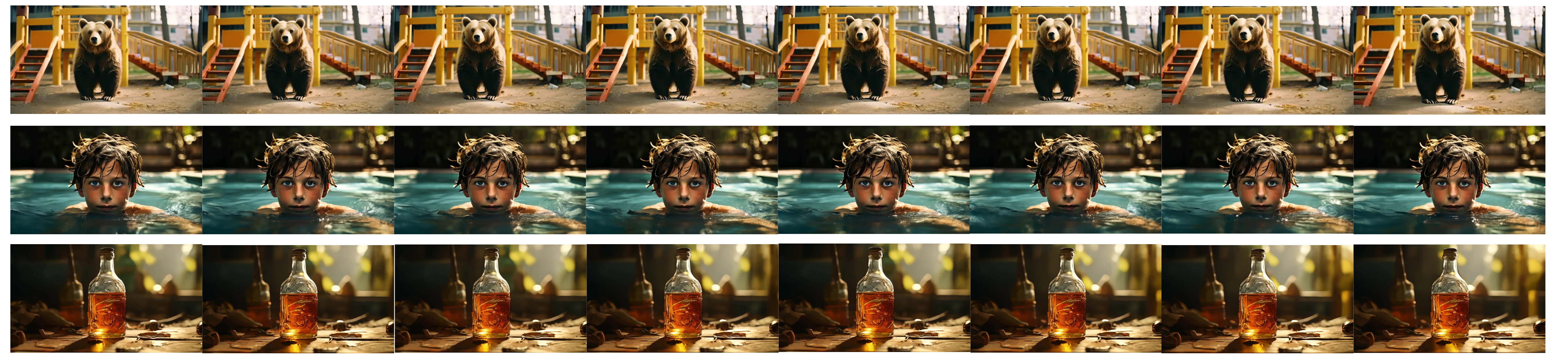}
	\caption{SVD~\citep{blattmann2023stable} generated samples visualization example.}
\end{figure*}

\begin{figure*}[h]
	\centering
	\includegraphics[width=0.99\linewidth]{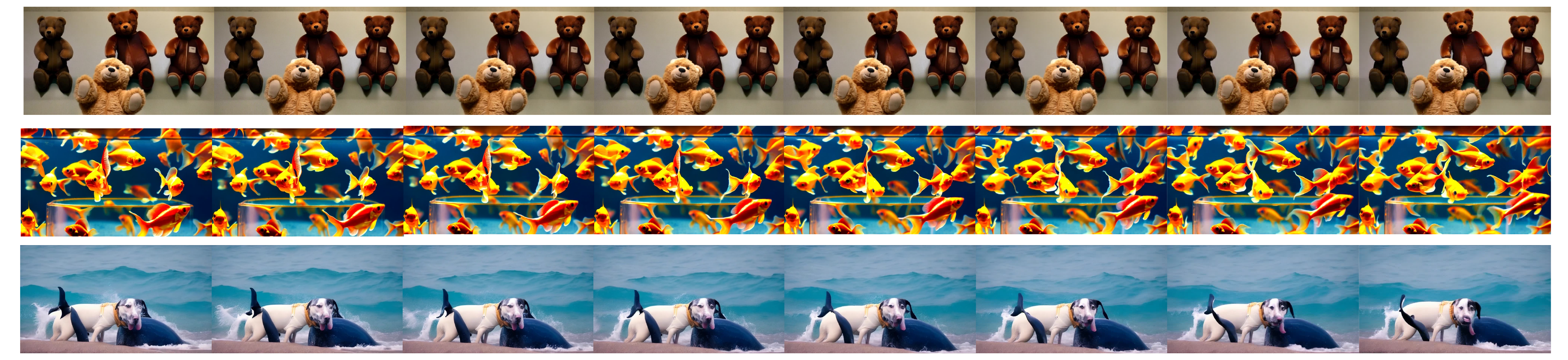}
	\caption{VideoCrafter~\citep{chen2024videocrafter2} generated samples visualization.}
\end{figure*}

\begin{figure*}[h]
	\centering
	\includegraphics[width=0.99\linewidth]{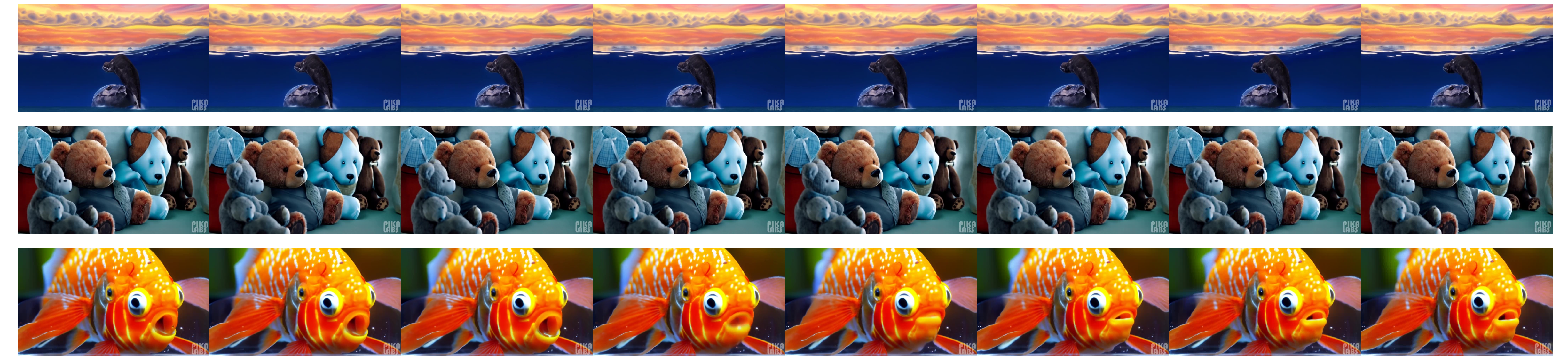}
	\caption{Pika~\citep{pika} generated samples visualization.}
\end{figure*}

\begin{figure*}[h]
	\centering
	\includegraphics[width=0.99\linewidth]{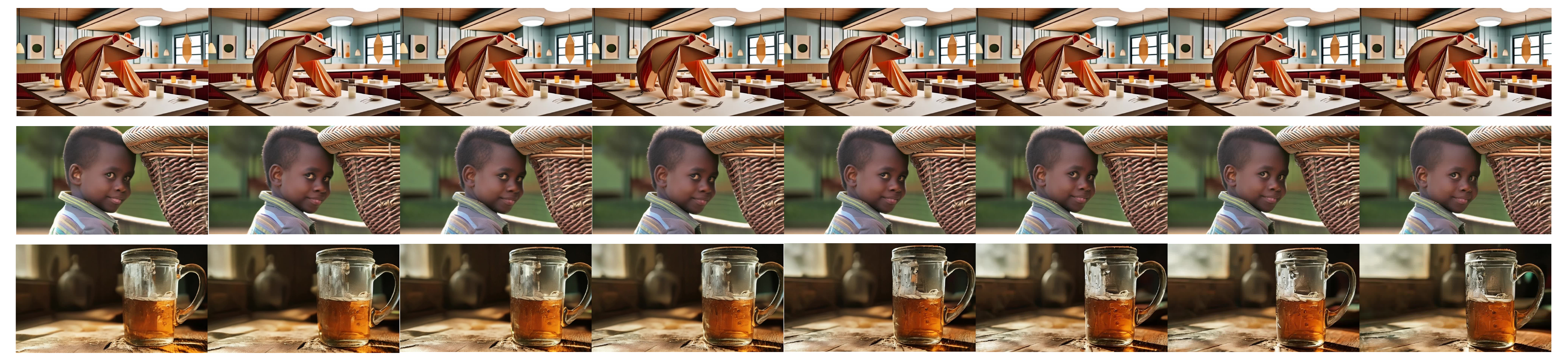}
	\caption{DynamicCrafter~\citep{xing2023dynamicrafter} generated samples visualization.}
\end{figure*}

\begin{figure*}[h]
	\centering
	\includegraphics[width=0.99\linewidth]{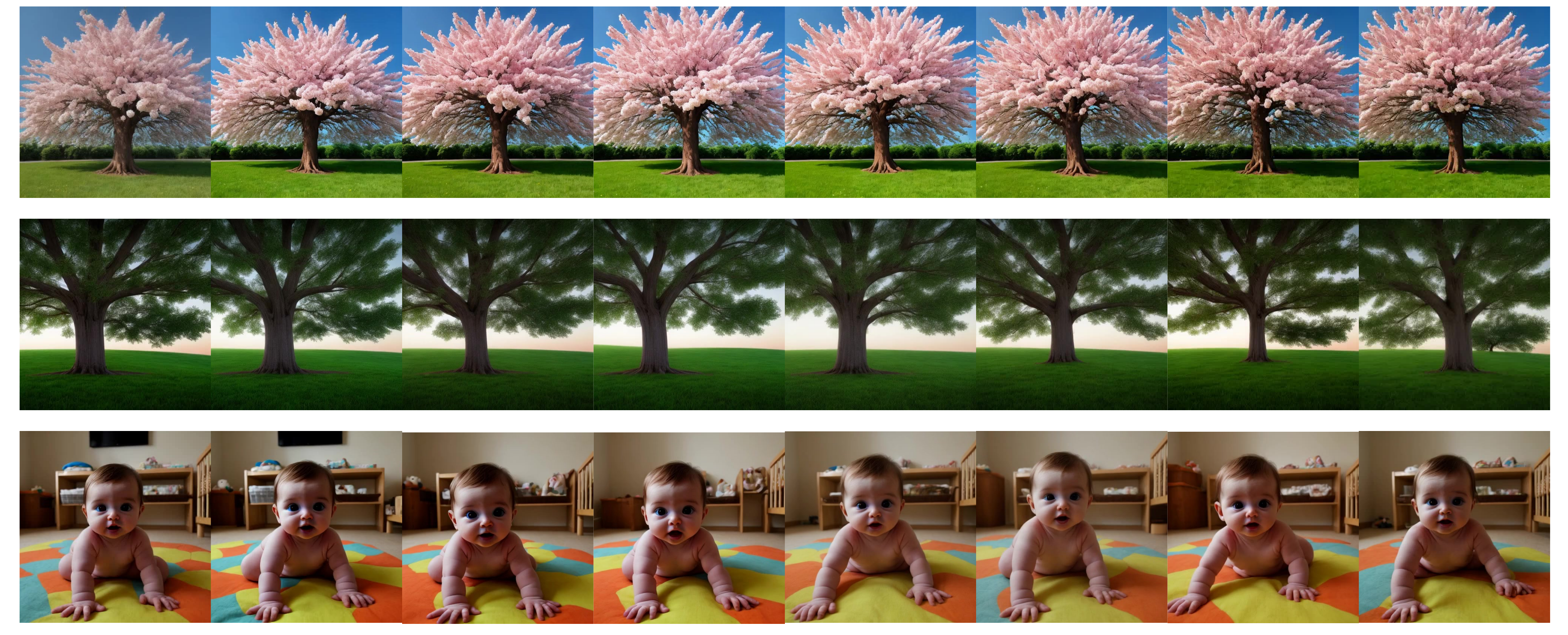}
	\caption{SD~\cite{zhang2023pia} generated samples visualization.}
\end{figure*}

\begin{figure*}[h]
	\centering
	\includegraphics[width=0.99\linewidth]{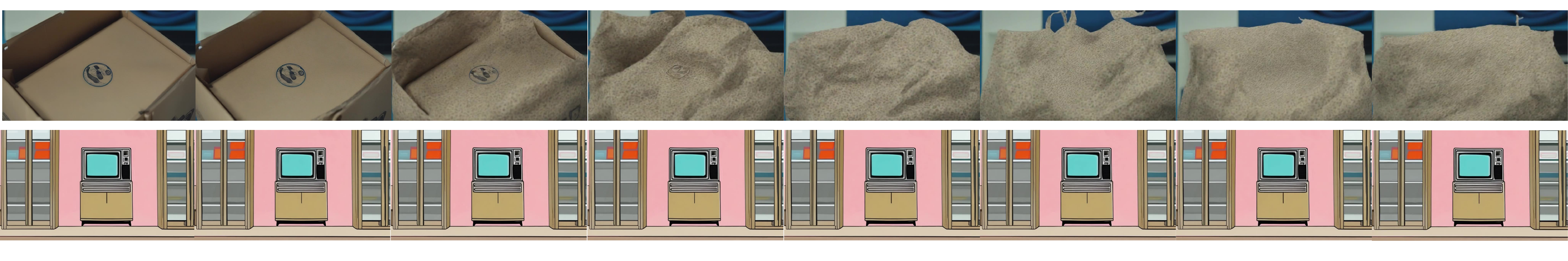}
	\caption{SEINE~\citep{chen2023seine} generated samples visualization.}
\end{figure*}

\begin{figure*}[h]
	\centering
	\includegraphics[width=0.99\linewidth]{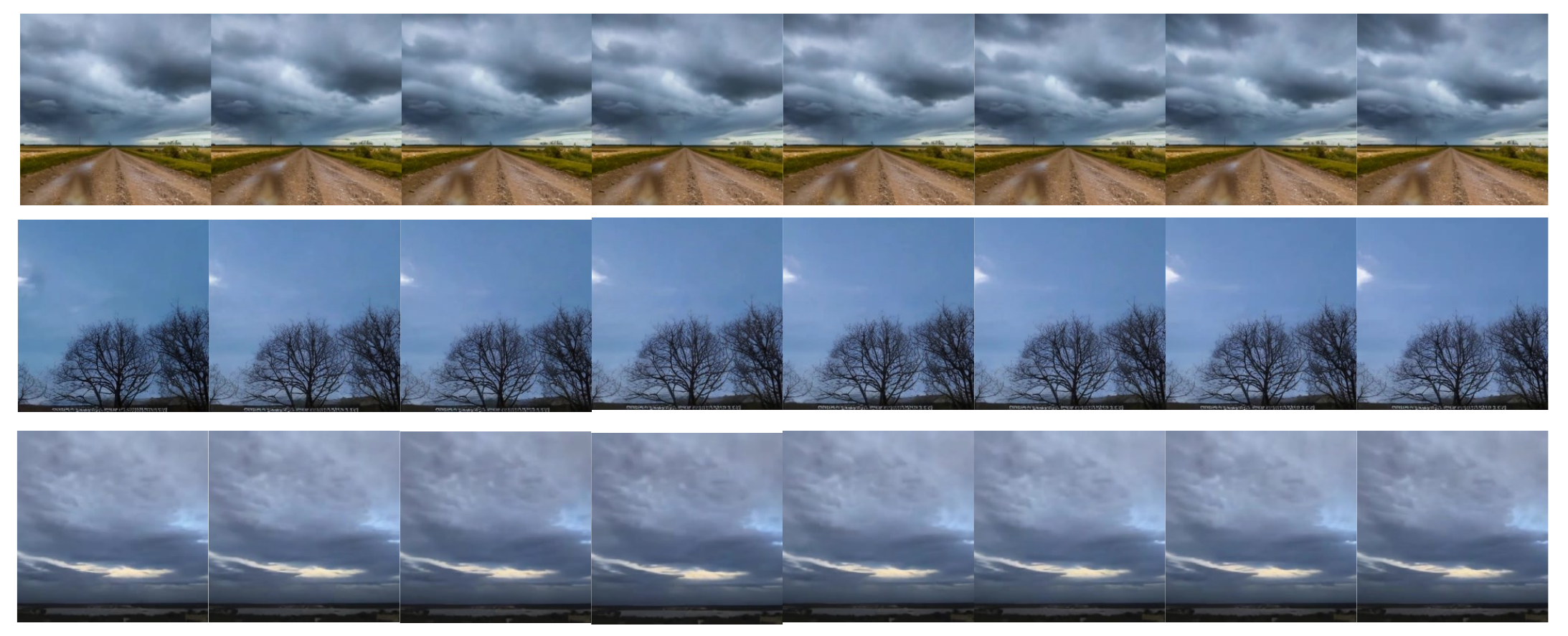}
	\caption{Latte~\citep{ma2024latte} generated samples visualization.}
\end{figure*}


\begin{figure*}[h]
	\centering
	\includegraphics[width=0.99\linewidth]{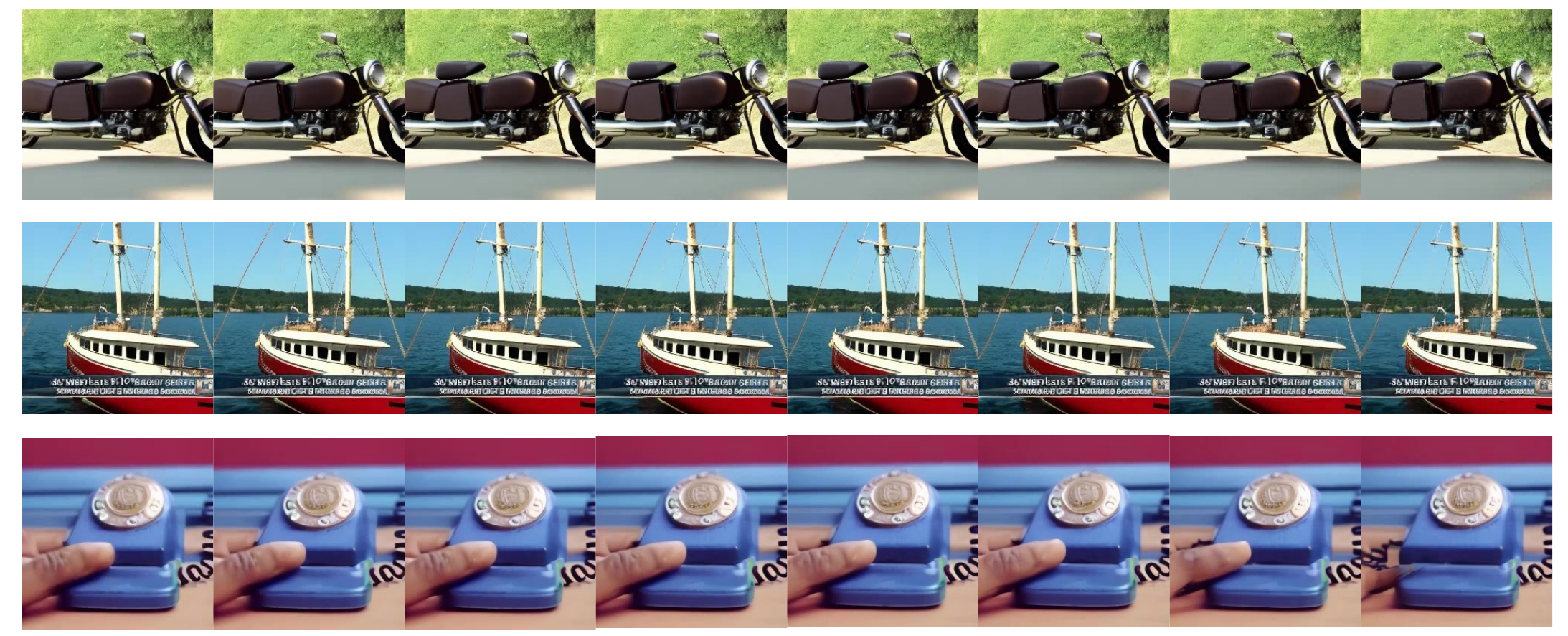}
	\caption{OpenSora~\citep{OpenSora} generated samples visualization.}
\end{figure*}
\begin{figure*}[h]
	\centering
	\includegraphics[width=0.99\linewidth]{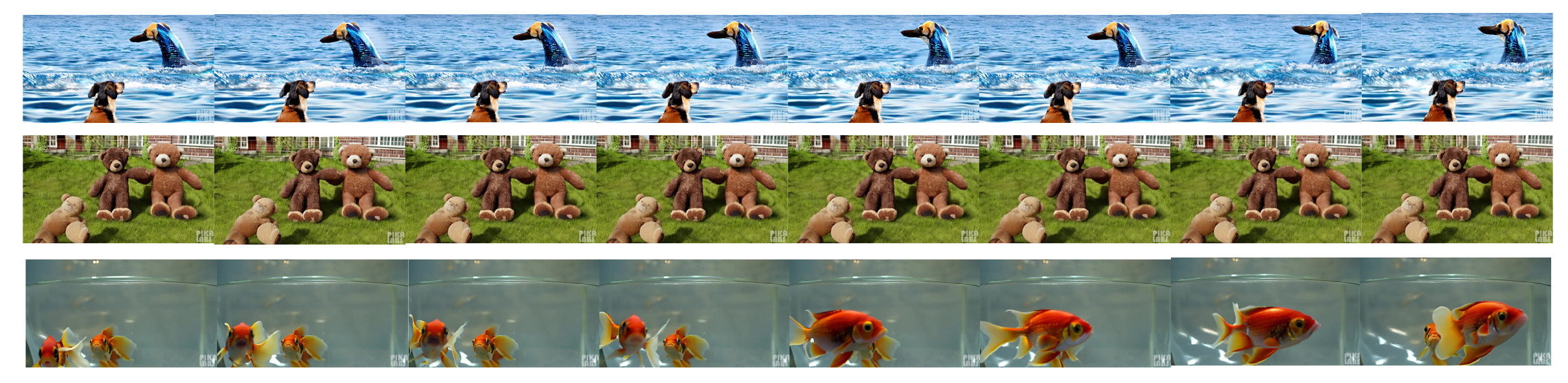}
	\caption{ModelScope~\citep{wang2023modelscope} generated samples visualization.}
\end{figure*}

\begin{figure*}[h]
	\centering
	\includegraphics[width=0.99\linewidth]{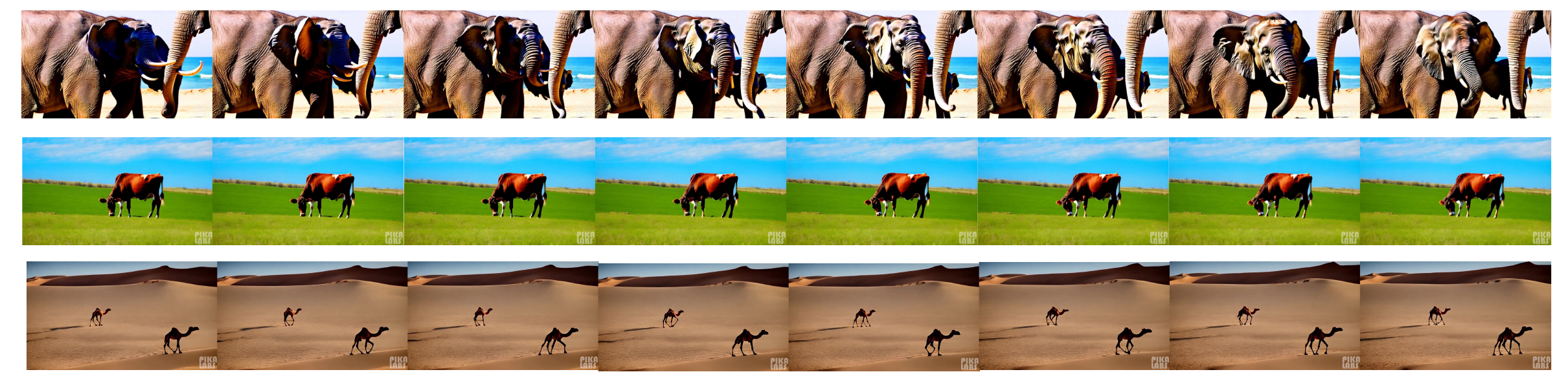}
	\caption{MorphStudio~\citep{Morphstudio} generated samples visualization.}
\end{figure*}

\begin{figure*}[h]
	\centering
	\includegraphics[width=0.99\linewidth]{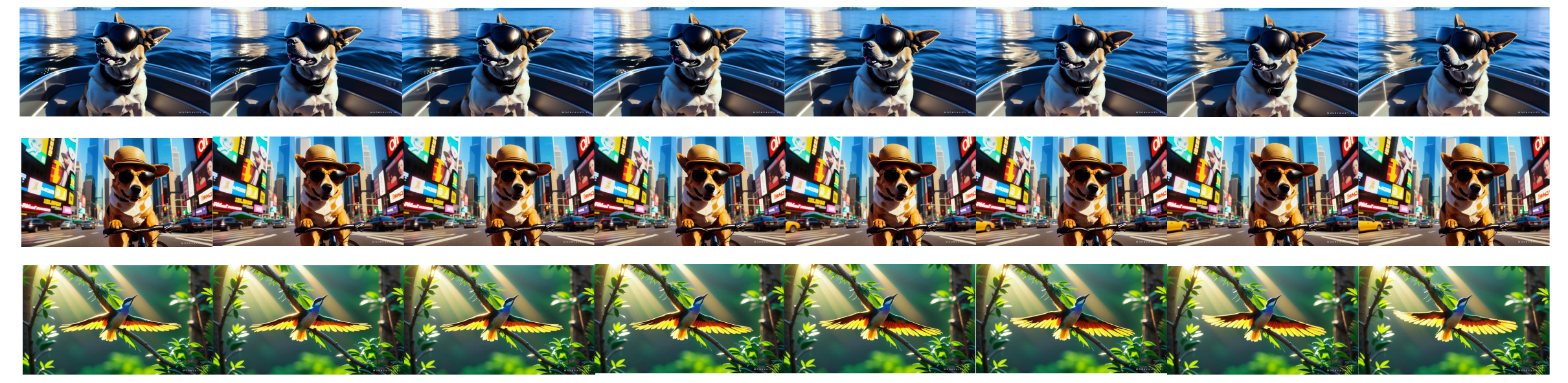}
	\caption{MoonValley~\citep{moonvalley} generated samples visualization.}
\end{figure*}

\begin{figure*}[h]
	\centering
	\includegraphics[width=0.99\linewidth]{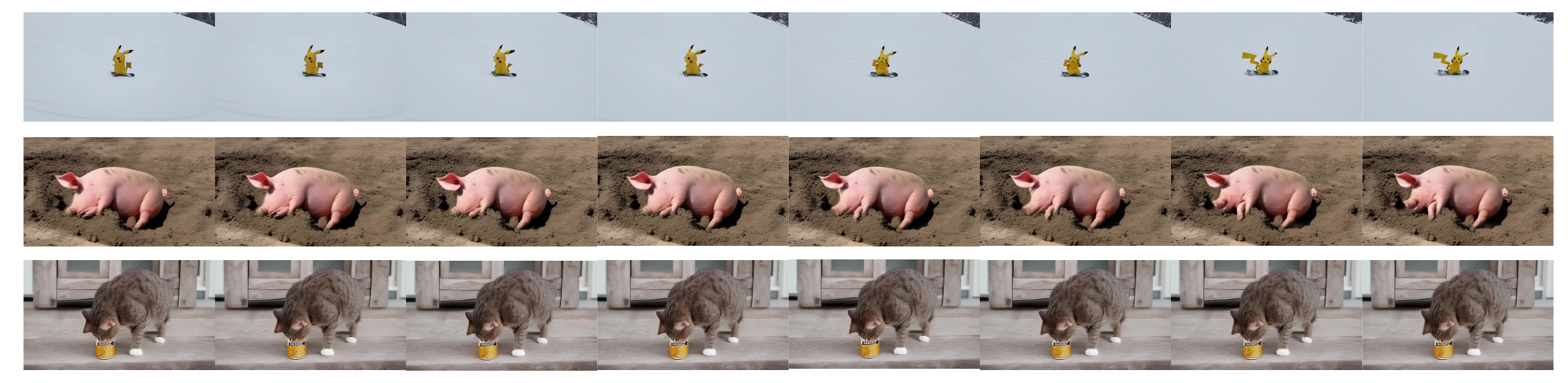}
	\caption{HotShot~\citep{Hotshot} generated samples visualization.}
\end{figure*}

\begin{figure*}[h]
	\centering
	\includegraphics[width=0.99\linewidth]{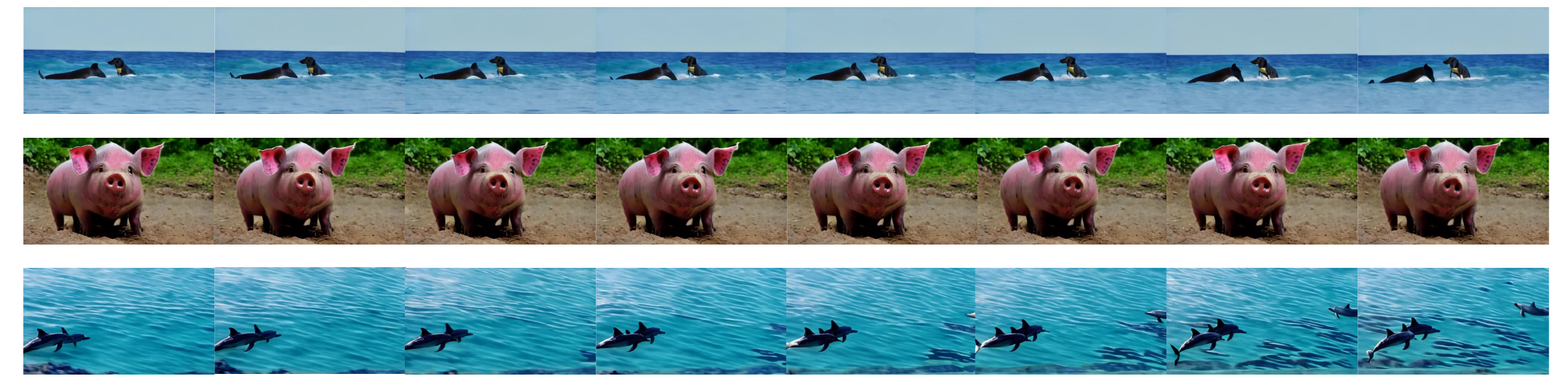}
	\caption{Show\_1~\citep{zhang2023show} generated samples visualization.}
\end{figure*}

\begin{figure*}[h]
	\centering
	\includegraphics[width=0.99\linewidth]{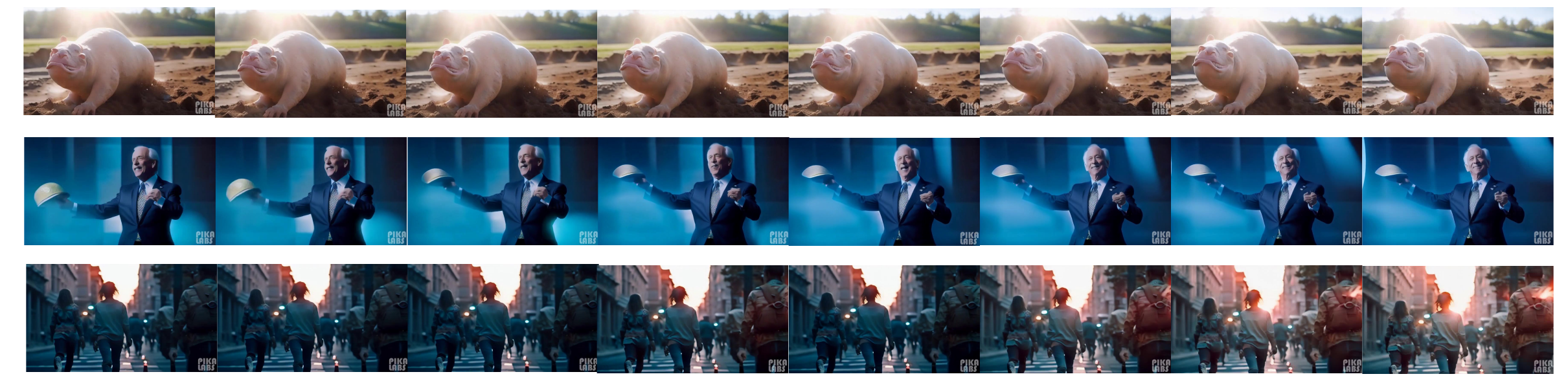}
	\caption{Gen2~\citep{esser2023structure} generated samples visualization.}
\end{figure*}

\begin{figure*}[h]
	\centering
	\includegraphics[width=0.99\linewidth]{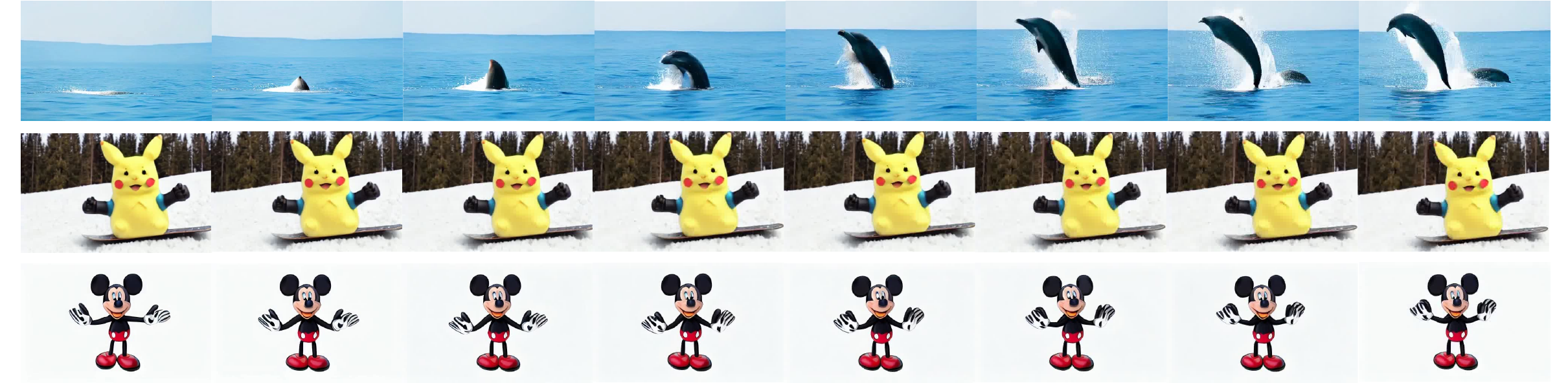}
	\caption{Lavie~\citep{Lavie} generated samples visualization.}
\end{figure*}
\begin{figure*}[h]
	\centering
	\includegraphics[width=0.99\linewidth]{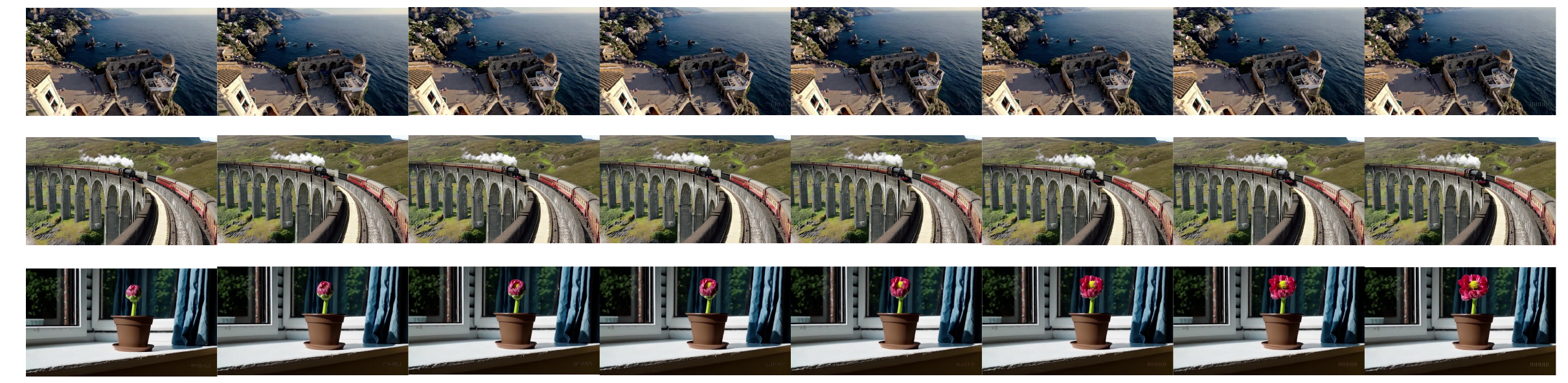}
	\caption{Sora~\citep{sora} generated samples visualization.}
\end{figure*}
\begin{figure*}[h]
	\centering
	\includegraphics[width=0.99\linewidth]{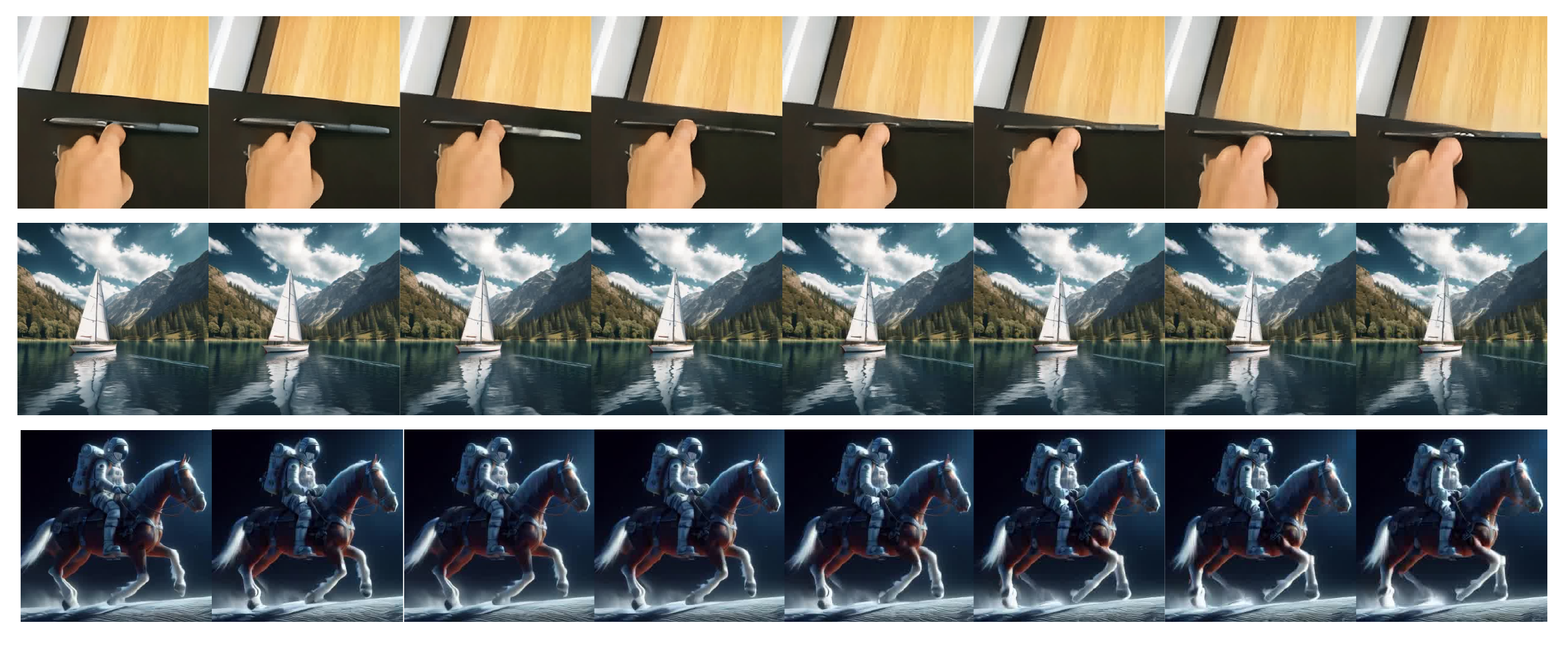}
	\caption{WildScrape~\cite{wei2023dreamvideo,feng2023dreamoving,xu2023magicanimate} sample visualization.}
\end{figure*}
\begin{figure*}[h]
	\centering
	\includegraphics[width=0.99\linewidth]{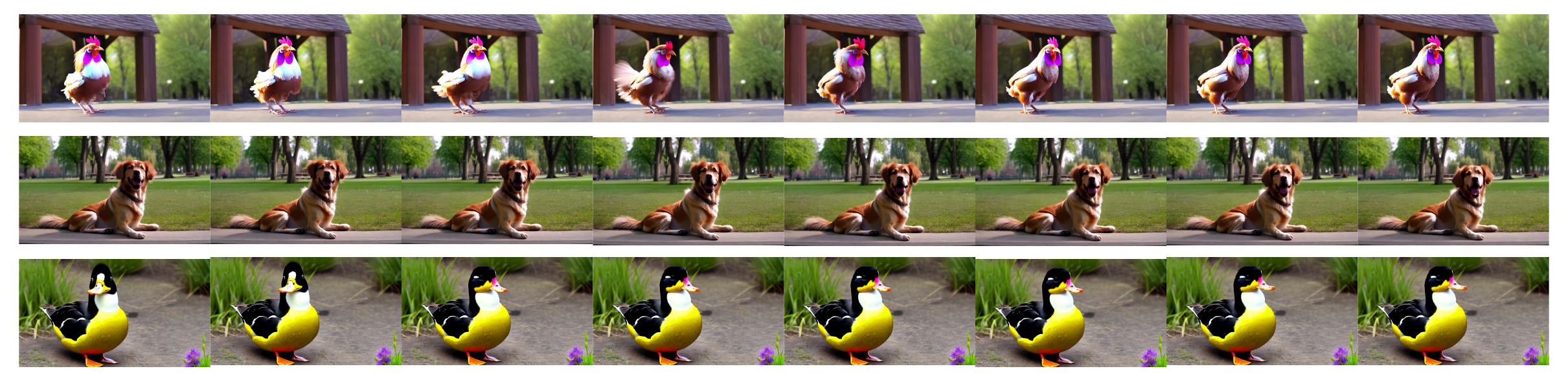}
	\caption{Crafter~\citep{chen2023videocrafter1} generated samples visualization.}
\end{figure*}

\end{document}